\documentclass{report}
\usepackage{authblk}
\usepackage{mathptmx}
\usepackage{url,latexsym,amsmath,amsthm,xspace,rotating,multirow,multicol,xspace,amssymb,paralist}
\usepackage{euscript}
\usepackage{fancybox,xcolor}
\usepackage{longtable}
\usepackage{paralist}
\usepackage[normalem]{ulem}
\usepackage[pdftex]{hyperref}
\usepackage{algorithmicx}
\usepackage{algpseudocode}
\usepackage{algorithm}
\usepackage{cancel}

\usepackage{url}
\usepackage{latexsym}

\usepackage{times}
\usepackage{amsmath}
\usepackage{amsthm}
\usepackage{amssymb}
\usepackage{graphicx}
\usepackage{xspace}
\usepackage{tabularx}
\usepackage{multicol}
\usepackage{multirow}
\usepackage{url}
\usepackage{wrapfig}
\usepackage{comment}
\usepackage{listings}
\usepackage{color}
\usepackage[utf8]{inputenc}
\usepackage{fancyvrb}
\usepackage{booktabs}
\usepackage{color}
\usepackage[normalem]{ulem}

\newcommand{\softmax}{\text{softmax}}

\newcommand{\bmx}[0]{\begin{bmatrix}}
\newcommand{\emx}[0]{\end{bmatrix}}

\newcommand{\vect}[1]{\mathbf{#1}}
\newcommand{\vects}[1]{\boldsymbol{#1}}
\newcommand{\matr}[1]{\mathbf{#1}}

\newcommand{\diag}[0]{\operatorname{diag}}

\newcommand{\va}[0]{\vect{a}}
\newcommand{\vb}[0]{\vect{b}}
\newcommand{\vc}[0]{\vect{c}}
\newcommand{\ve}[0]{\vect{e}}

\newcommand{\vh}[0]{\vect{h}}
\newcommand{\vv}[0]{\vect{v}}
\newcommand{\vx}[0]{\vect{x}}
\newcommand{\vp}[0]{\vect{p}}
\newcommand{\vz}[0]{\vect{z}}
\newcommand{\vw}[0]{\vect{w}}

\newcommand{\vf}[0]{\vect{f}}
\newcommand{\vi}[0]{\vect{i}}
\newcommand{\vo}[0]{\vect{o}}
\newcommand{\vd}[0]{\vect{d}}
\newcommand{\vy}[0]{\vect{y}}

\newcommand{\vu}[0]{\vect{u}}

\newcommand{\vr}[0]{\vect{r}}

\newcommand{\mW}[0]{\matr{W}}
\newcommand{\mE}[0]{\matr{E}}

\newcommand{\mX}[0]{\matr{X}}
\newcommand{\mY}[0]{\matr{Y}}
\newcommand{\mQ}[0]{\matr{Q}}
\newcommand{\mU}[0]{\matr{U}}

\newcommand{\mV}[0]{\matr{V}}

\newcommand{\mS}{\matr{S}}
\newcommand{\mI}{\matr{I}}

\newcommand{\TT}[0]{\vects{\theta}}

\newcommand{\vmu}[0]{\vects{\mu}}

\newcommand{\YY}[0]{\mathcal{Y}}
\newcommand{\BB}[0]{\mathcal{B}}

\newcommand{\HH}[0]{\mathcal{H}}
\newcommand{\RR}[0]{\mathbb{R}}
\newcommand{\MM}[0]{\mathcal{M}}
\newcommand{\OO}[0]{\mathbb{O}}
\newcommand{\II}[0]{\mathbb{I}}

\newcommand{\sigmoid}{\sigma}
\newcommand{\E}[0]{\mathbb{E}}

\newcommand{\eos}[0]{\ensuremath{\left< \text{eos}\right>}}

\newcommand{\dd}[1]{\ensuremath{\mbox{d}#1}}

\DeclareMathOperator*{\argmax}{\arg \max}

\newcommand{\BP}{\text{BP}}
\newcommand{\PPL}{\text{PPL}}
\newcommand{\PL}{\text{PL}}
\newcommand{\MatSum}{\text{MatSum}}
\newcommand{\MatMul}{\text{MatMul}}
\newcommand{\KL}{\text{KL}}
\newcommand{\data}{\text{data}}
\newcommand{\rect}{\text{rect}}
\newcommand{\maxout}{\text{maxout}}
\newcommand{\train}{\text{train}}
\newcommand{\val}{\text{val}}
\newcommand{\init}{\text{init}}
\newcommand{\fenc}{\text{fenc}}
\newcommand{\renc}{\text{renc}}
\newcommand{\enc}{\text{enc}}
\newcommand{\dec}{\text{dec}}
\newcommand{\test}{\text{test}}

\begin{document}

\title{Natural Language Understanding with Distributed Representation}
\author{Kyunghyun Cho}
\affil{
    Courant Institute of Mathematical Sciences and \\
    Center for Data Science,\\
    New York University 
}

\maketitle
\pagenumbering{arabic}

\abstract{
    This is a lecture note for the course DS-GA 3001 $\left<\right.$Natural
    Language Understanding with Distributed Representation$\left.\right>$ at the
    Center for Data Science\footnote{
        \url{http://cds.nyu.edu/}
    }, New York University in Fall, 2015. As the name of the course suggests,
    this lecture note introduces readers to a neural network based approach to
    natural language understanding/processing. In order to make it as
    self-contained as possible, I spend much time on describing basics of
    machine learning and neural networks, only after which how they are used for
    natural languages is introduced. On the language front, I almost solely
    focus on language modelling and machine translation, two of which I
    personally find most fascinating and most fundamental to natural language
    understanding.
    
    After about a month of lectures and about 40 pages of writing this lecture
    note, I found this fascinating note \cite{goldberg2015primer} by Yoav
    Goldberg on neural network models for natural language processing. This note
    deals with wider topics on natural language processing with
    distributed representations in more details, and I highly recommend you to read it
    (hopefully along with this lecture note.) I seriously wish Yoav had written
    it earlier so that I could've simply used his excellent note for my course.

    This lecture note had been written quite hastily as the course progressed,
    meaning that I could spare only about 100 hours in total for this note.
    This is my lame excuse for likely many mistakes in this lecture note, and I
    kindly ask for your understanding in advance. Again, how grateful I
    would've been had I found Yoav's note earlier.

    I am planning to update this lecture note gradually over time, hoping that I
    will be able to convince the Center for Data Science to let me teach the
    same course next year. The latest version will always be available both in
    pdf and in latex source code from
    \url{https://github.com/nyu-dl/NLP_DL_Lecture_Note}. The arXiv version will
    be updated whenever a major revision is made.

    I thank all the students and non-students who took\footnote{
        In fact, they are still taking the course as of 24 Nov 2015. They have
        two guest lectures and a final exam left until the end of the course.
    }
    this course and David Rosenberg for feedback.
}

\tableofcontents

\chapter{Introduction}
\label{chap:intro}

This lecture is going to be the only one where I discuss some philosophical,
meaning nonpractical, arguments, because according to Chris Manning and Hinrich
Schuetze, ``{\it even practically-minded people have to confront the issue of
what prior knowledge to try to build into their model}''
\cite{manning1999foundations}. 

\section{Route we will {\it not} take}
\label{sec:wrong_route}

\subsection{What is Language?}

The very first question we must ask ourselves before starting this course is the
question of what natural language is. Of course, the rest of this course does
not in any way require us to know what natural language is, but it is a
philosophical question I recommend everyone, including myself, to ponder upon
once a while. 

When I start talking about languages with anyone, there is a single person who
never misses to be mentioned, that is Noam Chomsky. His view has greatly
influenced the modern linguistics, and although many linguists I have talked to
claim that their work and field have long moved on from Chomsky's, I can feel
his shadow all over them. 

My first encounter with Chomsky was at the classroom of $<$Automata$>$ from my
early undergrad years. I was not the most attentive student back then, and all I
can remember is Chomsky's hierarchy and how it has shaped our view on languages,
in this context, programming/computer languages.  A large part of the course was
dedicated to explaining which class of languages emerges given a set of
constraints on a set of {\it generating rules}, or production rules. 

For instance, if we are given a set of generating rules that do not depend on
the context/meaning of non-terminal symbols (context-free grammar, CFG), we get
a context-free language. If we put a bit of constraints to CFG that each
generating rule is such that a non-terminal symbol is replaced by either a
terminal symbol, a terminal symbol by a non-terminal symbol or an empty symbol,
then we get a regular grammar. Similarly to CFG, we get a regular language from
the regular grammar, and the regular language is a subset of the context-free
language.

What Chomsky believes is that this kind of approach applies also to human
languages, or natural languages. There exists a set of generating rules that
{\it generates} a natural language. But, then, the obvious question to follow is
where those generating rules are. Where are they stored? How are they stored? Do
we have separate generating rules for different languages? 

\subsection{Language Understanding}
\label{sec:language_understanding_wrong}

\paragraph{Understanding Human Language}

Those questions are interesting, but out of scope for this course. Those
questions are the ones linguists try to answer. Generative linguistics aims at
figuring out what those rules are, how they are combined to form a valid
sentence, how they are adapted to different languages and so on. We will leave
these to linguists and continue on to our journey of {\it building a machine
that understands human languages}. 

\paragraph{Natural Language Understanding}

So, let's put these questions aside and trust Chomsky that we, humans, are
specially designed to store those generating rules somewhere in the brain
\cite{chomsky1959review,carnie2013syntax}. Or, better yet, let's trust Chomsky
that there's a universal grammar {\it built in} our brain. In other words, let's
say we were born with this set of generating rules for natural languages, and
while growing, we have adapted this universal grammar toward our native tongue
(language variation).

When we decide to speak of something (whatever that is and however implausible
that is), our brain quickly picks up a sequence of some of those generating
rules and starts generating a sentence accordingly. Of course, those rules do
not generate a sentence directly, but generates a sequence of control signals to
move our muscles to make sound. When heard by other people who understand your
language, the sound becomes a sentence.

In our case, we are more interested in a {\it machine} hearing that sound, or a
sentence from here on. When a machine heard this sentence, what would/should a
{\it language understanding machine} do to understand a language, or more simply
a sentence? Again, we are assuming that this sentence was generated from
applying a sequence of the existing generating rules. 

Under our assumption, a natural first step that comes to my mind is to figure
out that sequence of the generating rules which led to the sentence. Once the
sequence is found, or in a fancier term, inferred, the next step will be to
figure out what kind of mental state of the speaker led to those generating
rules. 

Let's take an example sentence ``{\it Our company is training workers}'' (from
Sec. 1.3 of \cite{manning1999foundations}), which is a horrible choice, because
this was used as an example of ambiguity in parsing. Regardless, a speaker
obviously has an awesome image of her company which trains its workers and wants
to tell a machine about this. This mental state is used to select the following
generating rules (assuming a phrase structure grammar)\footnote{
    Stanford Parser: \url{http://nlp.stanford.edu:8080/parser}
}:
\begin{verbatim}
    (ROOT
      (S
          (NP (PRP$ Our) (NN company))
              (VP (VBZ is)
                    (VP (VBG training)
                            (NP (NNS workers))))))
\end{verbatim}

\begin{figure}[ht]
    \centering
    \includegraphics[width=0.5\textwidth]{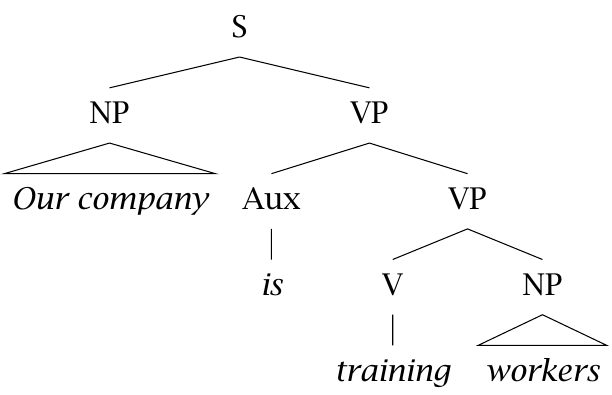}
    \caption{A parse of ``{\it Our company is training workers}''}
    \label{fig:manning_parse}
\end{figure}

The machine hears the sentence ``{\it Our company is training workers}'' and
infers the parse in Fig.~\ref{fig:manning_parse}. Then, we can make a simple set
of rules (again!) to let the machine answer questions about this sentence, kinds
of questions that imply that the machine has understood the sentence (language).
For instance, given a question ``{\it Who is training workers?}'', the machine
can answer by noticing that the question is asking for the subject of the verb
phrase
``{\it is training}'' acted on the object ``{\it workers}'' and that the subject
is ``{\it Our company}''.

\paragraph{Side Note: Bayesian Language Understanding} This generative view of
languages fits quite well with Bayesian modelling (see, e.g.,
\cite{perfors2006poverty}.) There exists a hidden mechanism, or a set of
generating rules and a rule governing their composition, which can be modelled
as a latent variable $Z$.  Given these rules, a language or a sentence $X$ is
generated according to the conditional distribution $P(X|Z)$. Then,
understanding language (by humans) is equivalent to computing the posterior
distribution over all possible sets of generating rules and their compositional
rules (i.e., $P(Z|X)$.) This answers the question of what is the most likely
mechanism underlying the observed language.

Furthermore, from the perspective of machines, Bayesian approach is attractive.
In this case, we assume to know {\it the} set of rules in advance and let the
latent variable $Z$ denote the specific configuration (use) of those rules.
Given this sequence of applying the rules, a sentence $X$ is generated via the
conditional distribution $P(X|Z)$. Machine understanding of language is
equivalent to inferring the posterior distribution over $Z$ given $X$.

For more details about Bayesian approaches (in the context of machine learning),
please, refer to \cite{bishop2006pattern} or take the course DS-GA 1005
Inference and Representation by Prof. David Sontag.

\paragraph{Understanding vs. Using} 
What's clear from this example is that in this generative view of languages,
there is a clear separation between understanding and using. Inferring the
generating rules from a given sentence is {\it understanding}, and answering a
question based on this understanding, {\it using}, is a separate activity.
Understanding part is done when the underlying (true) structure has been
determined regardless of how this understanding be used.

To put it in a slightly different wording, language understanding does not
require its use, or downstream tasks. In this road that we will {\em not} take
in this course, understanding exists as it is, regardless of what the understood
insight/knowledge will be used for. And, this is the reason why we do not walk
down this road.

\section{Road we {\it will} take}
\label{sec:intro}

\subsection{Language as a Function}

In this course, we will view a natural/human language as ``{\it a system
intended to communicate ideas from a speaker to a hearer}''
\cite{winograd1972understanding}. What this means is that we do not view a
language as a separate entity that exists on its own. Rather, we view a whole
system or behaviour of {\it communication} as a language. Furthermore, this view
dictates that we must take into account the world surrounding a speaker and a
hearer in order to understand language.

Under this view of language, language or rather its usage become somewhat
similar to action or behaviour. Speaking of something is equivalent to acting on
a listener, as both of them influence the listener in one way or another. The
purpose of language is then to influence another by efficiently communicate
one's will or intention.\footnote{
    Chomsky does not agree: ``{\it it is wrong to think of human use of language
    as characteristically informative, in fact or in intention.}''
    \cite{chomsky1968linguistic}.
} This hints at how language came to be (or may have come to be): (evolution)
language has evolved to facilitate the exchange of ideas among people (learning)
humans learn language by being either encouraged or punished for the use of
language. This latter view on how language came to be is similar in spirit to
the behaviourism of B. F. Skinner (``{\it necessary mediation of reinforcement by
another organism}'' \cite{skinner2014verbal}.)

This is a radical departure from the generative view of human language, where
language existed on its own and its understanding does not necessarily require
the existence of the outside world nor the existence of a listener. It is no
wonder why Chomsky was so harsh in criticizing Skinner's work in
\cite{chomsky1959review}. 
This departure, as I see it, is the departure toward a functional view of
language. {\it Language is a function of communication}. 

\subsection{Language Understanding as a Function Approximation}

Let's make a large jump here such that we consider this function as a
mathematical function.  This function (called language) takes as input the state
of the surrounding world, the speaker's speech, either written, spoken or signed
and the listener's mental state\footnote{
    We assume here that a such thing exists however it is represented in
    our brain.
} Inside the function, the listener's mental state is updated to incorporate the
new idea from the speaker's speech. The function then returns a response by the
listener (which may include ``no response'' as well) and a set of non-verbal
action sequences (what would be the action sequence if the speaker insulted
the listener?).

In this case, language understanding, both from humans' and machines'
perspective, boils down to figuring out the internal working of this function.
In other words, we understand language by learning the internal mechanism of the
function.  Furthermore, this view suggests that the underlying structures of
language are heavily dependent on the surrounding environment (context) as well
as on the target task. The former (context dependence) is quite clear, as the
function takes as input the context, but the latter may be confusing now.
Hopefully, this will become clearer later in the course.

How can we approximate this function? How can we figure out the internal working
mechanism of this function? What tools do we have?

\paragraph{Language Understanding by Machine Learning}
This functional view of languages suddenly makes machine learning a very
appealing tool for understanding human languages. After all, function
approximation is {\em the} core of machine learning. Classification is a
classical example of function approximation, clustering is a function
approximation where the target is not given, generative modeling learns a
function that returns a probability of an input, and so on.

When we approximate a function in machine learning, the prime ingredient is
data. We are given data which was either generated from this function
(unsupervised learning) or well fit this function (supervised learning), based
on which we adjust our approximation to the function, often iteratively, to best
fit the data. But, I must note here that it does not matter how well the approximated
function fits the data it was fitted to, but matters how well this approximation fits
{\em unseen} data.\footnote{
    This is a matter of generalization, and we will talk
    about this more throughout the course.
}

In language understanding, this means that we collect a large data set of input
and output pairs (or conversations together with the recording of the
surrounding environment) and fit some arbitrary function to well predict the
output given an input. We probably want to evaluate this approximation in a
novel conversation. If this function makes a conversation just like a person,
voil\`{a}, we made a machine that passed the Turing test.  Simple, right?

\paragraph{Problem}
Unfortunately, as soon as we try to do this, we run into a big problem. This
problem is not from machine learning nor languages, but the definition of this
function of language.

Properly approximating this function requires us to either simulate or record
the whole world (in fact, the whole universe.) For, this function takes as input
and maintains as internal state the surrounding world (context) and the mental
state of the individual (speaker.) This is unavoidable, if we wanted to very
well approximate this function as a whole.

It is unclear, however, whether we want to approximate the full function. For a
human to survive, yes, it is likely that the full function is needed. But, if
our goal is restricted to a certain task (such as translation, language
modelling, and so on), we may not want to approximate this function fully. We
probably want to approximate only a subset of this whole function. For instance,
if our goal is to understand the process of translation from one language to
another, we can perhaps ignore all but the speech input to the function and all
but the speech output from the function, because often a (trained) person can
translate a sentence in one language to another without knowing the whole
context.

This latter approach to language understanding--approximating a partial function
of languages-- will be at the core of this course. We will talk about various
language tasks that are a part of this whole function of language. These tasks
will include, but are not limited to, language modelling, machine translation,
image/video description generation and question answering. For these tasks and
potentially more, we will study how to use machine learning, or more
specifically deep learning, to solve these tasks by approximating sub-functions
of language.

\chapter{Function Approximation as Supervised Learning}
\label{chap:function_approx}

Throughout this course, we will extensively use artificial neural
networks\footnote{
    From here on, I will simply drop artificial and call them neural networks.
    Whenever I say ``neural network'', it refers to artificial neural networks.
}
to approximate (a part of) the function of natural language. This makes it
necessary for us to study the basics of neural networks first, and this lecture
and a couple of subsequent ones are designed to serve this purpose.

\section{Function Approximation: Parametric Approach}

\subsection{Expected Cost Function}

Let us start by defining a data distribution $p_{\text{data}}$.
$p_{\text{data}}$ is defined over a pair of input and output vectors, $\vx \in
\II^d$ and $\vy \in \OO^k$, respectively. $\II$ and $\OO$ are respectively sets
of all possible input and output values, such as $\RR$, $\left\{ 0, 1\right\}$
and $\left\{0, 1, \ldots, L\right\}$. This data distribution is not known to us.

The goal is to find a relationship between $\vx$ and $\vy$. More specifically,
we are interested in finding a function $f:\RR^d \to \OO^k$ that generates the
output $\vy$ given its corresponding input $\vx$. The very first thing we should
do is to put some constraints on the function $f$ to make our search for the
correct $f$ a bit less impossible. In this lecture, and throughout the course, I
will consider only a parametric function $f$, in which case the function is
fully specified with a set of parameters $\TT$.

Next, we must define a way to measure how well the function $f$ approximates the
underlying mechanism of generation ($\vx \to \vy$). Let's denote by $\hat{\vy}$ the 
output of the function with a particular set $\TT$ of parameters and a given input
$\vx$:
\begin{align*}
    \hat{\vy} = f_{\TT}(\vx)
\end{align*}
How well $f$ approximates the true generating function is equivalent to how far
$\hat{\vy}$ is from the correct output $\vy$. Let's use $D(\hat{\vy}, \vy)$ for
now call this distance\footnote{
    Note that we do not require this distance to satisfy the triangular
    inequality, meaning that it does not have to be a distance. However, I will
    just call it distance for now.
}
between $\hat{\vy}$ and $\vy$

It is clear that we want to find $\TT$ that minimizes $D(\hat{\vy}, \vy)$ for
every pair in the space ($RR^d \times \OO^k$). But, wait, every pair equally
likely? Probably not, for we do not care how well $f_{\TT}$ approximates the
true function, when a pair of input $\vx$ and output $\vy$ is unlikely, meaning
we do not care how bad the approximation is, if $p_\text{data}(\vx, \vy)$ is
small. However, this is a bit difficult to take into account, as we must decided
on the threshold below which we consider any pair irrelevant.

Hence, we {\em weight} the distance between the approximated $\hat{\vy}$ and the
correct $\vy$ of each pair $(\vx, \vy)$ in the space by its probability $p(\vx,
\vy)$. Mathematically saying, we want to find 
\begin{align*}
    \arg\min_{\TT} \int_{\vx} \int_{\vy} p_{\text{data}}(\vx, \vy) D(\hat{\vy}, \vy) \dd{\vx}
    \dd{\vy},
\end{align*}
where the integral $\int$ should be replaced with the summation $\sum$ if any of
$\vx$ and $\vy$ is discrete.

We call this quantity being minimized with respect to the parameters $\TT$ a
cost function $C(\TT)$. This is equivalent to computing the {\em expected}
distance between the predicted output $\hat{\vy}$ and the correct one $\vy$:
\begin{align}
    \label{eq:expected_cost}
    C(\TT) =& \int_{\vx} \int_{\vy} p_{\text{data}}(\vx, \vy) D(\hat{\vy}, \vy) \dd{\vx}
    \dd{\vy}, \\
    =& \E_{(\vx,\vy) \sim p_{\text{data}}}\left[ D(\hat{\vy}, \vy) \right] 
\end{align}
This is often called an expected loss or risk, and minimizing this cost function
is referred to as {\em expected risk minimization}~\cite{Vapnik1995}.

Unfortunately $C(\TT)$ cannot be (exactly) computed for a number of reasons. The
most important reason among them is simply that we don't know what the data
distribution $p_{\text{data}}$ is. Even if we have access to $p_{\text{data}}$,
we can exactly compute $C(\TT)$ only with heavy assumptions on both the data
distribution and the distance function.\footnote{Why?} 

\subsection{Empirical Cost Function}

This does not mean that we are doomed from the beginning. Instead of the
full-blown description of the data distribution $p_{\text{data}}$, we will
assume that someone miraculously gave us a finite set of pairs drawn from the
data distribution. We will call this a training set:
\begin{align*}
    \left\{ (\vx^1, \vy^1), \ldots, (\vx^N, \vy^N) \right\}.  
\end{align*}

As we have access to the samples from the data distribution, we can use Monte
Carlo method to approximate the expected cost function $C(\TT)$ such that
\begin{align}
    \label{eq:empirical_cost}
    C(\TT) \approx \tilde{C}(\TT) = \frac{1}{N} \sum_{n=1}^N D(\hat{\vy^{n}},
    \vy^{n}).
\end{align}
We call this approximate $\tilde{C}(\TT)$ of the expected cost function, an
empirical cost function (or empirical risk or empirical loss.)

Because empirical cost function is readily computable, we will mainly work with
the empirical cost function not with the expected cost function. However, keep
in mind that at the end of the day, the goal is to find a set of parameters that
minimizes the {\em expected} cost.

\section{Learning as Optimization}

We often call this process of finding a good set of parameters that minimizes
the expected cost {\em learning}. This term is used from the perspective of a
machine which implements the function $f_{\TT}$, as it {\em learns} to
approximate the true generating function $f$ from training data.

From what I have described so far, it may have become clear even without me
mentioning that learning is {\em optimization}. We have a clearly defined
function (the empirical cost function $\tilde{C}$) which needs to be minimized
with respect to its input $\TT$.

\subsection{Gradient-based Local Iterative Optimization}
There are many optimization algorithms one can use to find a set of parameters
that minimizes $\tilde{C}$. Sometimes, you can even find the optimal set of
parameters in a closed form equation.\footnote{
    One such example is a linear regression where 
    \begin{itemize}
        \item $f_{\TT=\left\{ \mW\right\}}(\vx) = \mW \vx$
        \item $D(\hat{\vy}, \vy) = \frac{1}{2} \| \hat{\vy} - \vy \|^2$
    \end{itemize}
    In this case, the optimal $\mW$ is 
    \begin{align}
        \label{eq:opt_lin}
        \mW = \mY \mX^\top (\mX \mX^\top)^{-1},
    \end{align}
    where
    \begin{align*}
        \mX = \left[ \vx^1; \ldots ;\vx^N\right], \mY = \left[ \vy^1; \ldots ;\vy^N\right]. 
    \end{align*}

    Try it yourself! 
}
In most cases, because there is no known closed-form solution, it is typical to
use an iterative optimization algorithm (see \cite{Fletcher1987} for in-depth
discussion on optimization.) 

By an {\em iterative} optimization, I mean an algorithm which refines its
estimate of the optimal set of parameters little by little until the values of
the parameters converge to the optimal (expected) cost function. Also, it is
worthwhile to note that most iterative optimization algorithms are {\em local},
in the sense that they do not require us to evaluate the whole parameter space,
but only a small subset along the path from the starting point to the
convergence point.\footnote{
    There are {\em global} optimization algorithms, but they are out of scope
    for this course. See, for instance, \cite{Brochu2010} for one such algorithm
    called Bayesian optimization.
}

Here I will describe the simplest one among those local iterative optimization
algorithms, called gradient descent (GD) algorithm. As the name suggests, this
algorithm depends entirely on the gradient of the cost function.\footnote{
    From here on, I will use the cost function to refer to the {\em empirical}
    cost function.
} 

\begin{figure}
    \centering
    \begin{minipage}{0.6\textwidth}
        \includegraphics[width=\columnwidth]{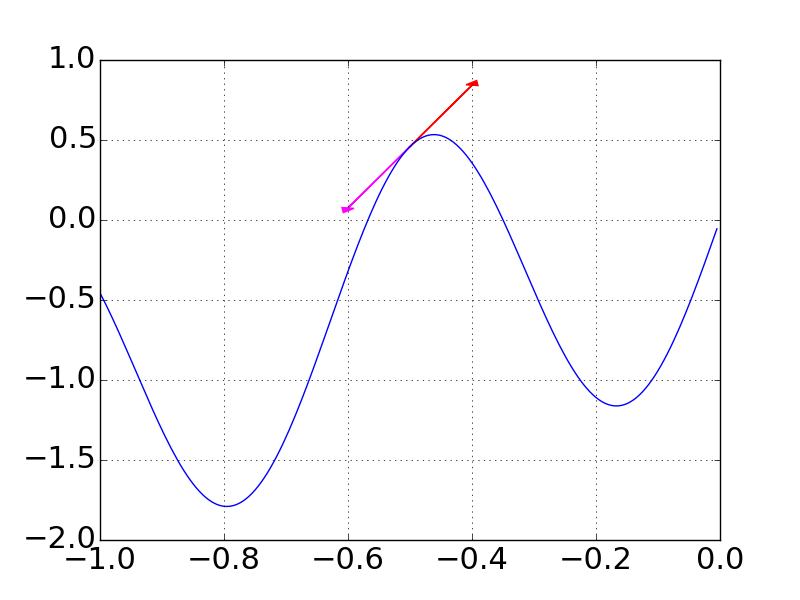}
    \end{minipage}
    \hfill
    \begin{minipage}{0.39\textwidth}
        \caption{(\textcolor{blue}{blue}) $f(x) = \sin(10 x) + x$.
            (\textcolor{red}{red}) a gradient at $x=-0.6$.
        (\textcolor{magenta}{magenta}) a negative gradient at $x=-0.6$.}
        \label{fig:grad}
    \end{minipage}
\end{figure}

The gradient of a function $\nabla \tilde{C}$ is a vector whose direction points
to the direction of the greatest rate of increase in the function's value and
whose magnitude measures this rate. At each point $\TT_t$ in the parameter
space, the gradient of the cost function $\nabla \tilde{C}(\TT_t)$ is the {\em
opposite} direction toward which we want to move the parameters. See
Fig.~\ref{fig:grad} for graphical illustration.

One important point of GD that needs to be mentioned here is on how large a step
one takes each time. As clear from the magenta line (the direction opposite to
the direction given by the gradient) in Fig.~\ref{fig:grad}, if too large a step
is taken toward the negative gradient direction, the optimization process will
overshoot and miss the (local) minimum around $x=-0.8$. This step size, or
sometimes called learning rate, $\eta$ is one most important hyperparameter of
the GD algorithm.

Now we have all the ingredients for the GD algorithm: $\nabla \tilde{C}$ and
$\eta$. The GD algorithm iterates the following step:
\begin{align}
    \label{eq:GD}
    \TT \leftarrow \TT - \eta \nabla \tilde{C}(\TT).
\end{align}
The iteration continues until a certain stopping criterion is met, which we will
discuss shortly.

\subsection{Stochastic Gradient Descent}
\label{sec:sgd}

This simple GD algorithm works surprisingly quite well, and it is a fundamental
basis upon which many advanced optimization algorithms have been built. I will
present a list of few of those advanced algorithms later on and discuss them
briefly. But, before going into those advanced algorithms, let's solve one tiny,
but significant issue of the GD algorithm.

This tiny, but significant issue arises especially often in machine learning.
That is, it is computationally very expensive to compute $\tilde{C}$ and
consequently its gradient $\nabla \tilde{C}$, thanks to the ever increasing size
of the training set $D$. 

Why is the growing size of the training set making it more and more
computationally demanding to compute $\tilde{C}$ and $\nabla \tilde{C}$?  This
is because both of them are essentially the sum of as many per-sample costs as
there are examples in the training set. In other words,
\begin{align*}
    &\tilde{C}(\TT) = \frac{1}{N} \sum_{n=1}^N \tilde{C}(\vx^n, \vy^n | \TT), \\
    &\nabla \tilde{C}(\TT) = \frac{1}{N} \sum_{n=1}^N \nabla \tilde{C}(\vx^n, \vy^n | \TT).
\end{align*}
And, $N$ goes up to millions or billions very easily these days.

This enormous computational cost involved in each GD step has motivated the {\em
stochastic gradient descent} (SGD) algorithm~\cite{Robbins1951,Bottou1998}. 

First, recall from Eq.~\eqref{eq:empirical_cost} that the cost function we
minimize is the {\em empirical} cost function $\tilde{C}$ which is the
sample-based approximation to the {\em expected} cost function $C$. This
approximation was done by assuming that the training examples were drawn
randomly from the data distribution $p_{\text{data}}$:
\begin{align*}
    C(\TT) \approx \tilde{C}(\TT) = \frac{1}{N} \sum_{n=1}^N D(\hat{\vy^{n}},
    \vy^{n}).
\end{align*}
In fact, as long as this assumption on the training set holds, we can always
approximate the expected cost function with a fewer number of training examples:
\begin{align*}
    C(\TT) \approx \tilde{C}_\MM (\TT) = \frac{1}{|\MM|} \sum_{m \in \MM}
    D(\hat{\vy^{m}}, \vy^m),
\end{align*}
where $M \ll N$ and $\MM$ is the indices of the examples in this much smaller
subset of the training set. We call this small subset a {\em minibatch}.

Similarly, this leads to a minibatch-based estimate of the gradient as well:
\begin{align*}
    \nabla \tilde{C}_\MM (\TT) = \frac{1}{|\MM|} \sum_{m \in \MM} \nabla
    D(\hat{\vy^m}, \vy^m).
\end{align*}

It must now be clear to you where I am headed toward. At each GD step, instead
of using the full training set, we will use a small subset $\MM$ which is
randomly selected to compute the gradient estimate. In other words, we use
$\tilde{C}_\MM$ instead of $\tilde{C}$, and $\nabla \tilde{C}_\MM$ instead of
$\nabla \tilde{C}$, in Eq.~\eqref{eq:GD}. 

Because computing $\tilde{C}_\MM$ and $\nabla \tilde{C}_\MM$ is independent of
the size of the training set, we can use SGD to make as many steps as we want
without worrying about the growing size of training examples. This is highly
beneficial, as regardless of how many training examples you used to compute the
gradient, we can only take a tiny step toward that descending direction.
Furthermore, the increased level of noisy in the gradient estimate due to the
small sample size has been suspected to help reaching a better solution in 
high-dimensional non-convex problems (such as those in training deep neural
networks) \cite{Lecun1998a}.\footnote{
    Why would this be the case? It is worth thinking about this issue further.
}

We can set $M$ to be any constant, and in an extreme, we can set it to $1$ as
well. In this case, we call it online SGD.\footnote{
    Okay, this is not true in a strict sense. SGD is an online algorithm with
    $M=1$ originally, and using $M>1$, is a variant of SGD, often called,
    minibatch SGD. However, as using minibatches ($M>1$) is almost always the
    case in practice, I will refer to minibatch SGD as SGD, and to the original
    SGD as online SGD.
} Surprisingly, already in 1951, it was shown that using a single example each
time is enough for the SGD to converge to a minimum (under certain conditions,
obviously) \cite{Robbins1951}.

This SGD algorithm will be at the core of this course and will be discussed
further in the future lectures.

\section{When do we stop learning?}
\label{sec:model_selection}

From here on, I assume that we approximate the ground truth function by
iteratively refining its set of parameters, in most cases using {\em stochastic
gradient descent}. In other words, learning of a machine that approximates the
true generating function $f$ happens gradually as the machine goes over the
training examples little by little over time.

Let us go over again what kind of constraints/issue we have first:
\begin{enumerate}
    \item Lack of access to the expected cost function $C(\TT)$
    \item Computationally expensive empirical cost function $\tilde{C}(\TT)$
    \item (Potential) non-convexity of the empirical cost function
        $\tilde{C}(\TT)$
\end{enumerate}

The most severe issue is that we do not have access to the expected cost
function which is the one we want to minimize in order to work well with {\em
any} pair of input $\vx$ and output $\vy$. Instead, we have access to the
empirical cost function which is a finite sample approximation to the expected
cost function. 

\begin{figure}
    \centering
    \begin{minipage}{0.6\textwidth}
        \includegraphics[width=\columnwidth]{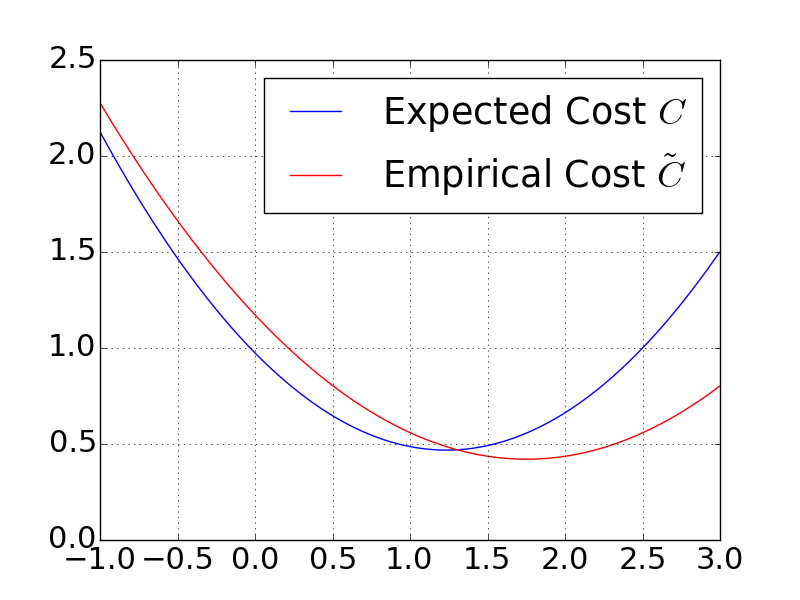}
    \end{minipage}
    \hfill
    \begin{minipage}{0.39\textwidth}
        \caption{(\textcolor{blue}{blue}) Expected cost function $C(\TT)$.
            (\textcolor{red}{red}) Empirical cost function $\tilde{C}(\TT)$.
            The underlying true generating function was $f(x) = \sin(10x) + x$.
            The cost function uses the squared Euclidean distance.
            The empirical cost function was computed based on 10 noisy
            examples of which $x$'s were sampled from the uniform distribution
        between $0$ and $1$. For each sample input $x$, noise from zero-mean Gaussian
    distribution with standard deviation $0.01$ was added to $f(x)$ to emulate
the noisy measurement channel.}
        \label{fig:cost}
    \end{minipage}
\end{figure}

Why is this a problem? Because, we do not have a guarantee that the (local)
minimum of the empirical cost function corresponds to the (local) minimum of the
expected cost function. An example of this mismatch between the expected and
empirical cost functions is shown in Fig.~\ref{fig:cost}.

As in the case shown in Fig.~\ref{fig:cost}, it is not desirable to minimize the
empirical cost function perfectly. The parameters that perfectly minimize the
empirical cost function (in the case of
Fig.~\ref{fig:cost}, the slope $a$ of a linear function $f(x) = a x$) will
likely be a sub-optimal cost for the expected cost function about which we
really care. 

\subsection{Early Stopping}
What should we do? There are many ways to avoid this weird contradiction where
we want to optimize the cost function well but not too well. Among those, one
most important trick is {\em early stopping}, which is only applicable when
iterative optimization is used.

First, we will split the training set $D$ into two partitions $D_{\train}$ and
$D_{\val}$.\footnote{
    Later on, we will split it further into three partitions.
}
We call them a training set and a validation set, respectively. In practice it
is a good idea to keep $D$ much larger than $D'$, because of the reasons that
will become clear shortly.

Further, let us define the training cost as
\begin{align}
    \label{eq:train_c}
    \tilde{C}(\TT) = C_{\train}(\TT) = \frac{1}{|D_{\train}|} \sum_{(x,y) \in
    D_{\train}} D_{\train}(\hat{y}, y),
\end{align}
and the validation cost as
\begin{align}
    \label{eq:val_c}
    C_{\val}(\TT) = \frac{1}{|D_{\val}|} \sum_{(x,y) \in D_{\val}} D(\hat{y}, y).
\end{align}
With these two cost functions we are all ready to use early stopping now. 

After every few updates using SGD (or GD), the validation cost function is
evaluated with the current set of parameters. The parameters are updated (i.e.,
the training cost function is optimized) until the validation cost does not
decrease, or starts to increase instead of decreasing.

That's it! It is almost free, as long as the size of the validation set is
reasonable, since each evaluation is at most as expensive as computing the
gradient of the empirical cost function. Because of the simplicity and
effectiveness, this early stopping strategy has become {\em de facto} standard
in deep learning and in general machine learning.

The question that needs to be asked here is what the validation cost function
does here. Clearly, it approximates the expected cost function $C$, similarly to
the empirical cost function $\tilde{C}$ as well as the training cost function
$C_{\train}$. In the infinite limit of the size of either training or validation
set, they should coincide, but in the case of a finite set, those two cost
functions differ by the noise in sampling (sampling pairs from the data
distribution) and observation (noise in $\vy=f(\vx)$.)

The fact that we explicitly optimize the training cost function implies that
there is a possibility (in fact, almost surely in practice) that the set of
parameters found by this optimization process may capture not only the
underlying generating function but also noise in the observation and sampling
procedure. This is an issue, because we want our machine to approximate the true
generating function not the noise process involved. 

The validation cost function measures both the true generating structure as well
as noise injected during sampling and observation. However, assuming that noise
is not correlated with the underlying generating function, noise introduced in
the validation cost function differs from that in the training cost function. In
other words, the set of parameters that perfectly minimizes the training cost
function (thereby capturing even noise in the training set) will be penalized
when measured by the validation cost function. 

\subsection{Model Selection}
\label{sec:hypothesis_space}

In fact, the use of the validation cost does not stop at the early stopping.
Rather, it has a more general role in model selection. First, we must talk about
model selection itself.

This whole procedure of optimization, or learning, can be cast as a process of
searching for the best {\em hypothesis} over the entire space $\HH$ of
hypotheses.  Here, each hypothesis corresponds to each possible function (with a
unique set of parameters and a unique functional form) that takes the input
$\vx$ and output $\vy$. In the case of regression ($\vx \in \RR^d$ and $\vy \in
\RR$), the hypothesis space includes an $n$-th order polynomial function 
\begin{align*}
    f(x) = \sum_{\sum_{k=1}^d i_k = n, i_k \geq 0}
    a_{i_1,i_2,\ldots,i_k} \prod_{k'=1}^d x_{k'}^{i_k},
\end{align*}
where $a_{i_1,i_2,\ldots,i_k}$'s are the coefficients, and 
any other functional form that you can imagine as long as it can process $\vx$
and return a real-valued scalar.  In the case of neural networks, this space
includes all the possible model architectures which are defined by the number of
layers, the type of nonlinearities, the number of hidden units in each layer and
so on. 

Let us use
$M \in \HH$ to denote one hypothesis.\footnote{
    $M$, because each hypothesis corresponds to one learning {\em machine}.
} One important thing to remember is that the parameter space is only a subset
of the hypothesis space, because the parameter space is defined by a family of
hypotheses (the parameter space of a linear function cannot include a set of
parameters for a second-order polynomial function.)

Given a definition of expected cost function, we can {\em score} each hypothesis
$M$ by the corresponding cost $C_M$. Then, the whole goal of function
approximation boils down to the search for a hypothesis $M$ with the minimal
expected cost function $C$. But, of course, we do not have access to the
expected cost function and resort to the empirical cost function based on a
given training set. 

The optimization-based approach we discussed so far searches for the best
hypothesis based on the empirical cost iteratively. However, because of the
issue of {\em overfitting} which means that the optimization algorithm overshot
and missed the local minimum of the expected cost function (because it was aimed
at the local minimum of the empirical cost function), I introduced the concept
of early stopping based on the validation cost.

This is unfortunately not satisfactory, as we have only searched for the best
hypothesis inside a small subset of the whole hypothesis space $\HH$. What if
another subset of the hypothesis space includes a function that better suits the
underlying generating function $f$? Are we doomed?

It is clearly better to try more than one subsets of the hypothesis space. For
instance, for a regression task, we can try linear functions ($\HH_1$),
quadratic (second-order polynomial) functions ($\HH_2$) and sinusoidal functions
($\HH_3$). Let's say for each of these subsets, we found the best hypothesis
(using iterative optimization and early stopping); $M_{\HH_1}$, $M_{\HH_2}$ and
$M_{\HH_3}$. Then, the question is how we should choose one of those hypotheses.

Similar to what we've done with early stopping, we can use the validation cost
to compare these hypotheses. Among those three we choose one that has the
smallest validation cost $C_{\val}(M)$.

This is one way to do {\em model selection}, and we will talk about another way
to do this later.

\section{Evaluation}

But, wait, if this is an argument for using the validation cost to {\em early
stop} the optimization (or learning), one needs to notice something weird. What
is it?

Because we used the validation cost to stop the optimization, there is a chance
that the set of parameters we found is optimal for the validation set (whose
structure consists of both the true generating function and sampling/observation
noise), but not to the general data distribution. This means that we cannot tell
whether the function estimate $\hat{f}$ approximating the true generating
function $f$ is a good fit by simply early stopping based on the validation
cost.  Once the optimization is done, we need yet another metric to see how well
the learned function estimate $\hat{f}$ approximates $f$.

Therefore, we need to split the training set not into two partitions but into
{\em three} partitions. We call them a training set $D_{\train}$, a validation
set $D_{\val}$ and a test set $D_{\test}$. Consequently, we will have three cost
functions; a training cost function $C_{\train}$, a validation cost function
$C_{\val}$ and a test cost function $C_{\test}$, similarly to
Eqs.~\ref{eq:train_c}--\ref{eq:val_c}.

This test cost function is the one we use to compare different hypotheses, or
models, fairly. Any hypothesis that worked best in terms of the test cost is the
one that you choose.

\paragraph{Let's not Cheat}
One most important lesson here is that you {\em must never look at a test set}.
As soon as you take a peak at the test set, it will influence your choice in the
model structure as well as any other hyperparameters biasing toward a better
test cost. The best option is to never ever look at the test set until it is
absolutely needed (e.g., need to present your result.)

\section{Linear Regression for Non-Linear Functions}

Let us start with a simple linear function to approximate a true generating
function such that
\begin{align*}
    \hat{\vy} = f(\vx) = \mW^\top \vx, 
\end{align*}
where $\mW \in \RR^{d \times l}$ is the weight matrix. In this case, this weight
matrix is the only parameter, i.e., $\TT=\left\{ \mW \right\}$.

The empirical cost function is then
\begin{align*}
    \tilde{C}(\TT) = \frac{1}{N} \sum_{n=1}^N \frac{1}{2} \left\| \vy^n - \mW^\top \vx^n
    \right\|^2_2.
\end{align*}

The gradient of the empirical cost function is
\begin{align}
    \label{eq:grad_lin}
    \nabla \tilde{C}(\TT) = -\frac{1}{N} \sum_{n=1}^N 
    \left( \vy^n - \mW^\top \vx^n\right)^\top \vx^n.
\end{align}

With these two well defined, we can use the iterative optimization algorithm,
such as GD or SGD, to find the best $\mW$ that minimizes the empirical cost
function.\footnote{
    In fact, looking at Eq.~\eqref{eq:grad_lin}, it's quite clear that you can
    compute the optimal $\mW$ analytically. See Eq.~\eqref{eq:opt_lin}.
}
Or, better is to use a validation set to stop the optimization
algorithm at the point of the minimal validation cost function (remember early
stopping?)

Now, but we are not too satisfied with a linear network, are we?

\subsection{Feature Extraction}
\label{sec:feature_extraction}

Why are we not satisfied?

First, we are not sure whether the true generating function $f$ was a linear
function. If it is not, can we expect linear regression to approximate the true
function well? Of course, not. We will talk about this shortly.

Second, because we were {\em given} $\vx$ (meaning we did not have much control
over what we want to measure as $\vx$), it is unclear how well $\vx$ represents
the input. For instance, consider doing a sales forecast of air conditioner at
one store which opened five years ago. The input $x$ is the number of days since
the opening date of the store (1 Jan 2009), and the output $y$ is the number of
units sold on each day.

Clearly, in this example, the relationship between $x$ and $y$ is not linear.
Furthermore, perhaps the most important feature for predicting the sales of air
conditioners is missing from the input $x$, which is a month (or a season, if
you prefer.) It is likely that the sales bottoms out during the winter (perhaps
sometime around December, January and February,) and it hits the peak during
summer months (around May, June and July.) In other words, if we look at how far
the month is away from July, we can predict the sales quite well even with 
linear regression.

Let us call this quantity $\phi(x)$, or equivalent {\em feature}, such that
\begin{align}
    \label{eq:feat_month}
    \phi(x) = \left| m(x) - \alpha \right|,
\end{align}
where $m(x) \in \left\{ 1,2,\ldots,12\right\}$ is the month of $x$ and
$\alpha=5.5$.  With this
feature, we can fit linear regression to better approximate the sales figure of
air conditioners. Furthermore, we can add yet another feature to improve the
predictive performance. For instance, one such feature can be which day of week
$x$ is. 

This whole process of extracting a good set of features that will make our
choice of parametric function family (such as linear regression in this case) is
called {\em feature extraction}. This feature extraction is an important step in
machine learning and has often been at the core of many applications such as
computer vision (the representative example is SIFT~\cite{lowe1999object}.)

Feature extraction often requires heavy knowledge of the domain in which this
function approximation is applied. To use linear regression for computer vision,
it is a good idea to use computer vision knowledge to extract a good set of
features. If we want to use it for environmental problems, we must first notice
which features must be important and how they should be represented for linear
regression to work. 

This is okay for a machine learning practitioner in a particular field, because
the person has in-depth knowledge about the field. There are however many cases
where there's simply not enough domain knowledge to exploit. To make the matter
worse, it is likely that the domain knowledge is not correct, making the whole
business of using manually extracted features futile.

\chapter{Neural Networks and Backpropagation Algorithm}
\label{chap:nn}

\section{Conditional Distribution Approximation}
\label{sec:distribution_approx}

I have mainly described so far as if the function we approximate or the function
we use to approximate returns only a constant value, as in one point $\vy$ in
the output space. This is however not true, and in fact, the function can return
anything including a {\em
distribution}~\cite{Bridle1990,denker1991transforming,bishop1994mixture}.

Let's first decompose the data distribution $p_{\data}$ into the product
of two terms:
\begin{align*}
    p_{\data}(\vx, \vy) = p_{\data}(\vx) p_{\data}(\vy |\vx).
\end{align*}
It becomes clear that one way to sample from $p_{\data}$ is to sample an input
$\vx^n$ from $p_{\data}(\vx)$ and subsequently sample the corresponding output
$\vy^n$ from the conditional distribution $p_{\data}(\vy | \vx^n)$. 

This implies that the function approximation of the generating function ($f:
\vx \to \vy$) is effectively equivalent to approximating the conditional
distribution $p_{\data}(\vy | \vx)$. This may suddenly sound much more
complicated, but it should not alarm you at all.  As long as we choose to use a
distribution parametrized by a small number of parameters to approximate the
conditional distribution $p_{\data}(\vy | \vx)$, this is quite manageable
without almost any modification to the expected and empirical cost functions we
have discussed. 

Let us use $\TT(\vx)$ to denote a set of parameters for the probability
distribution $\tilde{p}(\vy|\vx, \TT(\vx))$ approximating the true, underlying
probability distribution $p_{\data}(\vy|\vx)$. As the notation suggests, the
function now returns {\em the parameters of the distribution} $\TT(\vx)$ given
the input $\vx$. 

For example, let's say $\vy \in \left\{ 0, 1\right\}^k$ is a binary vector and
we chose to use independent Bernoulli distribution to approximate the
conditional distribution $p_{\data}(\vy | \vx)$. In this case, the parameters
that define the conditional distribution are the means of $k$ dimensions:
\begin{align}
    \label{eq:bernoulli}
    \tilde{p}(\vy | \vx) = \prod_{k'=1}^k p(y_{k'}|\vx) = 
    \prod_{k'=1}^k \mu_{k'}^{y_{k'}} (1-\mu_{k'})^{1-y_{k'}}.
\end{align}
Then the function $\TT(\vx)$ should output a $k$-dimensional vector of which
each element is between 0 and 1.

Another example: let's say $\vy \in \RR^k$ is a real-valued vector. It is quite
natural to use a Gaussian distribution with a diagonal covariance matrix to
approximate the conditional distribution $p(\vy | \vx)$:
\begin{align}
    \label{eq:gaussian}
    \tilde{p}(\vy | \vx) = \prod_{k'=1}^k \frac{1}{\sqrt{2\pi}\sigma_{k'}} 
\exp\left( \frac{(y_{k'} - \mu_{k'})^2}{2\sigma_{k'}^2} \right).
\end{align}
The parameters for this conditional distribution are $\TT(\vx) = \left\{ 
    \mu_1, \mu_2, \ldots, \mu_k, \sigma_1, \sigma_2, \ldots, \sigma_k
\right\}$, where $\mu_k \in \RR$ and $\sigma_k \in \RR_{>0}$.

In this case of probability approximation, it is natural to use
Kullback-Leibler (KL) divergence to measure the distance.\footnote{
    Again, we use a loose definition of the distance where triangular inequality
    is not enforced.
} 
The KL divergence from one distribution $P$ to the other $Q$ is
defined\footnote{
    Why don't I say the KL divergence between two distributions here? Because,
    the KL divergence is not a symmetric measure, i.e., $\KL(P\|Q) \neq
    \KL(Q\|P)$.
}
by
\begin{align*}
    \KL(P\|Q) = \int P(\vx) \log \frac{P(\vx)}{Q(\vx)} \dd{\vx}.
\end{align*}
In our case of function/distribution approximation, we want to minimize the KL
divergence from the data distribution $p_{\data}(\vy|\vx)$ to the approximate
distribution $\tilde{p}(\vy | \vx)$ averaged over the data distribution
$p_{\data}(\vx)$:
\begin{align*}
    C(\TT) = \int p_{\data}(\vx) \KL(p_{\data}\|\tilde{p}) \dd{\vx} = 
    \int p_{\data}(\vx) \int p_{\data}(\vy|\vx) \log \frac{p_{\data}
    (\vy|\vx)}{\tilde{p}(\vy|\vx)} \dd{\vy} \dd{\vx}.
\end{align*}
But again we do not have access to $p_{\data}$ and cannot compute this expected
cost function.

Similarly to how we defined the empirical cost function earlier, we must
approximate this expected KL divergence using the training set: 
\begin{align}
    \label{eq:kl_train}
    \tilde{C}(\TT) = \frac{1}{N} \sum_{n=1}^N -\log \tilde{p}(\vy^n|\vx^n).
\end{align}
As an example, if we choose to return the binary vector $\vy$ as in
Eq.~\eqref{eq:bernoulli}, the empirical cost function will be
\begin{align*}
    \tilde{C}(\TT) = -\frac{1}{N} \sum_{n=1}^N 
    \sum_{k'=1}^k y_{k'} \log \mu_{k'} + (1-y_{k'}) \log (1 - \mu_{k'}),
\end{align*}
which is often called a {\em cross entropy cost}. In the case of
Eq.~\eqref{eq:gaussian}, 
\begin{align}
    \label{eq:kl_gaussian}
    \tilde{C}(\TT) = -\frac{1}{N} \sum_{n=1}^N 
    \sum_{k'=1}^k \frac{(y_{k'} - \mu_{k'})^2}{2\sigma_{k'}^2} - \log
        \sigma_{k'}.
\end{align}

Do you see something interesting in Eq.~\eqref{eq:kl_gaussian}? If we assume
that the function outputs $1$ for all $\sigma_{k'}$'s, we see that this cost
function reduces to that using the Euclidean distance between the true output
$\vy$ and the mean $\vmu$. What does this mean?

There will be many occasions later on to discuss more about this perspective
when we discuss language modelling. However, one thing we must keep in our mind
is that there is nothing different between approximating a function and a
distribution.

\subsection{Why do we want to do this?}

Before we move on to the main topic of today's lecture, let's try to understand
why we want to output the distribution. Unlike returning a single point in the
space, the distribution returned by the function $f$ incorporates both the most
likely outcome $\hat{y}$ as well as the uncertainty associated with this value.

In the case of the Gaussian output in Eq.~\eqref{eq:gaussian}, the standard
deviation $\sigma_{k'}$, or the variance $\sigma_{k'}^2$, indicates how
uncertain the function is about the output centered at $\mu_{k'}$. Similarly,
the mean $\mu_{k'}$ of the Bernoulli output in Eq.~\eqref{eq:bernoulli} is
directly proportional to the function's confidence in predicting that the
$k'$-th dimension of the output is $1$.

\begin{figure}[ht]
    \centering
    \begin{minipage}{0.6\textwidth}
        \includegraphics[width=\columnwidth]{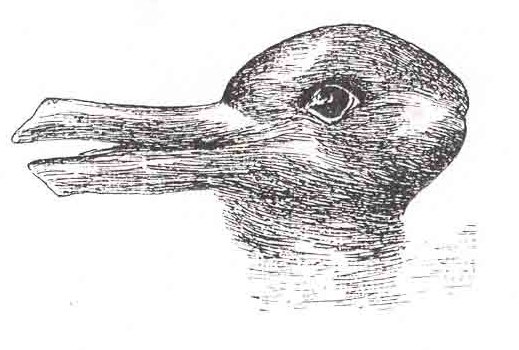}
    \end{minipage}
    \hfill
    \begin{minipage}{0.39\textwidth}
        \caption{Is this a duck or a rabbit? \cite{kuhn2012structure} At the end
        of the day, we want our function $f$ to return a conditional
    distribution saying that $p(\text{duck}|\vx) = p(\text{rabbit}|\vx)$,
instead of returning {\em the} answer out of these two possible answers.}
        \label{fig:duck_rabbit}
    \end{minipage}
\end{figure}

This is useful in many aspects, but one important aspect is that it reflects the
natural uncertainty of the underlying generating function. One input $\vx$ may
be interpreted in more than one ways, leading to two possible outputs, which
happens more often than not in the real world. For instance, the famous picture
in Fig.~\ref{fig:duck_rabbit} can be viewed as a picture of a duck or a picture
of a rabbit, in which case the function needs to output the probability
distribution by which the same probability mass is assigned to both a duck and a
rabbit. Furthermore, there is observational noise that cannot easily be
identified and ignored by the function, in which case the function should return
the uncertainty due to the observational noise along with the most likely (or
the average) prediction.

\subsection{Other Distributions}
\label{sec:other_dist}

I have described two distributions (densities) that are widely used: 
\begin{itemize}
    \itemsep 0em
    \item Bernoulli distribution: binary classification
    \item Gaussian distribution: real value regression
\end{itemize}
Here, let me present one more distribution which we will use almost everyday
through this course.

\paragraph{Categorical Distribution: Multi-Class Classification}

Multi-class classification is a task in which each example belongs to one of $K$
classes. For each input $x$, the problem reduces to find a probability $p_k(x)$
of the $k$-th class under the constraint that 
\begin{align*}
    \sum_{k=1}^K p_k(x)=1
\end{align*}

It is clear that in this case, the function $f$ returns $K$ values $\left\{
\mu_1, \mu_2, \ldots, \mu_K \right\}$, each of which is between $0$ and $1$.
Furthermore, the sum of $\mu_k$'s must sum to $1$. This can be achieved easily
by letting $f$ to compute affine transformation of $x$ (or $\phi(x)$) to return
$K$ (unbounded) real values followed by a so called {\em softmax}
function~\cite{Bridle1990}:
\begin{align}
    \label{eq:softmax}
    \mu_k = \frac{\exp(w_k^\top \phi(x) + b_k)}{\sum_{k'=1}^K \exp(w_{k'}^\top
    \phi(x) + b_k)},
\end{align}
where $w_k \in \RR^{\dim(\phi(x))}$ and $b_k \in \RR$ are the parameters of
affine transformation.

In this case, the (empirical) cost function based on the KL divergence is
\begin{align}
    \label{eq:cat_crossentropy}
    C(\TT) = -\frac{1}{N} \sum_{n=1}^N \sum_{k=1}^K \II_{k=y^n} \mu_k,
\end{align}
where 
\begin{align}
    \label{eq:indicator_function}
    \II_{k=y^n} = \left\{ 
        \begin{array}{c c}
            1, & \text{if }k=y^n \\
            0, & \text{otherwise}
        \end{array}
        \right.
\end{align}

\section{Feature Extraction is also a Function}
\label{sec:feature_ext}

We talked about the manual feature extraction in the previous lecture (see
Sec.~\ref{sec:feature_extraction}. But, this is quite unsatisfactory, because
this whole process of manual feature extraction is heavily dependent on the
domain knowledge, meaning that we cannot have a generic principle on which we
design features. This raises a question: instead of manually designing features
ourselves, is it possible for this to happen automatically?

One thing we notice is that the feature extraction process $\phi(\vx)$ is
nothing but a {\em function}. A function of a function is a function, right? In
other words, we will extend our definition of the function to include the
feature extraction function:
\begin{align*}
    \hat{\vy} = f(\phi(\vx)).
\end{align*}

We will assume that the feature extraction function $\phi$ is also parametrized,
and its parameters are included in the set of parameters which includes those of
$f$. As an example, $\alpha$ in Eq.~\eqref{eq:feat_month} is a parameter of the
feature extraction $\phi$.

A natural next question is which family of parametric functions we should use
for $\phi$. We run into the same issue we talked about earlier in
Sec.~\ref{sec:model_selection}: the size of hypothesis space is simply too
large!

Instead of choosing one great feature extraction function, we can go for a stack
of simple transformations which are all learned.\footnote{
    A great article about this was posted recently in
    \url{http://colah.github.io/posts/2014-03-NN-Manifolds-Topology/}.
}
Each transformation can be as simple as affine transformation followed by a
simple point-wise nonlinearity:
\begin{align}
    \label{eq:layer}
    \phi_0(\vx) = g(\mW_0 \vx + \vb_0),
\end{align}
where $\mW_0$ is the weight matrix, $\vb_0$ is the bias and $g$ is a point-wise
nonlinearity such as $\tanh$.\footnote{
    Some of the widely used nonlinearities are
    \begin{itemize}
        \item Sigmoid: $\sigma(x) = \frac{1}{1+\exp(-x)}$
        \item Hyperbolic function: $\tanh(x) = \frac{1-\exp(-2x)}{1+\exp(-2x)}$
        \item Rectified linear unit: $\rect(x) = \max(0, x)$
    \end{itemize}
}

One interesting thing is that if the dimensionality of the transformed feature
vector $\phi_0(\vx)$ is {\em much} larger than that of $\vx$, the function
$f(\phi_0(\vx))$ can approximate any function from $\vx$ to $\vy$ under some
assumptions, even when the parameters $\mW_0$ and $\vb_0$ are randomly
selected!~\cite{Cover1965} 

The problem solved, right? We just put a huge matrix $\mW_0$, apply some
nonlinear function $g$ to it and fit linear regression as I described earlier.
We don't even need to touch $\mW_0$ and $\vb_0$. All we need to do is replace
the input $\vx^n$ of all the pairs in the training set to $\phi_0(\vx^n)$.

In fact, there is a group of researchers claiming to have figured this out by
themselves less than a decade ago (as of 2015) who call this model an {\em
extreme learning machine}~\cite{Huang2006}. There have been some debates about
this so-called extreme learning machine. Here I will not make any comment
myself, but would be a good exercise for you to figure out why there has been
debates about this.

But, regardlessly, this is not what we want.\footnote{
    And, more importantly, I will not accept any final project proposal whose
    main model is based on the ELM.
}
What we want is to fully tune the whole thing.

\section{Multilayer Perceptron}
\label{sec:mlp}

The basic idea of multilayer perceptron is to stack a large number of those
feature extraction {\em layers} in Eq.~\eqref{eq:layer} between the input and
the output. This idea is as old as the whole field of neural network research,
dating back to early 1960s~\cite{Rosenblatt1962}. However, it took many more
years for people to figure out a way to tune the whole network, both $f$ and
$\phi$'s together. See \cite{schmidhuber2015deep} and \cite{lecun2015deep}, if you
are interested in the history.

\subsection{Example: Binary classification with a single hidden unit}

Let us start with the simplest example. The input $x \in \RR$ is a real-valued
scalar, and the output $y \in \left\{0,1\right\}$ is a binary value
corresponding to the input's label. The feature extractor $\phi$ is defined as
\begin{align}
    \label{eq:phi}
    \phi(x) = \sigma(u x + c),
\end{align}
where $u$ and $c$ are the parameters. The function $f$ returns the mean of the
Bernoulli conditional distribution $p(y|x)$:
\begin{align}
    \label{eq:f}
    \mu = f(x) = \sigma(w \phi(x) + b).
\end{align}
In both of these equations, $\sigma$ is a sigmoid function:
\begin{align}
    \label{eq:sigmoid}
    \sigma(x) = \frac{1}{1+\exp(-x)}.
\end{align}

We use the KL divergence to measure the distance between the true conditional
distribution $p(y|x)$ and the predicted conditional distribution $\hat{p}(y|x)$.
\begin{align*}
    \KL(p\|\hat{p}) =& \sum_{y \in \left\{0, 1\right\}} p(y|x) \log
    \frac{p(y|x)}{\hat{p}(y|x)} \\
    =& \sum_{y \in \left\{0, 1\right\}} p(y|x) \log p(y|x) - p(y|x) \log
    {\hat{p}(y|x)}.
\end{align*}
Note that the first term in the summation $p(y|x) \log p(y|x)$ can be safely
ignored in our case. Why? Because, this does not concern $\tilde{p}$ which is
one we change in order to minimize this KL divergence.

Let's approximate this KL divergence with a single sample from $p(y|x)$ and
leave only the relevant part. We will call this a per-sample cost:
\begin{align}
    \label{eq:cost_bernoulli}
    C_x =& - \log \hat{p}(y|x) \\
    =& - \log \mu^y (1-\mu)^{1-y} \\
    =& - y \log \mu - (1-y) \log (1-\mu),
\end{align}
where $\mu$ is from Eq.~\eqref{eq:f}.  It is okay to work with this per-sample
cost function instead of the full cost function, because the full cost function
is almost always the (unweighted) sum of these per-sample cost functions. See
Eq.~\eqref{eq:empirical_cost}.

We now need to compute the gradient of this cost function $C_x$ with respect to
all the parameters $w$, $b$, $u$ and $c$. First, let's start with $w$:
\begin{align*}
    \frac{\partial C_x}{\partial w} = \frac{\partial C_x}{\partial \mu}
    \frac{\partial \mu}{\partial \underline{\mu}}
    \frac{\partial \underline{\mu}}{\partial w},
\end{align*}
which is a simple application of chain rule of derivatives. Compare this to
\begin{align*}
    \frac{\partial C_x}{\partial b} = \frac{\partial C_x}{\partial \mu}
    \frac{\partial \mu}{\partial \underline{\mu}}
    \frac{\partial \underline{\mu}}{\partial b}.
\end{align*}
In both equations, $\underline{\mu} = w \phi(x) + b$ which is the input to $f$.

Both of these derivatives {\em share} $\frac{\partial C_x}{\partial
\mu}\frac{\partial \mu}{\partial \underline{\mu}}$, where
\begin{align}
    \label{eq:out_deriv}
    \frac{\partial C_x}{\partial \mu} \underbrace{\frac{\partial \mu}{\partial
    \underline{\mu}}}_{=\mu'} = 
    -\frac{y}{\mu}\mu' +
    \frac{1-y}{1-\mu}\mu' 
    = \frac{-y + y\mu + \mu - y\mu }{\mu(1-\mu)}\mu' 
    = \frac{\mu - y}{\mu(1-\mu)}\mu' 
    = \mu - y,
\end{align}
because the derivative of the sigmoid function $\frac{\partial \mu}{\partial
\underline{\mu}}$ is
\begin{align*}
    \mu' = \mu (1-\mu).
\end{align*}
Note that this corresponds to computing the difference between the correct label
$y$ and the predicted label (probability) $\mu$.

Given this output derivative $\frac{\partial C_x}{\partial \underline{\mu}}$, all we need to
compute are
\begin{align*}
    &\frac{\partial \underline{\mu}}{\partial w} = \phi(x) \\
    &\frac{\partial \underline{\mu}}{\partial b} = 1.
\end{align*}
From these computations, we see that
\begin{align}
    \label{eq:c_w}
    &\frac{\partial C_x}{\partial w} = (\mu - y) \phi(x), \\
    \label{eq:c_b}
    &\frac{\partial C_x}{\partial b} = (\mu - y).
\end{align}

Let us continue on to $u$ and $c$. We can again rewrite the derivatives w.r.t.
these into
\begin{align*}
    \frac{\partial C_x}{\partial u} =& \frac{\partial C_x}{\partial
    \underline{\mu}}
    \frac{\partial \underline{\mu}}{\partial \phi}
    \frac{\partial \phi}{\partial \underline{\phi}}
    \frac{\partial \underline{\phi}}{\partial u} \\
    \frac{\partial C_x}{\partial c} =& \frac{\partial C_x}{\partial
    \underline{\mu}}
    \frac{\partial \underline{\mu}}{\partial \phi}
    \frac{\partial \phi}{\partial \underline{\phi}}
    \frac{\partial \underline{\phi}}{\partial c},
\end{align*}
where $\underline{\phi}$ is the input to $\phi$ similarly to $\underline{\mu}$
was to the input to $\mu$.

There are two things to notice here. First, we already have $\frac{\partial
C_x}{\partial \underline{\mu}}$ from computing the derivatives w.r.t. $w$ and
$b$, meaning there is no need to re-compute it. Second, $\frac{\partial
\underline{\mu}}{\partial \underline{\phi}}$ is shared between the derivatives
w.r.t. $u$ and $c$.

Therefore, we first compute $\frac{\partial \underline{\mu}}{\partial
\underline{\phi}}$:
\begin{align*}
    \frac{\partial \underline{\mu}}{\partial \phi}\underbrace{\frac{\partial \phi}{\partial
    \underline{\phi}}}_{=\phi'} = w \phi' = w \phi(x) (1- \phi(x))
\end{align*}
Next, we compute 
\begin{align*}
    &\frac{\partial \underline{\phi}}{\partial u} = x \\
    &\frac{\partial \underline{\phi}}{\partial c} = 1.
\end{align*}

Now all the ingredients are there:
\begin{align*}
    \frac{\partial C_x}{\partial u} =& (\mu -y) w \phi(x) (1-\phi(x)) x \\
    \frac{\partial C_x}{\partial c} =& (\mu -y) w \phi(x) (1-\phi(x)).
\end{align*}

The most important lession to learn from here is that most of the computations
needed to get the derivatives in this seemingly complicated multilayered
computational graph (multilayer perceptron) are {\em shared}. At the end of the
day, the amount of computation needed to compute the gradient of the cost
function w.r.t. all the parameters in the network is only as expensive as
computing the cost function itself.

\subsection{Example: Binary classification with more than one hidden units}
\label{sec:example2}

Let us try to generalize this simple, or rather simplest model, into a slightly
more general setting. We will still look at the binary classification but with
multiple hidden units and a multidimensional input such that:
\begin{align*}
    \phi(x) = U x + c,
\end{align*}
where $U \in \RR^{l \times d}$ and $c \in \RR^l$. Consequently, $w$ will be a
$l$-dimensional vector.

The output derivative $\frac{\partial C_x}{\partial \mu}\frac{\partial
\mu}{\partial \underline{\mu}}$ stays same as before. See
Eq.~\eqref{eq:out_deriv}.
However, we note that the derivative of $\underline{\mu}$ with respect to $w$
should now differ, because it's a vector.\footnote{
    The Matrix Cookbook \cite{petersen2008matrix} is a good reference for this
    section.
}
Let's look at what this means.

The $\underline{\mu}$ can be expressed as
\begin{align}
    \label{eq:mu_vec}
    \underline{\mu} = w^\top \phi(x) + b = \sum_{i=1}^l w_i \phi_i(x) + b.
\end{align}
In this case, we can start computing the derivative with respect to each element
of $w_i$ separately:
\begin{align*}
    \frac{\partial \underline{\mu}}{\partial w_i} = \phi_i(x),
\end{align*}
and will put them into a vector:
\begin{align*}
    \frac{\partial \underline{\mu}}{\partial w} = \left[ 
        \frac{\partial \underline{\mu}}{\partial w_1}, \frac{\partial
        \underline{\mu}}{\partial w_2}, \ldots, \frac{\partial
        \underline{\mu}}{\partial w_l}
    \right]^\top = \left[ \phi_1(x), \phi_2(x), \ldots, \phi_l(x) \right]^\top
    = \phi(x)
\end{align*}

Then, the derivative of the cost function $C_y$ with respect to $w$ can be
written as
\begin{align*}
    \frac{\partial C_y}{\partial w} = (\mu -y) \phi(x),
\end{align*}
in which case nothing really changed from the case of a single hidden unit in
Eq.~\eqref{eq:c_w}.

Now, let's look at $\frac{\partial C_y}{\partial \phi}$. Again, because
$\phi(x)$ is now a vector, there has to be some changes. Because $\frac{\partial
C_y}{\partial \underline{\mu}}$ is already computed, we only need to look at
$\frac{\partial \underline{\mu}}{\partial \phi}$. In fact, the procedure for
computing this is identical to that for computing $\frac{\partial
\underline{\mu}}{\partial w}$ due to the symmetry in Eq.~\eqref{eq:mu_vec}. That
is,
\begin{align*}
    \frac{\partial \underline{\mu}}{\partial \phi} = w
\end{align*}

Next, what about $\frac{\partial \phi}{\partial \underline{\phi}}$? Because the
nonlinear activation function $\sigma$ is applied element-wise, we can simply
compute this derivative for each element in $\phi(x)$ such that
\begin{align*}
    \frac{\partial \phi}{\partial \underline{\phi}} = 
    \diag\left(\left[ \phi_1'(x), \phi_2'(x), \ldots, \phi_l'(x)
    \right]^\top\right),
\end{align*}
where $\diag$ returns a diagonal matrix of the input vector. In short, we will
denote this as $\phi'$

Overall so far, we have got
\begin{align*}
    \frac{\partial C_y}{\partial \underline{\phi}} = 
    (\mu - y)w^\top \phi'(x) = (\mu - y) (w \odot \diag(\phi'(x))),
\end{align*}
where $\odot$ is an element-wise multiplication.

Now it is time to compute $\frac{\partial
\underline{\phi}}{\partial U}$:
\begin{align*}
    \frac{\partial \underline{\phi}}{\partial U} = \frac{\partial U^\top
    x}{\partial U} = x,
\end{align*}
according to the Matrix Cookbook \cite{petersen2008matrix}. Then, let's look at
the whole derivative w.r.t. $U$:
\begin{align*}
    \frac{\partial C_y}{\partial U} = (\mu - y) (w \odot \diag(\phi'(x)))
    x^\top.
\end{align*}
Note that all the vectors in this lecture note are {\em column} vectors.

For $c$, it's straightforward, since 
\begin{align*}
    \frac{\partial \underline{\phi}}{\partial c} = 1.
\end{align*}

\section{Automating Backpropagation}
\label{sec:autodiff}

This procedure, presented as two examples, is called a {\em backpropagation}
algorithm. If you read textbooks on neural networks, you see a fancier way to
explain this backpropagation algorithm by introducing a lot of fancy terms such
as {\em local error} $\delta$ and so on. But, personally I find it much easier
to understand backpropagation as a clever application of the chain rule of
derivatives to a directed acyclic graph (DAG) in which each node computes a certain
function $\phi$ using the output of the previous nodes. I will refer to this
DAG as a computational graph from here on.

\begin{figure}[ht]
    \centering
    \begin{minipage}{0.48\textwidth}
        \centering
        \includegraphics[width=0.9\textwidth]{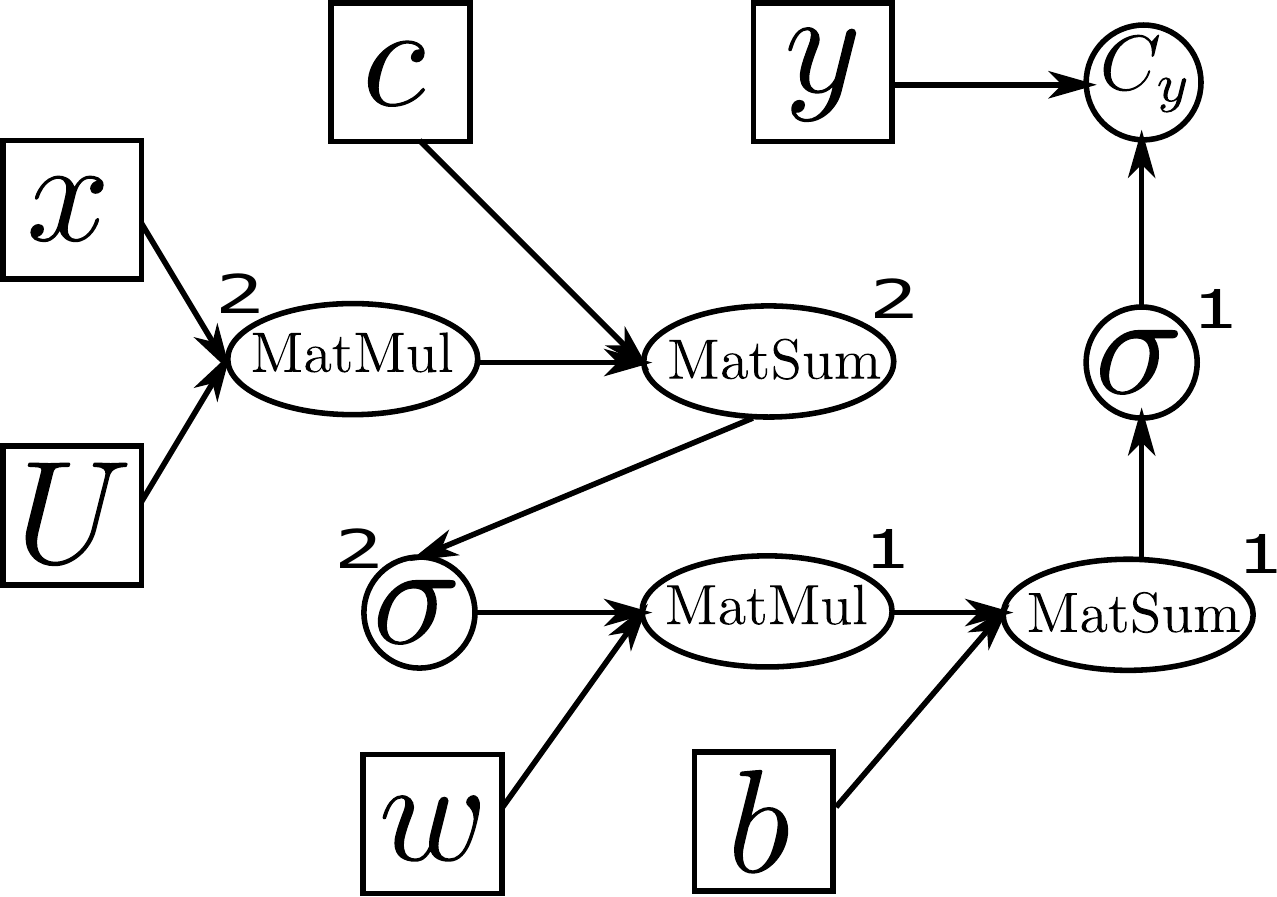}
    \end{minipage}
    \hfill
    \begin{minipage}{0.48\textwidth}
        \centering
        \includegraphics[width=0.9\textwidth]{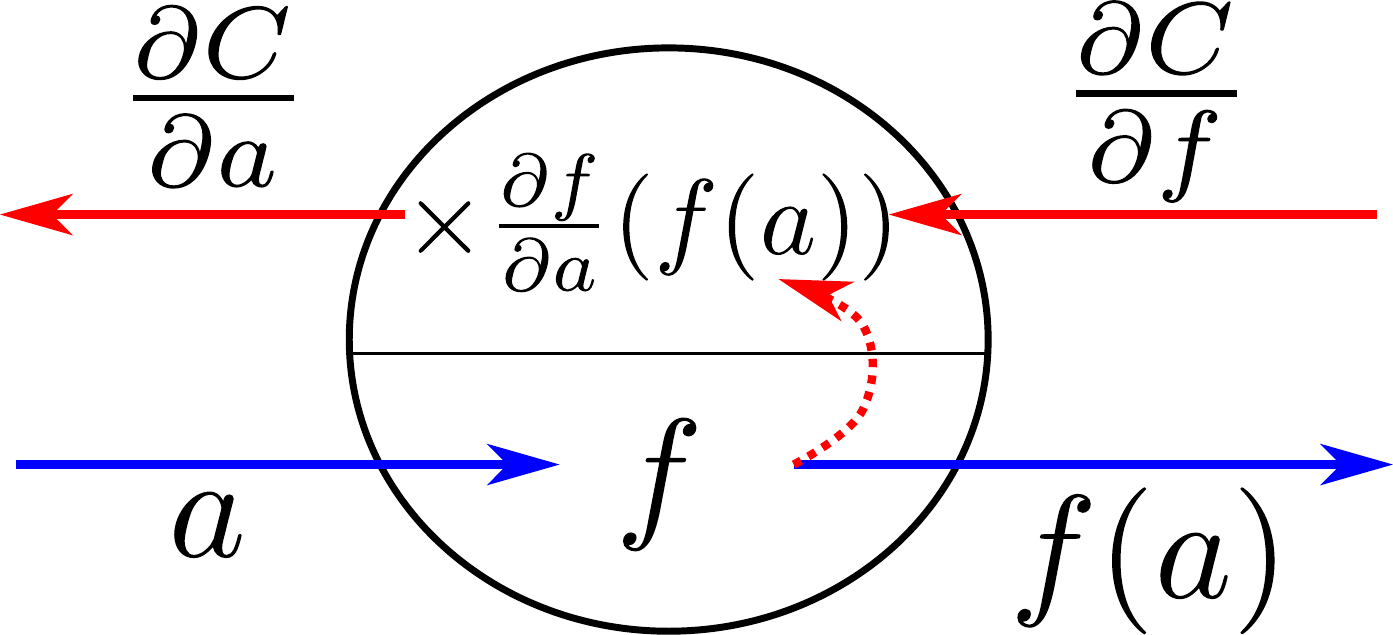}
    \end{minipage}

    \begin{minipage}{0.48\textwidth}
        \centering
        (a)
    \end{minipage}
    \hfill
    \begin{minipage}{0.48\textwidth}
        \centering
        (b)
    \end{minipage}
    \caption{(a) A graphical representation of the computational graph of the
    example network from Sec.~\ref{sec:example2}. (b) A graphical illustration
    of a function node ({\color{blue} $\to$}: forward pass, {\color{red}
$\leftarrow$}: backward pass.)}
    \label{fig:comp_graph}
\end{figure}

A typical computational graph looks like the one in Fig.~\ref{fig:comp_graph}~(a).
This computational graph has two types of nodes; (1) function node ($\bigcirc$)
and (2) variable node ($\Box$). There are four different types of function
nodes; (1) $\text{MatMul}(A,B) = AB$, (2) $\text{MatSum}(A,B)=A+B$, (3)
$\sigma$: element-wise sigmoid function  and (4) $C_y$: cost node. The variables
nodes correspond to either parameters or data ($x$ and $y$.) Each function node
has a number associated with it to distinguish between the nodes of the same
function.

Now, in this computational graph, let us start computing the gradient using the
backpropagation algorithm. We start from the last code, $C_y$, by computing
$\frac{\partial C_y}{\partial y}$ and $\frac{\partial C_y}{\partial \sigma^1}$.
Then, the function node $\sigma^1$ will compute its own derivative
$\frac{\partial \sigma^1}{\partial \text{MatSum}^1}$ and multiply it with
$\frac{\partial C_y}{\partial \sigma^1}$ passed {\em back} from the function
node $C_y$. So far we've computed
\begin{align}
    \label{eq:dC_dMS1}
    \frac{\partial C_y}{\partial \text{MatSum}^1} = \frac{\partial C_y}{\partial \sigma^1} \frac{\partial \sigma^1}{\partial
    \text{MatSum}^1}
\end{align}

The function node MatSum$^1$ has two inputs $b$ and the output of MatMul$^1$.
Thus, this node computes two derivatives $\frac{\partial
\text{MatSum}^1}{\partial b}$ and $\frac{\partial \text{MatSum}^1}{\partial
\text{MatMul}^1}$. Each of these is multiplied with the backpropagated
derivative $\frac{\partial C_y}{\partial \text{MatSum}^1}$ from
Eq.~\eqref{eq:dC_dMS1}. At this point, we already have the derivative of the
cost function $C_y$ w.r.t. one of the parameters $b$:
\begin{align*}
    \frac{\partial C_y}{\partial b} = \frac{\partial C_y}{\partial
    \text{MatSum}^1} \frac{\partial \text{MatSum}^1}{\partial b}
\end{align*}

This process continues mechanically until the very beginning of the graph (a set
of root variable nodes) is reached. All we need in this process of
backpropagating the derivatives is that each function node implements both {\em
forward computation} as well as {\em backward computation}. In the backward
computation, the function node received the derivative from the next function
node, evaluates its own derivative with respect to the inputs (at the point of
the forward activation) and passes theses
derivatives to the corresponding previous nodes. See
Fig.~\ref{fig:comp_graph}~(b) for the graphical illustration.

Importantly, the inner mechanism of a function node does not change depending on
its context (or equivalently where the node is placed in a computational graph.)
In other words, if each type of function nodes is implemented in advance, it
becomes trivial to build a complicated neural network (including multilayer
perceptrons) and compute the gradient of the cost function (which is one such
function node in the graph) with respect to all the parameters as well as all
the inputs.

This is a special case, called the reverse mode, of automatic
differentiation.\footnote{
    If anyone's interested in digging more into the whole field of automatic
    differentiation, try to Google it and you'll find tons of materials. One
    such reference is \cite{baydin2015automatic}.
} It is probably the most valuable tool in deep learning, and fortunately many
widely used toolkits such as Theano~\cite{bergstra2010theano,bastien2012theano}
have implemented this reverse mode of automatic differentiation with an
extensive number of function nodes used in deep learning everyday.

Before finishing this discussion on automating backpropagation, I'd like you to
think of pushing this even further. For instance, you can think of each function
node returning not its numerical derivative on its backward pass, but a
computational sub-graph computing its derivative. This means that it will return
a {\em computational graph of gradient}, where the output is the derivatives of
all the variable nodes (or a subset of them.) Then, we can use the same facility
to compute the second-order derivatives, right? 

\subsection{What if a Function is {\em not} Differentiable?}
\label{sec:nonlinearities}

From the description so far, one thing we notice is that backpropagation works
only when each and every function node (in a computational graph) is
differentiable. In other words, the nonlinear activation function must be chosen
such that almost everywhere it is differentiable. All three activation functions
I have presented so far have this property.

\paragraph{Logistic Functions}
A sigmoid function is defined as
\begin{align*}
    \sigma(x) = \frac{1}{1+\exp(-x)},
\end{align*}
and its derivative is
\begin{align*}
    \sigma'(x) = \sigma(x) (1 - \sigma(x)).
\end{align*}

A hyperbolic tangent function is
\begin{align*}
    \tanh(x) =  \frac{\exp(2x) -1}{\exp(2x) + 1},
\end{align*}
and its derivative is 
\begin{align*}
    \tanh'(x) = \left(\frac{2}{\exp(x)+\exp(-x)}\right)^2.
\end{align*}

\paragraph{Piece-wise Linear Functions}

I described a rectified linear unit (rectifier or ReLU,
\cite{nair2010rectified,glorot2011deep}) earlier:
\begin{align*}
    \rect(x) = \max(0, x).
\end{align*}
It is clear that this function is not strictly differentiable, because of the
discontinuity at $x=0$. However, the chance of the input to this rectifier lands
exactly at $0$ has zero probability, meaning that we can forget about this
extremely unlikely event. The derivative of the rectifier in this case is
\begin{align*}
    \rect'(x) = \left\{ 
        \begin{array}{l l}
            1, & \text{if }x > 0 \\
            0, & \text{if }x \leq 0
        \end{array}
        \right.
\end{align*}

Although the rectifier has become the most widely used nonlinearity, especially,
in deep learning's applications to computer vision,\footnote{
    Almost all the winning entries in ImageNet Large Scale Visual Recognition
    Challenges (ILSVRC) use a convolutional neural network with rectifiers. See
    \url{http://image-net.org/challenges/LSVRC/}.
}
there is a small issue with the rectifier. That is, for a half of the input
space, the derivative is zero, meaning that the error (the output derivative
from Eq.~\eqref{eq:out_deriv}) will be not well propagated through the rectifier
function node.

In \cite{goodfellow2013maxout}, the rectifier was extended to a maxout unit so
as to avoid this issue of the existence of zero-derivative region in the input
to the rectifier. The maxout unit of rank $k$ is defined as
\begin{align*}
    \maxout(x_1, \ldots, x_k) = \max(x_1, \ldots, x_k),
\end{align*}
and its derivative as
\begin{align*}
    \frac{\partial \maxout}{\partial x_i}(x_1, \dots, x_k) = 
    \left\{
        \begin{array}{l l}
            1, & \text{if } \max(x_1, \ldots, x_k) = x_i \\
            0, & \text{otherwise}
        \end{array}
        \right.
\end{align*}
This means that the derivative is backpropagated only through one of the $k$
inputs.

\paragraph{Stochastic Variables}
These activation functions work well with the backpropagation algorithm, because
they are differentiable almost everywhere in the input space. However, what
happens if a function is non-differentiable at all. One such example is a binary
stochastic node, which is computed by
\begin{enumerate}
    \itemsep 0em
    \item Compute $p = \sigma(x)$, where $x$ is the input to the function node.
    \item Consider $p$ as a mean of a Bernoulli distribution, i.e., $\BB(p)$.
    \item Generate one sample $s \in \left\{ 0, 1\right\}$ from the Bernoulli distribution.
    \item Output $s$.
\end{enumerate}
Clearly there is no derivative of this function node. 

Does it mean that we're doomed in this case? Fortunately, no. Although I will
not discuss about this any further in this course,
Bengio~et~al.~\cite{bengio2013estimating} provide an extensive list of
approaches we can take in order to compute the derivative of the stochastic
function nodes.

\chapter{Recurrent Neural Networks and Gated Recurrent Units}
\label{chap:rnn}

After the last lecture I hope that it has become clear how to build a multilayer
perceptron. Of course, there are so many details that I did not mention, but are
extremely important in practice. For instance, how many layers of simple
transformations Eq.~\eqref{eq:layer} should a multilayer perceptron have for a
certain task? How wide (equiv. $\dim(\phi_0(\vx))$) should each transformation
be? What other transformation layers are there? What kind of learning rate
$\eta$ (see Eq.~\eqref{eq:GD}) should we use? How should we schedule this
learning rate over training? Answers to many of these questions are
unfortunately heavily task-, data- and model-dependent, and I cannot provide any
general answer to them.


\section{Recurrent Neural Networks}
\label{sec:rnn}

Instead, I will move on to describing how we can build a neural
network\footnote{
    Now, let me begin using a term neural network instead of a general function.
} to handle a variable length input. Until now the input $x$ was assumed to be
either a scalar or a vector of the fixed number of dimensions. From here on
however, we remove this assumption of a fixed size input and consider the case
of having a variable length input $x$. 

What do I mean by a {\em variable length input}? A variable length input $x$ is
a {\em sequence} where each input $x$ has a different number of elements. For
instance, the first training example's input $x^1$ may consist of $l^1$ elements
such that
\begin{align*}
    x^1 = (x^1_1, x^1_2, \ldots, x^1_{l^1}).
\end{align*}
Meanwhile, another example's input $x^n$ may be a sequence of $l^n \neq l^1$ elements:
\begin{align*}
    x^n = (x^n_1, x^n_2, \ldots, x^n_{l^n}).
\end{align*}

Let's go back to very basic about dealing with these kinds of sequences.
Furthermore, let us assume that each element $x_i$ is binary, meaning that it is
either $0$ or $1$. What would be the most natural way to write a function that
returns the number of $1$'s in an input sequence $x=(x_1,x_2,\ldots,x_{l})$? My
answer is to first build a recursive function called {\sc add1}, shown in
Alg.~\ref{alg:add1}.  This function {\sc add1} will be called for each element of
the input $x$, as in Alg.~\ref{alg:sum}.

\begin{algorithm}[ht]
\begin{algorithmic}
    \State $s \leftarrow 0$
    \Function{add1}{$v$,$s$}
    \If{$v=0$} 
    \Return{$s$} 
    \Else~
    \Return{$s+1$} 
    \EndIf
    \EndFunction
\end{algorithmic}
\caption{A function {\sc add1}}
\label{alg:add1}
\end{algorithm}

\begin{algorithm}[ht]
\begin{algorithmic}
    \State $s \leftarrow 0$
    \For{$i\leftarrow 1,2,\ldots,l$}
    $s \leftarrow \text{\sc{add1}}(x_i,s)$
    \EndFor
\end{algorithmic}
\caption{A function {\sc add1}}
\label{alg:sum}
\end{algorithm}

There are two important components in this implementation. First, there is a
memory $s$ which counts the number of $1$'s in the input sequence $x$. Second,
a single function {\sc add1} is applied to each symbol in the sequence {\em one
at a time} together with the memory $s$. Thanks to these two properties, our
implementation of the function {\sc add1} can be used with the input sequence of
{\em any length}.

Now let us generalize this idea of having a memory and a recursive function that
works over a variable length sequence. One likely most general case of this idea
is a digital computer we use everyday. A computer program is a sequence $x$ of
instructions $x_i$. A central processing unit (CPU) reads each instruction of
this program and manipulates its registers according to what the instruction
says. Manipulating registers is often equivalent to manipulating any
input--output (I/O) device attached to the CPU. Once one instruction is
executed, the CPU moves on to the next instruction which will be executed with
the content of the registers from the previous step. In other words, these
registers work as a memory in this case ($s$ from Alg.~\ref{alg:sum},) and the
execution of an instruction by the CPU corresponds to a recursive function
({\sc add1} from Alg.~\ref{alg:add1}.)

Both {\sc add1} and CPU are {\em hard coded} in the sense that they do what they
have been designed and manufactured to do. Clearly, this is not what we want,
because nobody knows how to design a CPU or a recursive function for natural
language understanding, which is our ultimate goal. Instead what we want is to
have a parametric recursive function that is able to read a sequence of
(linguistic) symbols and use a memory in order to {\em understand natural
languages}.

To build this parametric recursive function\footnote{
    In neural network research, we call this function a {\em recurrent neural
    network.}
}
that works on a variable-length
input sequence $x=(x_1, x_2, \ldots, x_l)$, we now know that there needs to be a
memory. We will use one vector $\vh \in \RR^{d_h}$ as this memory vector. As is
clear from Alg.~\ref{alg:add1}, this recursive function takes as input both one
input symbol $x_t$ and the memory vector $\vh$, and it returns the updated memory
vector. It often helps to {\em time index} the memory vector as well, such that
the input to this function is $\vh_{t-1}$ (the memory after processing the
previous symbol $x_{t-1}$,) and we use $\vh_t$ to denote the memory vector
returned by the function. This function is then
\begin{align*}
    h_t = f(x_t, \vh_{t-1})
\end{align*}

Now the big question is what kind of parametric form this recursive function $f$
takes? We will follow the simple transformation layer from Eq.~\eqref{eq:layer},
in which case we get
\begin{align}
    \label{eq:rnn_layer}
    f(x_t, \vh_{t-1}) = g(\mW \phi(x_t) + \mU \vh_{t-1}),
\end{align}
where $\phi(x_t)$ is a function that transforms the input symbol (often
discrete) into a $d$-dimensional real-valued vector. $\mW\in \RR^{d_h\times d}$
and $\mU^{d_h \times d_h}$ are parameters of this function. A nonlinear
activation function $g$ can be any function, but for now, we will assume that it
is an element-wise nonlinear function such as $\tanh$.

\subsection{Fixed-Size Output $y$}
\label{sec:rnn_fixed_y}

Because our goal is to approximate an underlying, true function, we now need to
think of how we use this recursive function to {\em return} an output $y$. As
with the case of variable-length sequence input $x$, $y$ can only be either a
fixed-size output, such as a category to which the input $x$ belongs, or a
variable-length sequence output. Here let us discuss the case of having a
fixed-size output $y$.

The most natural approach is to use the last memory vector $\vh_l$ to produce
the output (or more often output distribution.) Consider a task of binary
classification where $y$ is either positive ($1$) or negative ($0$), in which
case a Bernoulli distribution fits perfectly. A Bernoulli distribution is fully
characterized by a single parameter $\mu$. Hence,
\begin{align*}
    \mu = \sigma(\vv^\top \vh_l),
\end{align*}
where $\vv \in \RR^{d_h}$ is a weight vector, and $\sigma$ is a sigmoid
function.

This now looks very much like the multilayer perceptron from Sec.~\ref{sec:mlp}.
The whole function given an input sequence $x$ computes
\begin{align}
    \label{eq:rnn_unrolled}
    \mu = \sigma(\vv^\top 
    \underbrace{
        g(\mW \phi(x_l) + \mU 
        g(\mW \phi(x_{l-1}) + \mU
        g(\mW \phi(x_{l-2}) + \cdots
    g(\mW \phi(x_{1}) + \mU \vh_0) \cdots)))}_{(a) \text{ recurrence}}),
\end{align}
where $\vh_0$ is an initial memory state which can be simply set to an all-zero
vector. 

\begin{figure}[ht]
    \centering
    \includegraphics[width=0.8\textwidth]{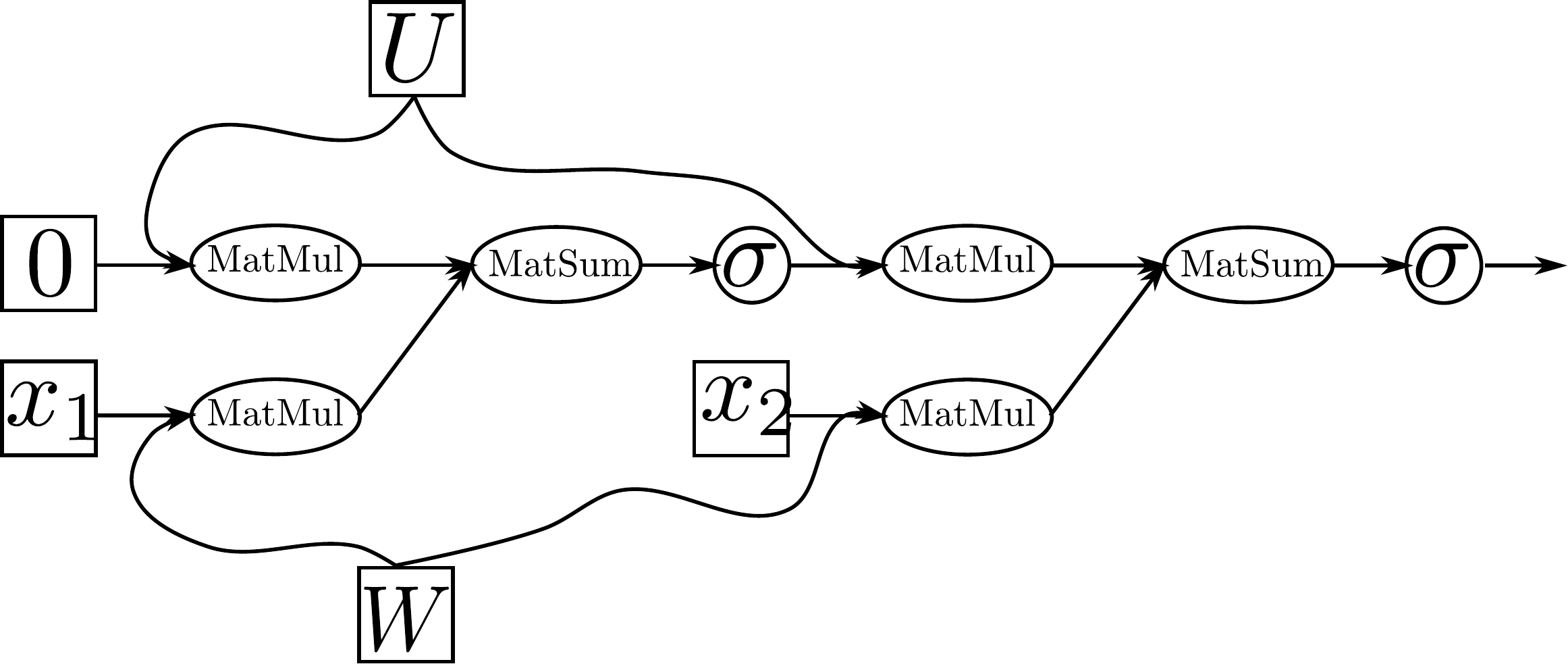}
    \caption{
        Sample computational graph of the recurrence in
        Eq.~\eqref{eq:rnn_unrolled}.
    }
    \label{fig:rec_unrolled}
\end{figure}

The main difference is that the input is not given only to the first simple
transformation layer, but is given to all those transformation layers (one at a
time.) Also, each transformation layer {\em shares} the parameters $\mW$ and
$\mU$.\footnote{
    Note that for brevity, I have omitted bias vectors. This should not matter
    much, as having a bias vector is equivalent to augmenting the input with a
    constant element whose value is fixed at $1$. Why? Because,
    \begin{align*}
        \left[ \mW ; \vb \right]
    \left[ \begin{array}{c} \vx \\1 \end{array} \right]
        = \mW \vx + \vb
    \end{align*}
    Note that as I have declared before all vectors are {\em column} vectors.
} The first two steps of the recurrence part (a) of Eq.~\eqref{eq:rnn_unrolled}
are shown as a computational graph in Fig.~\ref{fig:rec_unrolled}.

As this is not any special computational graph, the whole discussion on how to
automate backpropagation (computing the gradient of the cost function w.r.t. the
parameters) in Sec.~\ref{sec:autodiff} applies to recurrent neural networks
directly, except for one potentially confusing point.

\subsection{Multiple Child Nodes and Derivatives}
\label{sec:rnn_backprop}

It may be confusing how to handle those parameters that are shared across
multiple time steps; $\mW$ and $\mU$ in Fig.~\ref{fig:rec_unrolled}. In fact, in
the earlier section (Sec.~\ref{sec:autodiff}), we did not discuss about what to
do when the output of one node is fed into multiple function nodes.
Mathematically saying, what do we do in the case of 
\begin{align*}
    c = g(f_1(x), f_2(x), \ldots, f_n(x))?
\end{align*}

$g$ can be any function, but let us look at two widely used cases:
\begin{itemize}
    \item Addition: $g(f_1(x), \ldots, f_n(x)) = \sum_{i=1}^n f_i(x)$
        \begin{align*}
            \frac{\partial c}{\partial x} = \frac{\partial c}{\partial g} 
            \sum_{i \in \left\{ 1, 2, \ldots, n\right\}} 
            \frac{\partial f_i}{\partial x}.
        \end{align*}
    \item Multiplication: $g(f_1(x), \ldots, f_n(x)) = \prod_{i=1}^n f_i(x)$
        \begin{align*}
            \frac{\partial c}{\partial x} = \frac{\partial c}{\partial g} 
            \sum_{i \in \left\{ 1, 2, \ldots, n\right\}} 
            \left(\prod_{j \neq i} f_j(x)\right)
            \frac{\partial f_i}{\partial x}.
        \end{align*}
\end{itemize}

From these two cases, we can see that in general
\begin{align*}
    \frac{\partial c}{\partial x} = \frac{\partial c}{\partial g} 
    \sum_{i \in \left\{ 1, 2, \ldots, n\right\}} 
    \frac{\partial g}{\partial f_i}
    \frac{\partial f_i}{\partial x}.
\end{align*}
This means that when multiple derivatives are {\em backpropagated} into a single
node, the node should first {\em sum them} and multiply its summed derivative
with its own derivative.

What does this mean for the shared parameters of the recurrent neural network?
In an equation,
\begin{align}
    \label{eq:bptt}
    \frac{\partial C}{\partial W} =&
    \underbrace{
        \frac{\partial C}{\partial \MatSum^l}
    }_{(a)}
    \frac{\partial \MatSum^l}{\partial \MatMul^l}
    \frac{\partial \MatMul^l}{\partial W} \\
    +& 
    \underbrace{
        \frac{\partial C}{\partial \MatSum^l}
    }_{(a)}
    \underbrace{
        \frac{\partial \MatSum^l}{\partial \MatSum^{l-1}}
    }_{(b)}
    \frac{\partial \MatSum^{l-1}}{\partial \MatMul^{l-1}}
    \frac{\partial \MatMul^{l-1}}{\partial W} 
    \nonumber \\
    +& 
    \underbrace{
        \frac{\partial C}{\partial \MatSum^l}
    }_{(a)}
    \underbrace{
        \frac{\partial \MatSum^l}{\partial \MatSum^{l-1}}
    }_{(b)}
    \underbrace{
        \frac{\partial \MatSum^{l-1}}{\partial \MatSum^{l-2}}
    }_{(c)}
    \frac{\partial \MatSum^{l-2}}{\partial \MatMul^{l-2}}
    \frac{\partial \MatMul^{l-2}}{\partial W} 
    \nonumber \\
    +& 
    \cdots,
    \nonumber
\end{align}
where the superscript $l$ of each function node denotes the layer at which the
function node resides. 

Similarly to what we've observed in Sec.~\ref{sec:autodiff}, many derivatives
are shared across the terms inside the summation in Eq.~\eqref{eq:bptt}. This
allows us to compute the derivative of the cost function w.r.t. the
parameter $W$ efficiently by simply running the recurrent neural network
backward. 

\subsection{Example: Sentiment Analysis}

There is a task in natural language processing called {\em sentiment analysis}.
As the name suggests, the goal of this task is to predict the sentiment of a
given text. This is definitely one function that a human can do fairly well:
when you read a critique's review of a movie, you can easily tell whether the
critique likes, hates or is neutral to the movie. Also, even without a star
rating of a product on Amazon, you can quite easily tell whether a user like it
by reading her/his review of the product.

In this task, an input sequence $x$ is a given text, and the fixed-size output
is its label which is almost always one of positive, negative or neutral. Let us
assume for now that the input is a {\em sequence of words}, where each word
$\vx_i$ is represented as a so-called one-hot vector.\footnote{
    A one-hot vector is a way to represent a discrete symbol as a binary vector.
    The one-hot vector $\vv_i$ of a symbol $i \in V = \left\{ 1, 2, \ldots, |V|
    \right\}$ is 
    \begin{align*}
        \vv_i = [ \underbrace{0, \ldots, 0}_{1,\ldots,i-1}, 
        \underbrace{1}_{i}, \underbrace{0, \ldots, 0}_{i+1, \ldots,
    |V|}]^\top.
    \end{align*}
} In this case, we can use 
\begin{align*}
    \phi(x_t) = \vx_t
\end{align*}
in Eq.~\eqref{eq:rnn_layer}.

Once the input sequence, or paragraph in this specific example, is read, we get
the last memory state $\vh_l$ of the recurrent neural network. We will
affine-transform $\vh_l$ followed by the {\em softmax} function to obtain the
conditional distribution of the output $y \in \left\{ 1, 2, 3 \right\}$ ($1$:
positive, $2$: neutral and $3$: negative):
\begin{align}
    \label{eq:multiclass}
    \mu = \left[ \mu_1, \mu_2, \mu_3 \right]^\top = \softmax(\mV \vh_l),
\end{align}
where $\mu_1$, $\mu_2$ and $\mu_3$ are the probabilities of ``positive'',
``neural'' and ``negative''. See Eq.~\eqref{eq:softmax} for more details on the
softmax function.

Because this network returns a categorial distribution, it is natural to use the
(categorical) cross entropy as the cost function. See
Eq.~\eqref{eq:cat_crossentropy}. A working example of this sentiment analyzer
based on recurrent neural networks will be introduced and discussed during the
lab session.\footnote{
    For those eager to learn more, see
    \url{http://deeplearning.net/tutorial/lstm.html} in advance of the lab
    session.
}

\subsection{Variable-Length Output $y$: $|x|=|y|$}
\label{sec:rnn_x_y}

Let's generalize what we have discussed so far to recurrent neural networks
here. Instead of a fixed-size output $y$, we will assume that the goal is to
label each input symbol, resulting in the output sequence $y=(y_1, y_2, \ldots,
y_l)$ of the same length as the input sequence $x$.

What kind of applications can you think of that returns the output sequence as
long as the input sequence? One of the most widely studied problems in natural
language processing is a problem of classifying each word in a sentence into one
of part-of-speech tags, often called POS tagging (see Sec.~3.1 of
\cite{manning1999foundations}.) Unfortunately, in my personal opinion, this is
perhaps the least interesting problem of all time in natural language
understanding, but perhaps the most well suited problem for this section.

In its simplest form, we can view this problem of POS tagging as classifying
each word in a sentence as one of {\em noun}, {\em verb}, {\em adjective} and
{\em others}. As an example, given the following input sentence $x$
\begin{align*}
    x = (\text{Children}, \text{eat}, \text{sweet}, \text{candy}),
\end{align*}
the goal is to output
\begin{align*}
    y = (\text{noun}, \text{verb}, \text{adjective}, \text{noun}).
\end{align*}

This task can be solved by a recurrent neural network from the preceding section
(Sec.~\ref{sec:rnn_fixed_y}) after a quite trivial modification. Instead of
waiting until the end of the sentence to get the last memory state of the
recurrent neural network, we will use the {\em immediate memory state} to
predict the label at each time step $t$. 

At each time $t$, we get the immediate memory state $\vh_t$ by
\begin{align}
    \label{eq:rnn_h_x}
    \vh_t = f(x_t, \vh_{t-1}),
\end{align}
where $f$ is from Eq.~\eqref{eq:rnn_layer}. Instead of continuing on to
processing the next word, we will first predict the label of the $t$-th input
word $x_t$.

This can be done by 
\begin{align}
    \label{eq:rnn_y_h}
    \mu_t = \left[ \mu_{t,1}, \mu_{t,2}, \mu_{t,3}, \mu_{t,4} \right]^\top =
    \softmax(\mV \vh_{t}).
\end{align}
Four $\mu_{t,i}$'s correspond to the probabilities of the four categories; (1)
noun, (2) verb, (3) adjective and (4) others.

From this output distribution at time step $t$, we can define a {\em per-step},
{\em per-sample} cost function:
\begin{align}
    \label{eq:cost_per_step}
    C_{x,t}(\TT) = -\sum_{k=1}^K \II_{k=y} \mu_{t,k},
\end{align}
where $K$ is the number of categories, four in this case.  We discussed earlier
in Eq.~\eqref{eq:cat_crossentropy}.
Naturally a per-sample cost function is defined as the sum of these per-step,
per-sample cost functions:
\begin{align}
    \label{eq:cost_per_sample}
    C_x(\TT) = -\sum_{t=1}^l \sum_{k=1}^K \II_{k=y} \mu_{t,k}.
\end{align}

\paragraph{Incorporating the Output Structures} 
This formulation of the cost function is equivalent to maximizing the
log-probability of the correct output sequence given an input sequence, where
the conditional log-probability is defined as
\begin{align}
    \label{eq:log_y_x_seq}
    \log p(y|x) = \underbrace{\sum_{t=1}^l \underbrace{\log
    p(y_t|x_1,\ldots,x_t)}_{\text{Eq.~\eqref{eq:cost_per_step}}}}_{
        \text{Eq.~\eqref{eq:cost_per_sample}}}.
\end{align}
This means that the network is predicting the label of the $t$-th input symbol
using only the input symbols read up to that point (i.e., $x_1, x_2, \ldots,
x_t$.)

In other words, this means that the recurrent neural network is {\em not} taking
into account the structure of the output sequence. For instance, even without
looking at the input sequence, in English it is well known that the probability
of the next word being a noun increases if the current word is an
adjective.\footnote{
    Okay, this requires a more thorough analysis, but for the sake of the
    argument, which does not have to do anything with actual POS tags, let's
    believe that this is indeed the case.
} This kind of structures in the output are effectively ignored in this
formulation.

Why is this so in this formulation? Because, we have made an assumption that the
output symbols $y_1, y_2, \ldots, y_l$ are mutually independent conditioned on
the input sequence. This is clear from Eq.~\eqref{eq:log_y_x_seq} and the
definition of the conditional independence:
\begin{align*}
    &\text{$Y_1$ and $Y_2$ are conditionally independent dependent on $X$} \\
    \iff&
    p(Y_1, Y_2|X) = p(Y_1|X)p(Y_2|x).
\end{align*}

If the underlying, true conditional distribution obeyed this assumption of
conditional independence, there is no worry. However, this is a very strong
assumption for many of the tasks we run into, apparently from the example of POS
tagging. Then, how can we exploit the structure in the output sequence?

One simple way is to make a less strong assumption about the conditional
probability of the output sequence $y$ given $x$. For instance, we can assume
that
\begin{align*}
    \log p(y|x) = \sum_{i=1}^l \log p(y_i|y_{<i}, x_{\leq i}),
\end{align*}
where $y_{<i}$ and $x_{\leq i}$ denote all the output symbols before the $i$-th one 
and all the input symbols up to the $i$-th one, respectively.

Now the question is how we can incorporate this into the existing formulation of
a recurrent neural network from Eq.~\eqref{eq:rnn_h_x}. It turned out that the
answer is extremely simple. All we need to do is to compute the memory state of
the recurrent neural network based not only on the current input symbol $x_t$
and the previous memory state $\vh_{t-1}$, but also on the previous output
symbol $y_{t-1}$ such that
\begin{align*}
    \vh_t = f(x_t, y_{t-1}, \vh_{t-1}).
\end{align*}
Similarly to Eq.~\eqref{eq:rnn_layer}, we can think of implementing $f$ as
\begin{align*}
    f(x_t, y_{t-1}, \vh_{t-1}) = g(\mW_x \phi_x(x_t) + \mW_y \phi_y(y_{t-1}) +
    \mW_h \vh_{t-1}).
\end{align*}

There are two questions naturally arising from this formulation. First, what do
we do when computing $\vh_1$? This is equivalent to saying what $\phi_y(y_{0})$
is. There are two potential answers to this question:
\begin{enumerate}
    \itemsep 0em
    \item Fix $\phi_y(y_0)$ to an all-zero vector
    \item Consider $\phi_y(y_0)$ as an additional parameter
\end{enumerate}
In the latter case, $\phi_y(y_0)$ will be estimated together with all the other
parameters such as those weight matrices $\mW_x$, $\mW_y$, $\mW_h$ and $\mV$.

\paragraph{Inference}
The second question involves how to handle $y_{t-1}$. During training, it is
quite straightforward, as our cost function (KL-divergence between the
underlying, true distribution and the parametric conditional distribution
$p(y|x)$, approximated by Monte Carlo method) says that we use the groundtruth
value for $y_{t-1}$'s. 

It is however not clear what we should do when we test the trained network,
because then we are not given the groundtruth output sequence. This process of
finding an output that maximizes the conditional (log-)probability is called
{\em inference}\footnote{
    Okay, I confess. The term {\em inference} refers to a much larger class of
    problems, even if we consider only machine learning. However, let me simply
    use this term to refer to a task of finding the most likely output of a
    function.
}:
\begin{align*}
    \hat{y} = \argmax_y \log p(y|x)
\end{align*}

The exact inference is quite straightforward. One can simply evaluate $\log
p(y|x)$ for every possible output sequence and choose the one with the highest
conditional probability. Unfortunately, this is almost always intractable, as
the number of every possible output sequence grows exponentially with respect to
the length of the sequence:
\begin{align*}
    |\YY| = K^l,
\end{align*}
where $\YY$, $K$ and $l$ are the set of all possible output sequences, the
number of labels and the length of the sequence, respectively. Thus, this is
necessary to resort to approximate search over the set $\YY$.

The most naive approach to approximate inference is a greedy one. With the
trained model, you predict the first output symbol $\hat{y}_1$ based on the
first input symbol $x_1$ by selecting the category of the highest probability
$p(y_1|x_1)$. Now, given $\hat{y}_1$, $x_1$ and $x_2$, we compute
$p(y_2|x_1,x_2,y_1)$ from which we select the next output symbol $\hat{y}_2$
with the highest probability. We continue this process iteratively until the
last output symbol $\hat{y}_l$ is selected.

This is greedy in the sense that any early choice with a high conditional
probability may turn out to be unlikely one due to extremely low conditional
probabilities later on. It is highly related to the so-called {\em garden path
sentence} problem. To know more about this, read, for instance, Sec.~3.2.4 of
\cite{manning1999foundations}.

It is possible to alleviate this issue by considering $N<K$ best hypotheses of
the output sequence at each time step. This procedure is called {\em beam
search}, and we will discuss more about this in a later lecture on neural
machine translation.

\section{Gated Recurrent Units}

\subsection{Making Simple Recurrent Neural Networks {\em Realistic}}

Let us get back to the analogy we made in Sec.~\ref{sec:rnn}. We compared a
recurrent neural network to how CPU works. Executing a recurrent function $f$ is
equivalent to executing one of the instructions on CPU, and the memory state of
the recurrent neural network is equivalent to the registers of the CPU. This
analogy does sound plausible, except that it is not.

In fact, how a simple recurrent neural network works is far from being similar
to how CPU works. I am now talking about how they are implemented in practice,
but rather I'm talking at the conceptual level. What is it at the conceptual
level that makes the simple recurrent neural network unrealistic?

An important observation we make about the simple recurrent neural network is
that it {\em refreshes} the whole memory state at each time step. This is almost
opposite to how the registers on a CPU are maintained. Each time an instruction
is executed, the CPU does not clear up the whole registers and repopulate them.
Rather, it works only on a small number of registers. All the other registers'
values are stored as they were before the execution of the instruction.

Let's try to write this procedure mathematically. Each time, based on the choice
of instruction to be executed, a subset of the registers of a CPU, or a subset
of the elements in the memory state of a recurrent neural network, is selected.
This can be written down as a binary vector $\vu \in \left\{ 0, 1\right\}^{n_h}$:
\begin{align*}
    u_i = \left\{ 
        \begin{array}{l l}
            0, & \text{ if the register's value does not change} \\
            1, & \text{ if the register's value will change}
        \end{array}
        \right.
\end{align*}

With this binary vector, which I will call an {\em update gate}, a new memory
state or a new register value at time $t$ can be computed as a convex
interpolation such that
\begin{align}
    \label{eq:gru_leaky}
    \vh_{t} = (1 - \vu) \odot \vh_{t-1} + \vu \odot \tilde{\vh}_t,
\end{align}
where $\odot$ is as usual an element-wise multiplication. $\tilde{\vh}_t$
denotes a new memory state or a new register value, after executing the
instruction at time $t$.

Another unrealistic point about the simple recurrent neural network is that each
execution considers the whole registers. It is almost impossible to imagine
designing an instruction on a CPU that requires to read the values of all the
registers. Instead, what almost always happens is that each instruction will
consider only a small subset of the registers, which again we can use a binary
vector to represent. Let me call it a {\em reset gate} $\vr \in \left\{ 0,
1\right\}^{n_h}$:
\begin{align*}
    r_i = \left\{ 
        \begin{array}{l l}
            0, & \text{ if the register's value will not be used} \\
            1, & \text{ if the register's value will be used}
        \end{array}
        \right.
\end{align*}

This reset gate can be multiplied to the register values {\em before} being used
by the instruction at time $t$.\footnote{
    It is important to note that this is {\em not} resetting the actual values
    of the registers, but only the input to the instruction/recursive function.
} If we use a recursive function $f$ from Eq.~\eqref{eq:rnn_layer}, it means that
\begin{align} 
    \label{eq:gru_cand}
    \tilde{\vh}_t = f(x_t, \vr \odot \vh_{t-1}) = g(\mW \phi(x_t) + \mU (\vr
    \odot \vh_{t-1})).
\end{align}

\begin{figure}[ht]
    \centering
    \begin{minipage}{0.50\textwidth}
        \centering
        \includegraphics[width=0.9\columnwidth]{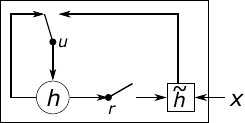}
    \end{minipage}
    \hfill
    \begin{minipage}{0.47\textwidth}
        \caption{
            A graphical illustration of a gated recurrent
            unit~\cite{Cho-et-al-EMNLP2014}.
        }
        \label{fig:gru}
    \end{minipage}
\end{figure}

Now, let us put these two gates that are necessary to make the simple recurrent
neural network more realistic into one piece. At each time step, the {\em
candidate} memory state is computed based on a subset of the elements of the
previous memory state:
\begin{align*}
    \tilde{\vh}_t = g(\mW \phi(x_t) + \mU (\vr \odot \vh_{t-1}))
\end{align*}
A new memory state is computed as a linear interpolation between the previous
memory state and this candidate memory state using the update gate:
\begin{align*}
    \vh_t = (1 - \vu) \odot \vh_{t-1} + \vu \odot \tilde{\vh}_t
\end{align*}
See Fig.~\ref{fig:gru} for the graphical illustration.

\subsection{Gated Recurrent Units}
\label{sec:gru}

Now here goes a big question: {\em How are the update $\vu$ and reset $\vr$
gates computed?}

If we stick to our analogy to the CPU, those gates must be pre-configured {\em
per instruction}. Those binary gates are dependent on the instruction. Again
however, this is not what we want to do in our case. There is no set of
predefined instructions, but the execution of any instruction corresponds to
computing a recurrent {\em function} based on the input symbol and the memory
state from the previous time step (see, e.g., Eq.~\eqref{eq:rnn_layer}.)
Similarly to this what we want with the update and reset gates is that they are
computed by a function which depends on the input symbol and the previous memory
state.

This sounds like quite straightforward, except that we defined the gates to be
binary. This means that whatever the function we use to compute those gates, the
function will be a discontinuous function with zero derivative almost
everywhere, except at the point where a sharp transition from $0$ to $1$
happens. We discussed the consequence of having an activation function with zero
derivative almost everywhere in Sec.~\ref{sec:nonlinearities}, and the
conclusion was that it becomes very difficult to compute the gradient of the
cost function efficiently {\em and} exactly with these discrete activation
functions in a computational graph.

One simple solution which turned out to be extremely efficient is to consider
those gates not as binary vectors but as real-valued coefficient vectors. In
other words, we redefine the update and reset gates to be
\begin{align*}
    \vu \in \left[ 0, 1 \right]^{n_h}, \vr \in \left[ 0, 1\right]^{n_h}.
\end{align*}
This approach makes these gates {\em leaky} in the sense that they always allow
some leak of information through the gate.

In the case of the reset gate, rather than making a hard decision on which
subset of the registers, or the elements of the memory state, will be used, it
now decides {\em how much} information from the previous memory state will be
used. The update gate on the other hand now controls how much content in the
memory state will be replaced, which is equivalent to saying that it controls
how much information will be {\em kept} from the previous memory state.

Under this definition we can simply use a sigmoid function from
Eq.~\eqref{eq:sigmoid} to compute these gates:
\begin{align*}
    \vr =& \sigmoid(\mW_r \phi(x_t) + \mU_r \vh_{t-1}), \\
    \vu =& \sigmoid(\mW_u \phi(x_t) + \mU_u (\vr \odot \vh_{t-1})), \\
\end{align*}
where $\mW_r$, $\mU_r$, $\mW_u$ and $\mU_u$ are the additional
parameters.\footnote{
    Note that this is not {\em the} formulation available for computing the
    reset and update gates. For instance, one can use the following definitions
    of the reset and update gates:
    \begin{align*}
        \vr =& \sigmoid(\mW_r \phi(x_t) + \mU_r \vh_{t-1}), \\
        \vu =& \sigmoid(\mW_u \phi(x_t) + \mU_u \vh_{t-1}), \\
    \end{align*}
    which is more parallelizable than the original formulation from
    \cite{Cho-et-al-EMNLP2014}. This is because there is no more direct
    dependency between $\vr$ and $\vu$, which makes it possible to compute them
    in parallel.
}
Since
the sigmoid function is differentiable everywhere, we can use the
backpropagation algorithm (see Sec.~\ref{sec:autodiff}) to compute the
derivatives of the cost function with respect to these parameters and estimate
them together with all the other parameters.

We call this recurrent activation function with the reset and update gates a
{\em gated recurrent unit} (GRU), and a recurrent neural network having this GRU
as a gated recurrent network.

\subsection{Long Short-Term Memory}
\label{sec:lstm}

The gated recurrent unit (GRU) is highly motivated by a much earlier work on
long short-term memory (LSTM) units \cite{hochreiter1997long}.\footnote{
    Okay, let me confess here. I was not well aware of long short-term memory
    when I was designing the gated recurrent unit together with Yoshua Bengio
    and Caglar Gulcehre in 2014.
} The LSTM was proposed in 1997 with the goal of building a recurrent neural
network that can learn long-term dependencies across many number of timsteps,
which was deemed to be difficult to do so with a simple recurrent neural
network. 

Unlike the element-wise nonlinearity of the simple recurrent neural network and
the gated recurrent unit, the LSTM explicitly separates the memory state $\vc_t$
and the output $\vh_t$. The output is a small subset of the {\em hidden} memory
state, and only this subset of the memory state is visibly {\em exposed} to any
other part of the whole network. 

How does a recurrent neural network with LSTM units decide how much of the
memory state it will reveal?  As perhaps obvious at this point, the LSTM uses a
so-called {\em output} gate $\vo$ to achieve this goal. Similarly to the reset
and update gates of the GRU, the output gate is computed by
\begin{align*}
    \vo = \sigmoid(\mW_o \phi(x_t) + \mU_o \vh_{t-1}).
\end{align*}
This output vector is multiplied to the memory state $\vc_t$ point-wise to
result in the output:
\begin{align*}
    \vh_t = \vo \odot \tanh(\vc_t).
\end{align*}

Updating the memory state $\vc_t$ closely resembles how it is updated in the GRU
(see Eq.~\eqref{eq:gru_leaky}.) A major difference is that instead of using a
single update gate, the LSTM uses two gates, forget and input gates, such that
\begin{align*}
    \vc_t = \vf \odot \vc_{t-1} + \vi \odot \tilde{\vc}_t,
\end{align*}
where $\vf \in \RR^{n_h}$, $\vi \in \RR^{n_h}$ and $\tilde{\vc}_t$ are the
forget gate, input gate and the candidate memory state, respectively.

The roles of those two gates are quite clear from their names. The forget gate
decides how much information from the memory state will be {\em forgotten},
while the input gate controls how much informationa about the new input
(consisting of the input symbol and the previous output) will be {\em inputted}
to the memory. They are computed by
\begin{align}
    \label{eq:lstm_forget}
    \vf =& \sigmoid(\mW_f \phi(x_t) + \mU_f \vh_{t-1}), 
    \\
    \vi =& \sigmoid(\mW_i \phi(x_t) + \mU_i \vh_{t-1}).
    \nonumber
\end{align}

The candidate memory state is computed similarly to how it was done with the GRU
in Eq.~\eqref{eq:gru_cand}:
\begin{align}
    \tilde{\vc}_t = g(\mW_c \phi(x_t) + \mU_c \vh_{t-1}),
\end{align}
where $g$ is often an element-wise $\tanh$.

All the additional parameters specific to the
LSTM--$\mW_o,\mU_o,\mW_f,\mU_f,\mW_i,\mU_i,\mW_c$ and $\mU_c$-- are estimated
together with all the other parameters. Again, every function inside the LSTM is
differentiable everywhere, and we can use the backpropagation algorithm to
efficient compute the gradient of the cost function with respect to all the
parameters.

Although I have described one formulation of the long short-term memory unit
here, there are many other variants proposed over more than a decade since it
was first proposed. For instance, the forget gate in Eq.~\eqref{eq:lstm_forget}
was not present in the original work \cite{hochreiter1997long} but was fixed to
$1$. Gers et al. \cite{gers2000learning} proposed the forget gate few years
after the LSTM was originally proposed, and it turned out to be one of the most
crucial component in the LSTM. For more variants of the LSTM, I suggest you to
read \cite{greff2015lstm,jozefowicz2015empirical}.\footnote{
    Interestingly, based on the observation in \cite{jozefowicz2015empirical},
    it seems like the plain LSTM with a forget gate and the GRU seem to be close
    to the optimal gated unit we can find. 
}

\section{Why not Rectifiers?}

\subsection{Rectifiers Explode}
\label{sec:rect_explode}

Let us go back to the {\em simple recurrent neural network} which uses the
simple transformation layer from Eq.~\eqref{eq:rnn_layer}:
\begin{align*}
    f(x_t, \vh_{t-1}) = g(\mW \phi(x_t) + \mU \vh_{t-1}),
\end{align*}
where $g$ is an element-wise nonlinearity. 

One of the most widely used nonlinearities is a hyperbolic tangent function
$\tanh$. This is unlike the case in feedforward neural networks (multilayer
perceptrons) where a (unbounded) piecewise linear function, such as a rectifier
and maxout, has become standard. In the case of feedforward neural networks, you
can safely assume that everyone uses some kind of piecewise linear function as
an activation function in the network. This has become pretty much standard
since Krizhevsky et al. \cite{krizhevsky2012imagenet} shocked the (computer
vision) research community by outperforming all the more traditional computer
vision teams in the ImageNet Large Scale Visual Recognition Challenge
2012.\footnote{
    \url{http://image-net.org/challenges/LSVRC/2012/results.html}
}

The main difference between logistic functions
($\tanh$ and sigmoid function) and piecewise linear functions (rectifiers and
maxout) is that the former is bounded from both above and below, while the
latter is bounded only from below (or in some cases, not bounded at all
\cite{he2015delving}.\footnote{
    A parametric rectifier, or PReLU, is defined as
    \begin{align*}
        g(x) = \left\{ 
            \begin{array}{l l}
                x, & \text{if }x \geq 0 \\
                ax, & \text{otherwise}
            \end{array}
            \right.,
    \end{align*}
    where $a$ is a parameter to be estimated together with all the other
    parameters of a network.
})

This unbounded nature of piece-wise linear functions makes it difficult for them
to be used in recurrent neural networks. Why is this so?

Let us consider the simplest case of unbounded element-wise nonlinearity; a
linear function:
\begin{align*}
    g(a) = a.
\end{align*}

The hidden state after $l$ symbols is
\begin{align}
    \vh_l =& \mU(\mU (\mU (\mU (\cdots) + \mW\phi(x_{l-3})) + \mW\phi(x_{l-2})) +
    \mW\phi(x_{l-1})) + \mW\phi(x_l) 
    \nonumber
    \\
    =& \left(\prod_{l'=1}^{l-1} \mU\right) \mW\phi(x_{1})
    + \left(\prod_{l'=1}^{l-2} \mU\right) \mW\phi(x_{2})
    + \cdots 
    + \mU \mW \phi(x_{l-1}) 
    + \mW \phi(x_l), 
    \nonumber
    \\
    \label{eq:rnn_forward}
    =& \sum_{t=1}^l \underbrace{\left( \prod_{l'=1}^{l-t} \mU\right) \mW\phi(x_t)}_{(a)}
\end{align}
where $l$ is the length of the input sequence.

Let us assume that
\begin{itemize}
    \itemsep 0em
    \item $\mU$ is a full rank matrix
    \item The input sequence is sparse: $\sum_{t=1}^l \II_{\phi(x_t) \neq 0} =
        c$, where $c=O(1)$
    \item $\left[ \mW \phi(x) \right]_i > 0$ for all $i$
\end{itemize}
and consider Eq.~\eqref{eq:rnn_forward} (a):
\begin{align}
    \label{eq:rnn_forward_one}
    \vh_l^{t'} = \left( \prod_{l'=1}^{l-t'} \mU\right) 
    \mW\phi(x_{t'}).
\end{align}

Now, let's look at what happens to Eq.~\eqref{eq:rnn_forward_one}. First, the
eigendecomposition of the matrix $\mU$:
\begin{align*}
    \mU = \mQ \mS \mQ^{-1},
\end{align*}
where $\mS$ is a diagonal matrix whose non-zero entries are eigenvalues.
$\mQ$ is an orthogonal matrix. Then
\begin{align*}
    \prod_{l'=1}^{l-t'} \mU = \mQ \mS^{l-t'} \mQ^{-1},
\end{align*}
and
\begin{align*}
    \left(\prod_{l'=1}^{l-t'} \mU\right) \mW\phi(x_{t'})
    = \diag(\mS^{l-t'}) \odot (\underbrace{\mQ \mQ^{-1}}_{=\mI} \mW\phi(x_{t'})),
\end{align*}
where $\odot$ is an element-wise product.

What happens if the largest eigenvalue $e_{\max} = \max \diag(\mS)$ is larger
than $1$, the norm of $\vh_l$ will {\em explode}, i.e., $\|\vh_l\| \to \infty$.
Furthermore, due to the assumption that $\mW\phi(x_{t'}) > 0$, each element of
$\vh_l$ will explode to infinity as well. The rate of growth is exponentially
with respect to  the length of the input sequence, meaning that even when the
input sequence is not too long, the norm of the memory state grows quickly if
$e_{\max}$ is reasonably larger than $1$. 

This happens, because the nonlinearity $g$ is {\em unbounded}. If $g$ is bounded
from both above and below, such as the case with $\tanh$, the norm of the memory
state is also bounded. In the case of $\tanh:\RR \to \left[ -1, 1\right]$,
\begin{align*}
    \|\vh_l\|\leq \dim(\vh_l).
\end{align*}

This is one reason why a logistic function, such as $\tanh$ and $\sigmoid$, is
most widely used with recurrent neural networks, compared to piecewise linear
functions.\footnote{
    However, it is not to say that piecewise linear functions are never used for
    recurrent neural networks. See, for instance,
    \cite{le2015simple,bengio2013advances}.
} I will call this recurrent neural network with $\tanh$ as an element-wise
nonlinear function a {\em simple recurrent neural network}. 

\subsection{Is $\tanh$ a Blessing?}
\label{sec:vanishing_grad}

Now, the argument in the previous section may sound like $\tanh$ and $\sigmoid$
are {\em the} nonlinear functions that one should use. This seems quite
convincing for recurrent neural networks, and perhaps so for feedforward neural
networks as well, if the network is {\em deep} enough.

Here let me try to convince you otherwise by looking at how the norm of
backpropagated derivative behaves. Again, this is much easier to see if we
assume the following:
\begin{itemize}
    \itemsep 0em
    \item $\mU$ is a full rank matrix
    \item The input sequence is sparse: $\sum_{t=1}^l \II_{\phi(x_t) \neq 0} =
        c$, where $c=O(1)$
\end{itemize}

Similarly to Eq.~\eqref{eq:rnn_forward}, let us consider a forward computational
path until $\vh_l$, however without assuming a linear activation function:
\begin{align*}
    \vh_l = g\left(\mU g\left(\mU g\left(\mU g\left(\mU \left( \cdots \right)
                + \mW \phi\left(x_{l-3}\right)\right) + \mW
        \phi\left(x_{l-2}\right)\right)+\mW \phi\left(x_{l-1}\right)\right) +
    \mW\phi\left(x_{l}\right)\right).
\end{align*}
We will consider a subsequence of this process, in which all the input symbols
are $0$ except for the first symbol:
\begin{align*}
    \vh_{l_1} = g\left(\mU g\left(\mU \left( \cdots g\left(\mU \vh_{l_0} + \mW
    \phi\left(x_{l_0+1}\right)\right)\right)\right)\right).
\end{align*}
It should be noted that as $l$ approaches
infinity, there will be at least one such subsequence whose length also
approaches infinity due to the sparsity of the input we assumed.

From this equation, let's look at 
\begin{align*}
    \frac{\partial \vh_{l_1}}{\partial \phi\left(x_{l_0+1}\right)}.
\end{align*}
This measures the effect of the $(l_0+1)$-th input symbol $x_{l_0+1}$ on the
$l_1$-th memory state of the simple recurrent neural network. This is also the
crucial derivative that needs to be computed in order to compute the gradient of
the cost function using the automated backpropagation procedure described in
Sec.~\ref{sec:autodiff}.

This derivative can be rewritten as
\begin{align*}
    \frac{\partial \vh_{l_1}}{\partial \phi\left(x_{l_0+1}\right)} = 
    \underbrace{\frac{\partial \vh_{l_1}}{\partial
    \vh_{l_0+1}}}_{(a)}
    \frac{\partial \vh_{l_0+1}}{\partial \underline{\vh}_{l_0+1}}
    \frac{\partial \underline{\vh}_{l_0+1}}{\partial
    \phi\left(x_{l_0+1}\right)}.
\end{align*}
Among these three terms in the left hand side, we will focus on the first one
(a) which can be further expanded as
\begin{align}
    \label{eq:tderiv}
    \frac{\partial \vh_{l_1}}{\partial \vh_{l_0+1}} = 
    \left(
    \underbrace{\frac{\partial \vh_{l_1}}{\partial \underline{\vh}_{l_1}}}_{(b)}
    \underbrace{\frac{\partial \underline{\vh}_{l_1}}{\partial \vh_{l_1-1}}}_{(c)}
\right)
\left(
    \underbrace{\frac{\partial \vh_{l_1-1}}{\partial
    \underline{\vh}_{l_1-1}}}_{(b)}
        \underbrace{\frac{\partial \underline{\vh}_{l_1-1}}{\partial
        \vh_{l_1-2}}}_{(c)}
\right)
    \cdots
\left(
    \underbrace{\frac{\partial \vh_{l_0+2}}{\partial
    \underline{\vh}_{l_0+2}}}_{(b)}
        \underbrace{\frac{\partial \underline{\vh}_{l_0+2}}{\partial
        \vh_{l_0+1}}}_{(c)}
    \right).
\end{align}

Because this is a recurrent neural network, we can see that the analytical forms
for the terms grouped by the parentheses in the above equation are identical
except for the subscripts indicating the time index. In other words, we can
simply only on one of those groups, and the resulting analytical form will be
generally applicable to all the other groups.

First, we look at Eq.~\eqref{eq:tderiv} (b), which is nothing but a derivative
of a nonlinear activation function used in this simple recurrent neural network.
The derivatives of the widely used logistic functions are 
\begin{align*}
    \sigma'(x) =& \sigma(x) (1 - \sigma(x)), \\
    \tanh'(x) =& 1 - \tanh^2(x),
\end{align*}
as described earlier in Sec.~\ref{sec:nonlinearities}. 
Both of these functions' derivatives are {\em bounded}:
\begin{align}
    \label{eq:sigmoid_bound}
    &0 < \sigma'(x) \leq 0.25, \\
    \label{eq:tanh_bound}
    &0 < \tanh'(x) \leq 1.
\end{align}

In the simplest case in which $g$ is a linear function (i.e., $x=g(x)$,) we do
not even need to look at $\left\| \frac{\partial \vh_{t}}{\partial
\underline{\vh}_{t}} \right\|$. We simply ignore all the $\frac{\partial
\vh_t}{\partial \underline{\vh}_t}$ from Eq.~\eqref{eq:tderiv}.

Next, consider Eq.~\eqref{eq:tderiv} (c). In this case of simple recurrent
neural network, we notice that we have already learned how to compute this
derivative earlier in Sec.~\ref{sec:example2}:
\begin{align*}
    \frac{\partial \underline{\vh}_{t+1}}{\partial \vh_{t}} = \mU
\end{align*}

From these two, we get
\begin{align*}
    \frac{\partial \vh_{l_1}}{\partial \vh_{l_0+1}} =&
    \left(
    \frac{\partial \vh_{l_1}}{\partial \underline{\vh}_{l_1}}
    \mU
\right)
\left(
    \frac{\partial \vh_{l_1-1}}{\partial \underline{\vh}_{l_1-1}}
    \mU
\right)
    \cdots
\left(
    \frac{\partial \vh_{l_0+2}}{\partial \underline{\vh}_{l_0+2}}
    \mU
    \right) 
    \\
    =& \prod_{t=l_0+2}^{l_1} \left( \frac{\partial \vh_{t}}{\partial
\underline{\vh}_{t}} \mU\right).
\end{align*}

Do you see how similar it looks like Eq.~\eqref{eq:rnn_forward_one}? If the recurrent
activation function $f$ is linear, this whole term reduces to
\begin{align*}
    \frac{\partial \vh_{l_1}}{\partial \vh_{l_0+1}} = \mU^{l_1-l_0+1},
\end{align*}
which according to Sec.~\ref{sec:rect_explode}, will explode as $l\to\infty$ if
\begin{align*}
    e_{\max} > 1,
\end{align*}
where $e_{\max}$ is the largest eigenvalue of $\mU$. When $e_{\max} < 1$, it
will {\em vanish}, i.e., $\|\frac{\partial \vh_{l_1}}{\partial \vh_{l_0+1}}\|
\to 0$, exponentially fast.

What if the recurrent activation function $f$ is {\em not} linear at all? Let's
look at $\frac{\partial \vh_{t}}{\partial \underline{\vh}_{t}} \mU$ as 
\begin{align*}
    \frac{\partial \vh_{t}}{\partial \underline{\vh}_{t}} \mU = 
        \underbrace{
    \left[ 
            \begin{array}{c c c c}
                f_1' & 0 & \cdots & 0 \\
                0 & f_2' & \cdots & 0 \\
                \vdots & \vdots & \cdots & \vdots \\
                0 & 0 & \cdots & f_{n_h}'
            \end{array}
    \right] 
        }_{=\diag\left( \frac{\partial \vh_{t}}{\partial \underline{\vh}_{t}}\right)}
    \left(\mQ \mS \mQ^{-1}\right),
\end{align*}
where we used the eigendecomposition of $\mU=\mQ \mS \mQ^{-1}$. This can be
re-written into
\begin{align*}
    \frac{\partial \vh_{t}}{\partial \underline{\vh}_{t}} \mU = 
    \mQ \left( \diag\left(\frac{\partial \vh_{t}}{\partial
            \underline{\vh}_{t}}\right)
    \odot \mS \right) \mQ^{-1}.
\end{align*}
This means that the eigenvalue of $\mU$ will be scaled by the derivative of the
recurrent activation function at each timestep. In this case, we can bound the
maximum eigenvalue of $\frac{\partial \vh_{t}}{\partial \underline{\vh}_{t}}
\mU$ by
\begin{align*}
    e_{\max}^t \leq \lambda e_{\max},
\end{align*}
where $\lambda$ is the upperbound on $g'=\frac{\partial \vh_{t}}{\partial
\underline{\vh}_{t}}$. See Eqs.~\eqref{eq:sigmoid_bound}--\eqref{eq:tanh_bound}
for the upperbounds of the sigmoid and hyperbolic tangent functions.

In other words, if the largest eigenvalue of $\mU$ is larger than
$\frac{1}{\lambda}$, it is {\em likely} that this temporal derivative of
$\vh_{l_1}$ with respect to $\vh_{l_0+1}$ will {\em explode}, meaning that its
norm will grow exponentially large. In the opposite case of $e_{\max} <
\frac{1}{\lambda}$, the norm of the temporal derivative likely shrinks toward
$0$. The former case is referred to as {\em exploding gradient}, and the latter
{\em vanishing gradient}. These cases were studied already at the very early
years of research in recurrent neural
networks~\cite{bengio1994learning,hochreiter2001gradient}.

Using $\tanh$ is a blessing in recurrent neural networks when running the
network forward, as I described in the previous section. This is however not
necessarily true in the case of backpropagating derivaties. Especially because, 
there is a higher chance of vanishing gradient with $\tanh$, or even worse with
$\sigmoid$. Why? Because $\frac{1}{\lambda} > 1$ for almost everywhere.

\subsection{Are We Doomed?}
\label{sec:rnn_vanish_grad}

\paragraph{Exploding Gradient}

Fortunately it turned out that the phenomenon of exploding gradient is quite
easy to address. First, it is straightforward to detect whether the exploding
gradient happened by inspecting the norm of the gradient fo the cost with
respect to the parameters $\left\| \nabla_{\TT} \tilde{C} \right\|$. If the
gradient's norm is larger than some predefined threhold $\tau > 0$, we can
simply renormalize the norm of the gradient to be $\tau$. Otherwise, we leave it
as it is.

In mathematics,
\begin{align*}
    \tilde{\nabla} = 
    \left\{
        \begin{array}{l l}
            \tau \frac{\nabla}{\|\nabla\|}, & \text{ if } \|\nabla\| > \tau \\
            \nabla, & \text{ otherwise}
        \end{array}
    \right.,
\end{align*}
where we used the shorthand notiation $\nabla$ for $\nabla_{\TT} \tilde{C}$.
$\tilde{\nabla}$ is a rescaled gradient update direction which will be used by
the stochastic gradient descent (SGD) algorithm from Sec.~\ref{sec:sgd}. This
algorithm is referred to as {\em gradient clipping}
\cite{pascanu2013difficulty}.

\paragraph{Vanishing Gradient}

What about vanishing gradient? But, first, what does vanishing gradient mean? We
need to understand the meaning of this phenomenon in order to tell whether this
is a problem at all from the beginning.

Let us consider a case the variable-length output where $|x|=|y|$ from
Sec.~\ref{sec:rnn_x_y}. Let's assume that there exists a clear dependency
between the output label $y_t$ and the input symbol $x_{t'}$, where $t' \ll t$.
This means that the empirical cost will decrease when the weights are adjusted
such that 
\[
    \log p(y_t=y_t^*| \ldots, \phi(x_{t'}), \ldots)
\]
is maximized, where $y_t^*$ is the ground truth output label at time $t$. The
value of $\phi(x_{t'})$ has great influence on the $t$-th output $y_t$, and the
influence can be measured by 
\[ 
    \frac{\partial \log p(y_t=y_t^*| \ldots)}{\partial \phi(x_{t'})}.
\]

Instead of exactly computing $\frac{\partial \log p(y_t=y_t^*| \ldots)}{\partial
\phi(x_{t'})}$, we can approximate it by the finite difference method. Let
$\epsilon \in \RR^{\dim(\phi(x_{t'}))}$ be a vector of which each element is a
very small real value ($\epsilon \approx 0$.) Then,
\begin{align*}
    \frac{\partial \log p(y_t=y_t^*|\ldots)}{\partial \phi(x_{t'})} 
    = \lim_{\epsilon \to 0}& \left(\log p(y_t=y_t^*|\ldots,
    \phi(x_{t'})+\epsilon, \ldots)
    \right.
    \\
&- \left. \log p(y_t=y_t^*|\ldots, \phi(x_{t'}), \ldots, )\right) \oslash \epsilon,
\end{align*}
where $\oslash$ is an element-wise division.  This shows that $\frac{\partial
\log p(y_t=y_t^*|\ldots)}{\partial \phi(x_{t'})}$ computes the difference in the
$t$-th output probability with respect to the change in the value of the $t'$-th
input. 

In other words $\frac{\partial \log p(y_t=y_t^*|\ldots)}{\partial \phi(x_{t'})}$
directly reflects the degree to which the $t$-th output $y_t$ {\em depends} on
the $t'$-th input $x_{t'}$, {\em according to the network}.  To put it in
another way, $\frac{\partial \log p(y_t=y_t^*|\ldots)}{\partial\phi(x_{t'})}$
reflects how much dependency the {\em recurrent neural network has captured the
dependency between $y_t$ and $x_{t'}$}.

Let's rewrite 
\begin{align*}
    \frac{\partial \log p(y_t=y_t^*|\ldots)}{\partial \phi(x_{t'})} 
    = \frac{\partial \log p(y_t=y_t^*|\ldots)}{\partial \vh_t}
    \underbrace{
        \frac{\partial \vh_t}{\partial \vh_{t-1}} \cdots
        \frac{\partial \vh_{t'+1}}{\partial \vh_{t'}} 
    }_{(a)}
    \frac{\partial \vh_{t'}}{\partial \phi(x_t)}.
\end{align*}
The terms marked with (a) looks exactly identical to Eq.~\eqref{eq:tderiv}. We
have already seen that this term can easily {\em vanish} toward zero with a high
probability (see Sec.~\ref{sec:vanishing_grad}.)

This means that the recurrent neural network is unlikely to capture this
dependency. This is especially true when the (temporal) distance between the
output and input, i.e., $|t - t'| \gg 0$.

The biggest issue with this vanishing behaviour is that there is no
straightforward way to avoid it. We cannot tell whether $\frac{\partial \log
p(y_t=y_t^*|\ldots)}{\partial \phi(x_{t'})} \approx 0$ is due to the lack of
this dependency in the true, underlying function or due to the wrong
configuration (parameter setting) of the recurrent neural network. If we are
certain that there are indeed these long-term dependencies, we may
simultaneously minimize the following auxiliary term together with the cost
function:
\begin{align*}
    \sum_{t=1}^T \left( 1 - \frac{\left\| 
            \frac{\partial \tilde{C}}{\partial \vh_{t+1}}
            \frac{\partial \vh_{t+1}}{\partial \vh_t}
    \right\|}{\left\| 
        \frac{\partial \tilde{C}}{\partial \vh_{t+1}}
\right\|} \right)^2.
\end{align*}
This term, which was introduced in \cite{pascanu2013difficulty}, is minimized
when the norm of the derivative does not change as it is being backpropagated,
effectively forcing the gradient {\em not} to vanish.

This term however was found to help significantly only when the target task, or
the underlying function, does indeed exhibit long-term dependencies. How can we
know in advance? Pascanu et al. \cite{pascanu2013difficulty} showed this with
the well-known toy tasks which were specifically designed to exhibit long-term
dependencies \cite{hochreiter2001gradient}.

\subsection{Gated Recurrent Units Address Vanishing Gradient}

Will the same problems of vanishing gradient happen with the gated recurrent
units (GRU) or the long short-term memory units (LSTM)? Let us write the memory 
state at time $t$:
\begin{align*}
    \vh_t =& \vu_t \odot \tilde{\vh}_t + 
    (1 - \vu_t) \odot \left(
        \vu_{t-1} \odot \tilde{\vh}_{t-1} + 
        (1 - \vu_{t-1}) \odot \left(
            \vu_{t-2} \odot \tilde{\vh}_{t-2} + 
            (1 - \vu_{t-2}) \odot \left(
                \cdots
            \right)
        \right)
    \right) 
    \\
    =& \vu_t \odot \tilde{\vh}_t +
    (1-\vu_t) \odot \vu_{t-1} \odot \tilde{\vh}_{t-1} +
    (1-\vu_t) \odot (1-\vu_{t-1}) \odot \vu_{t-2} \odot \tilde{\vh}_{t-2} +
    \cdots
\end{align*}
Let's be more specific and see what happens to this with respect to $x_{t'}$:
\begin{align}
    \vh_t =& \vu_t \odot \tilde{\vh}_t +
    (1-\vu_t) \odot \vu_{t-1} \odot \tilde{\vh}_{t-1} +
    (1-\vu_t) \odot (1-\vu_{t-1}) \odot \vu_{t-2} \odot \tilde{\vh}_{t-2} +
    \cdots 
    \nonumber \\
    \label{eq:gru_expand}
    & +
    \underbrace{
        \left(\prod_{k=t,\ldots,t'+1} (1-\vu_k)\right) \odot \vu_{t'} 
    }_{(a)}
    \odot \tanh \left( \mW \phi(x_{t'}) + \mU \left( \vr_{t'} \odot \vh_{t'-1} \right)
    \right),
\end{align}
where $\prod$ is for element-wise multiplication.

What this implies is that the GRU effectively introduces a {\em shortcut} from
time $t'$ to $t$. The change in $x_{t'}$ will directly influence the value of
$\vh_t$, and subsequently the $t$-th output symbol $y_t$. In other words, all
the issue with the simple recurrent neural network we discussed earlier in
Sec.~\ref{sec:rnn_vanish_grad}.

The update gate controls the strength of these shortcuts. Let's assume for now
that the update gate is fixed to some predefined value between $0$ and $1$. This
effectively makes the GRU a leaky integration unit \cite{bengio2013advances}.
However, as it is perhaps clear from Eq.~\eqref{eq:gru_expand} that we will
inevitably run into an issue. Why is this so? 

Let's say we are sure that there are many long-term dependencies in the data.
It is natural to choose a large coefficient for the leaky integration unit,
meaning the update gate is close to $1$. This will definitely help carrying the
dependency across many time steps, but this inevitably carries {\em unnecessary}
information as well. This means that much of the representational power of the
output function $g_{\text{out}}(\vh_t)$ is wasted in {\em ignoring} those
unnecessary information.

If the update gate is fixed to something substantially smaller than $1$, all the
shortcuts (see Eq.~\eqref{eq:gru_expand} (a)) will exponentially vanish. Why?
Because it is a repeated multiplication of a scalar small than 1. In other
words, it does not really help to have a leaky integration unit in the place of
a simple $\tanh$ unit.

This is however not the case with the actual GRU or LSTM, because those update
gates are {\em not} fixed but are adaptive with respect to the input. If the
network detects that there is an important dependency being captured, the update
gate will be closed ($u_j \approx 0$.) This will effectively strengthen the
shortcut connection (see Eq.~\eqref{eq:gru_expand} (a).) When the network
detects that there is no dependency anymore, it will open the update gate ($u_j
\approx 1$), which effectively cuts off the shortcut. How does the network know,
or detect, the existence or lack of these dependencies? Do we need to manually
code this up? I will leave these questions for you to figure out.

%

\chapter{Neural Language Models}
\label{chap:nlm}

\section{Language Modeling: First Step}

What does it mean for a machine to understand natural language? In other words,
how can we tell that the machine understood natural language?  These are the two
equivalent questions that are at the core of this course. 

One of the most basic capability of a machine that can signal us that it indeed
understands natural language is for the machine to tell how likely a given
sentence is. Of course, this is extremely ill-defined, as we probably cannot
define the {\em likeliness} of a sentence, because there are many different
types of unlikeliness. For instance, a sentence ``Colorless green ideas sleep
furiously'' from Chomsky's \cite{chomsky2002syntactic} is {\em unlikely}
according to our common sense, because 
\begin{enumerate}
    \itemsep 0em
    \item An object (``idea'') cannot be both ``colorless'' and ``green.''
    \item An object cannot ``sleep'' ``furiously.''
    \item An ``idea'' does not ``sleep.''
\end{enumerate}
On the other hand, this sentence is a grammatically correct sentence.

Let's take a look at another sentence ``Jane and me went to see a movie
yesterday.'' Grammatically, this is not the most correct sentence one can make.
It should be ``Jane and I went to see a movie yesterday.'' Even with a
grammatical error in the original sentence however, the meaning of the sentence
is clear to me, and perhaps is much more understandable than the sentence
``colorless green ideas sleep furiously.'' Furthermore, many people likely say
this (saying ``me'' instead of ``I'') quite often. This sentence is thus likely
according to our common sense, but is not likely according to the grammar.

This observation makes us wonder what is the criterion to use. Is it correct for
a machine to tell whether the sentence is likely by analyzing its grammatical
correctness? Or, is it possible that the machine should deem a sentence likely
only when its meaning agrees well with common sense regardless of its
grammatical correctness in the most strict sense?

As we discussed in the first lecture of the course, we are more interested in
approaching natural language as a means for one to communicate ideas to a
listener. In this sense, language use is a function which takes as input the
surrounding environment, including the others' speech, and returns linguistic
response, and this function is {\em not} given but learned via observing others'
use of language and the reinforcement by the surrounding
environment~\cite{skinner2014verbal}. Also, throughout this course, we are not
concerned too much about the existing syntactic, or grammatical, structures
underlying natural language, which makes it difficult for us to say anything
about the grammatical correctness of a given sentence.

In short, we take the route here that the likeliness of a sentence be determined
based on its agreement with common sense. The common sense here is captured by
everyday use of natural language, which consequently implies that the statistics
of natural language use can be a strong indicator for determining the likely of
a natural language sentence. 

\subsection{What if those linguistic structures do exist}
\label{sec:linguistic_lm}

Of course, as we discussed earlier in Sec.~\ref{sec:wrong_route} and in this
section, not everyone agrees.  This is due to the fact that a perfect
grammatical sentence may be considered unlikely, just because it does not happen
often. In other words, statistical approaches to {\em language modeling} may
conclude that a sentence with perfectly valid grammatical construction is
unlikely. Is this a problem? 

This problem of telling how likely a given sentence is can be viewed very
naturally as building a probabilistic model of sentences. In other words, given
a sentence $S$, what is the probability $p(S)$ of $S$?  Let us briefly talk
about what this means for the case of viewing the likeliness of a sentence as
equivalent to its grammatical correctness.\footnote{
    Why briefly and why here? Because, we will not pursue this line at all after
    this section.
}

We first assume that there is an underlying linguistic structure $G$ which has
generated the observed sentence $S$. Of course, we do not know the correct $G$
in advance, and unfortunately no one will tell us what the correct $G$
is.\footnote{
    Here, the correct $G$ means the $G$ that generated $S$, not the whole
    structure of $G$ which is assumed to exist according to a certain set of
    rules.
}
Thus,
$G$ is a hidden variable in this case. This hidden structure $G$ generates the
observed sentence $S$ according to an unknown conditional distribution $p(S|G)$.
Each and every grammatical structure $G$ is assigned a prior probability which
is also unknown in advance.\footnote{
    This is not necessarily true. If we believe that each and every grammatical
    correct sentence is equally likely and that each correct grammatical
    structure generates a single corresponding sentence, the prior distribution
    over the hidden linguistic structure is such that any correct structure is
    given an equal probability while any incorrect structure is given a zero
    probability. But, of course, if we think about it, there are clearly certain
    structures that are more prevalent and others that are not.
}

With the conditional distribution $S|G$ and the prior distribution $G$, we
easily get the joint distribution $S,G$ by
\begin{align*}
    p(S,G) = p(S|G) p(G),
\end{align*}
from the definition of conditional probability.\footnote{
    A conditional probability of $A$ given $B$ is defined as
    \begin{align*}
        p(A|B) = \frac{p(A, B)}{p(B)}
    \end{align*}
} 
From this joint distribution we get the distribution over a given sentence $S$
by marginalizing out $G$:
\begin{align*}
    p(S) = \sum_{G} p(S, G).
\end{align*}
This means that we should compute how likely a given sentence $S$ is with
respect to all possible underlying linguistic structure. This is very likely
intractable, because there must be infinite possible such structures.

Instead of computing $p(S)$ exactly we can simply look at its lowerbound. For
instance, one simplest, and probably not the best, way to do so is
\begin{align*}
    p(S) = \sum_G p(S,G) \geq p(S, \hat{G}),
\end{align*}
where $\hat{G} = \argmax_G p(S,G) = \argmax_G p(G|S)$.\footnote{
    This inequality holds due to the definition of probability, which states
    that $p(X)\geq 0$ and $\sum_X p(X) = 1$.
}

This lowerbound is tight, i.e., $p(S) = p(S, \hat{G})$, when there is only a
single true underlying linguistic structure $\hat{G}$ given $S$. What this says
is that there is no other possible linguistic structure possible for a single
observed sentence, i.e., no ambiguity in inferring the correct linguistic
structure.  In other words, we can compute the probability or likeliness of a
given sentence by inferring its correct underlying linguistic structure.

However, there are a few issues here. First, it is not clear which formalism $G$
follows, and we have briefly discussed about this at the very beginning of this
course. Second, it is quite well known that most of the formalisms do indeed
have uncertainty in inference. Again, we looked at one particular example in
Sec.~\ref{sec:language_understanding_wrong}. These two issues make many people,
including myself, quite uneasy about this type of {\em model-based} approaches. 

In the remaining of this chapter, I will thus talk about model-free approaches
(as opposed to these model-based approaches.)

\subsection{Quick Note on Linguistic Units}
\label{sec:ling_unit}

Before continuing, there is one question that must be bugging you, or at least
has bugged me a lot: {\em what is the minimal linguistic unit?} 

If we think about written text, the minimal unit does seem like a character.
With spoken language, the minimal unit seems to be a phoneme. But, is this the
level at which we want to model the process of {\em understanding} natural
language? In fact, to most of the existing natural language processing
researchers as well as some (or most) linguists, the answer to this question is
a hard ``no.''

The main reason is that these low-level units, both characters and phonemes, do
not convey any meaning themselves. Does a Latin alphabet ``q'' have its own
meaning? The answer by most of the people will be ``no.'' Then, starting from
this alphabet ``q'', how far should we climb up in the hierarchy of linguistic
units to reach a level at which the unit begins to convey its own meaning?
``qu'' does not seem to have its own meaning still. ``qui'' in French means
``who'', but in English it does not really say much. ``quit'' in English is a
valid word that has its own meaning, and similarly ``quiet'' is a valid word
that has its own meaning, quite apart from that of ``quit.''

It looks like a word is the level at which meaning begins to form itself.
However, this raises a follow-up question on the definition of a word: {\em What
is a word?}

It is tempting to say that a sequence of non-blank characters is a word. This
makes everyone's life so much easier, because we can simply split each sentence
by a blank space to get a sequence of words. Unfortunately this is a very bad
strategy. The simplest counter example to this definition of words is a token
(which I will use to refer to a sequence of non-blank characters) consisting of
a word followed by a punctuation. If we simply split a sentence into words by
blank spaces, we will get a bunch of redundant words. For instance, ``llama'',
``llama,'', ``llama.'', ``llama?'', ``"llama'', ``llama"'' and ``llama!'' will
all be distinct words. We will run into an issue of exploding vocabulary with
any morphologically rich language. Furthermore, in some languages such as
Chinese, there is no blank space at all inside a sentence, in which case this
simple strategy will completely fail to give us any meaningful, small linguistic
unit other than sentences.

Now at this point it almost seems like the best strategy is to use each
character as a linguistic unit. This is not necessarily true due to the highly
nonlinear nature of orthography.\footnote{
    Orthography is defined as ``the study of spelling and how letters combine to
    represent sounds and form words.''
} There are many examples in which this nonlinear nature shows its difficulty.
One such example is to consider the following three words: ``quite'', ``quiet''
and ``quit''.\footnote{
    I would like to thank Bart van Merrienboer for this example.
}
All three character sequences have near identical forms, but their corresponding
meanings differ from each other substantially. In other words, any function that
maps from these character sequences to the corresponding meanings will have to
be extremely nonlinear and thus difficult to be learned from data. Of course,
this is an area with active research, and I hope I am not giving you an
impression that characters are not the units to use (see, e.g.,
\cite{kim2015character}.)

Now then the question is whether there is some middle ground between characters
and words (or blank-space-separated tokens) that are more suitable to be used as
elementary linguistic units (see, e.g., \cite{sennrich2015neural}.)
Unfortunately this is again an area with active research.  Hopefully, we will
have time later in the course to discuss this issue further.  For now, we will
simply use blank-space-separated tokens as our linguistic units.

\section{Statistical Language Modeling}
\label{sec:lm}

Regardless of which linguistic unit we use, any natural language sentence $S$
can be represented as a sequence of $T$ discrete symbols such that
\begin{align*}
    S = (w_1, w_2, \ldots, w_T).
\end{align*}
Each symbol is one element from a {\em vocabulary} $V$ which contains all
possible symbols:
\begin{align*}
    V = \left\{ v_1, v_2, \ldots, v_{|V|}\right\},
\end{align*}
where $|V|$ is used to mean the size of the vocabulary, or the number of all
symbols.

The problem of language modeling is equivalent to finding a model that assigns a
probability $p(S)$ to a sentence:
\begin{align}
    \label{eq:sentence_prob}
    p(S) = p(w_1, w_2, \ldots, w_T).
\end{align}
Of course, we are not given this distribution and need to learn this from data.

Let's say we are given data $D$ which contains $N$ sentences such that
\begin{align*}
    D = \left\{ S^1, S^2, \ldots, S^N \right\},
\end{align*}
where each sentence $S^n$ is 
\begin{align*}
    S^n = (w_1^n, w_2^n, \ldots, w_{T^n}^n),
\end{align*}
meaning that each sentence has a different length.

Given this data $D$, let us estimate the probability of a certain sentence $S$.
This is quite straightforward:
\begin{align}
    \label{eq:sentence_mle}
    p(S) = \frac{\sum_{n=1}^N \II_{S = S^n}}{N},
\end{align}
where $\II$ is the indicator function defined earlier in
Eq.~\eqref{eq:indicator_function} which is defined as
\begin{align*}
    \II_{S=S^n} = \left\{ 
        \begin{array}{c c}
            1, & \text{if }S=S^n \\
            0, & \text{otherwise}
        \end{array}
        \right.
\end{align*}
This is equivalent to counting how many times $S$ occurs in the data.\footnote{
    A data set consisting of (written) text is often referred to as a {\em
    corpus}. 
}

\subsection{Data Sparsity/Scarcity}
\label{sec:data_sparsity}

Has this solved the whole problem of language model? No, unfortunately not. The
very major issue here is that however large your corpus is, it is unlikely to
contain all reasonable sentences in the world. Let's do simple counting here.

There are $|V|$ symbols in a vocabulary. Each sentence can be as long as $T$
symbols. Then, there are $|V|^T$ possible sentences. A reasonable range for the
sentence length $T$ is roughly between $1$ to $50$, meaning that there are
\begin{align*}
    \sum_{T=1}^{50} |V|^T
\end{align*}
possible sentences. As it's quite clear, this is a huge space of sentences.

Of course, not all those sentences are plausible. This is however conceivable
that even the fraction of that space will be gigantic, especially considering
that the size of vocabulary often goes up to $100$k to 1M words. Many of the
plausible sentences will not appear in the corpus. Is this true? In fact, yes,
it is. 

\begin{figure}[ht]
    \centering
    \includegraphics[width=0.6\textwidth]{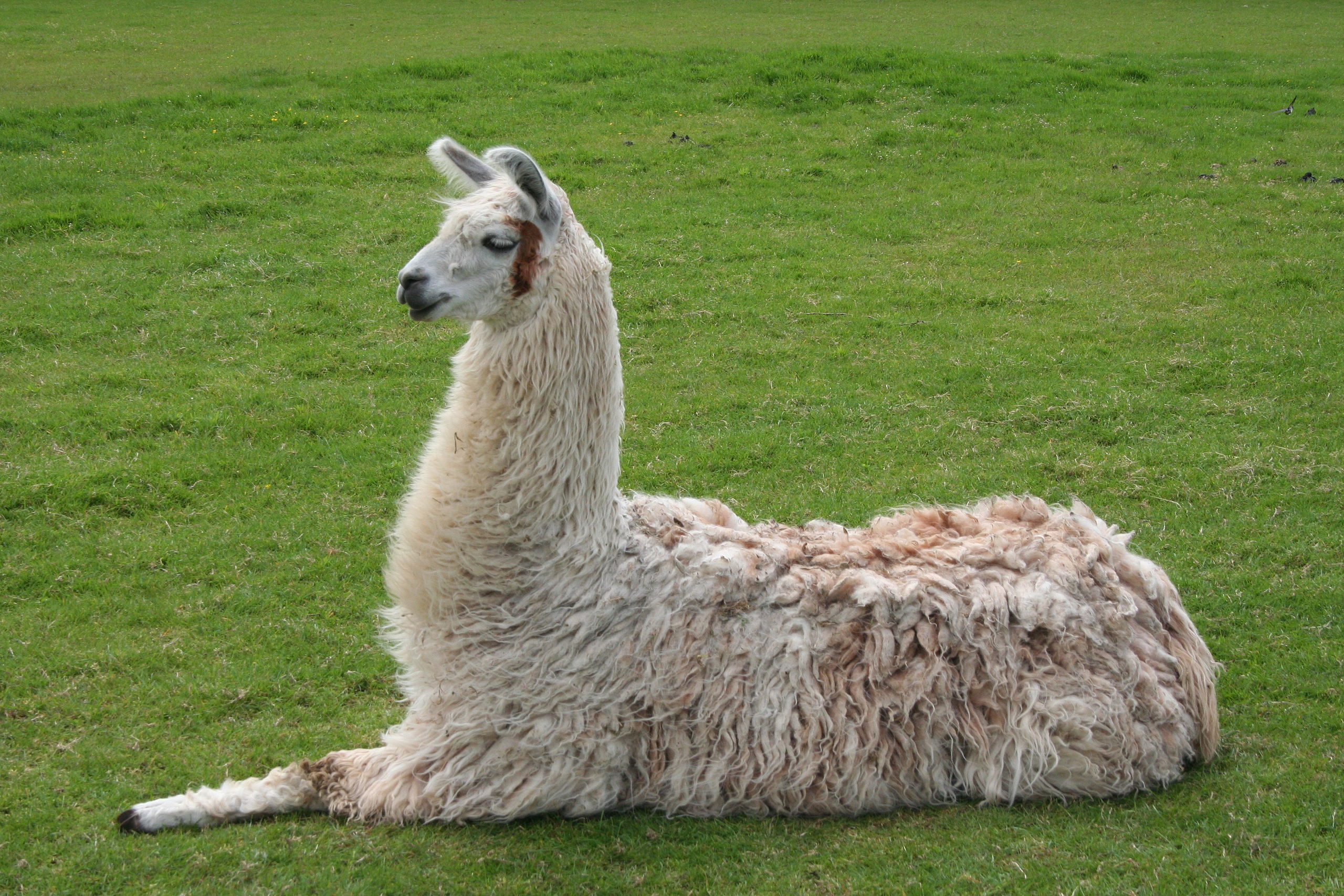}
    \caption{A picture of a llama lying down. From
    \url{https://en.wikipedia.org/wiki/Llama}}
    \label{fig:llama}
\end{figure}

\begin{figure}[ht]
    \centering
    \includegraphics[width=\textwidth]{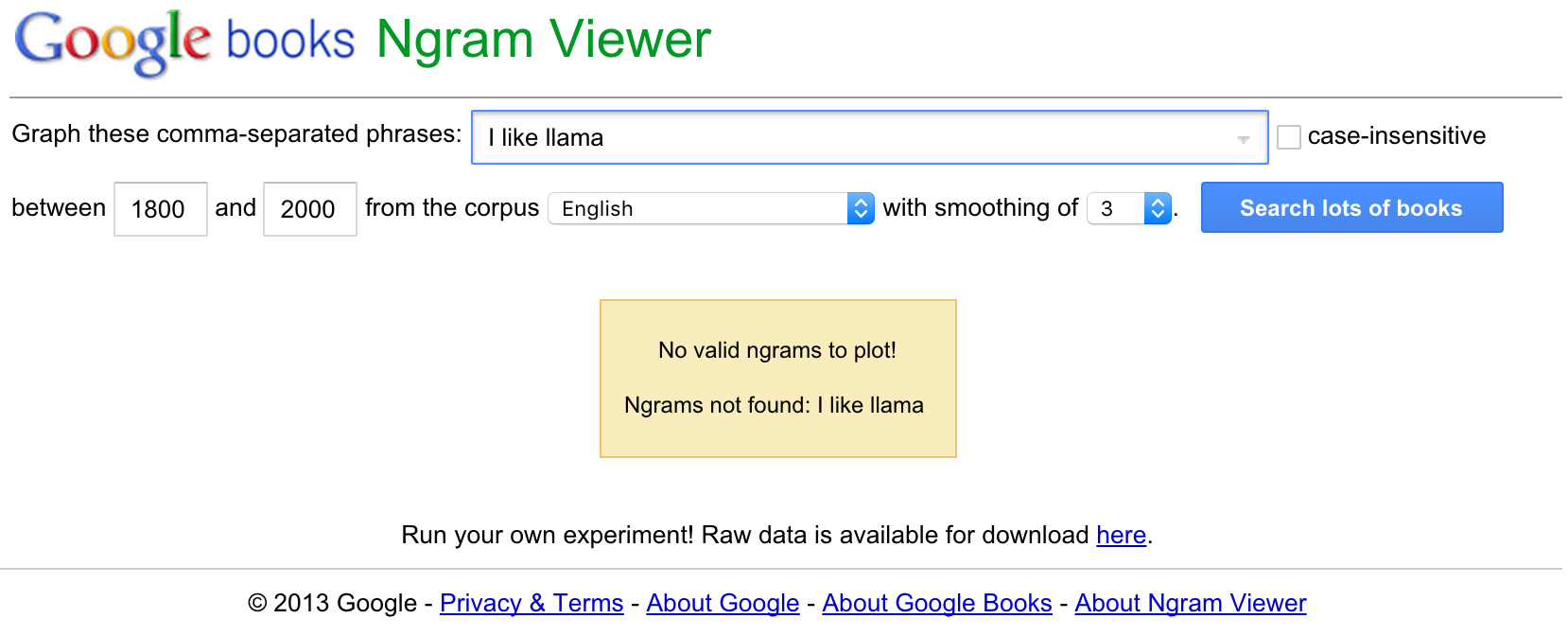}
    \caption{A resulting page of Google Books Ngram Viewer for the query ``I
    like llama''.}
    \label{fig:ngram_fail}
\end{figure}

It is quite easy to find such an example. For instance, Google Books Ngram
Viewer\footnote{
    \url{https://books.google.com/ngrams}
}
lets you search for a sentence or a sequence of up to five English words from
the gigantic corpus of Google Books. Let me try to search for a very plausible
sentence ``I like llama,'' and the Google Books Ngram\footnote{
    We will discuss what Ngrams are in the later sections.
} 
Viewer returns an error
saying that ``Ngrams not found: I like llama.'' (see Fig.~\ref{fig:llama} in the case you
are not familiar with a llama.) See Fig.~\ref{fig:ngram_fail} as an evidence.

What does this mean for the estimate in Eq.~\eqref{eq:sentence_mle}? It means
that this estimator will be too harsh for many of the plausible sentences that
do not occur in the data. As soon as a given sentence does not appear {\em
exactly} as it is in the corpus, this estimator will say that there is a {\em
zero} probability of the given sentence. Although the sentence ``I like llama''
is a likely sentence, according to this estimator in
Eq.~\eqref{eq:sentence_mle}, it will be deemed extremely unlikely.

This problem is due to the issue of {\em data sparsity}. Data sparsity here
refers to the phenomenon where a training set does not cover the whole space of
input sufficiently. In more concrete terms, most of the points in the input
space, which have non-zero probabilities according to the true, underlying
distribution, do not appear in the training set. If the size of a training set
is assumed to be fixed, the severity of data sparsity increases as the average,
or maximum length of the sentences. This follows from the fact that the size of
the input space, the set of all possible sentences, grows with respect to the
maximum possible length of a sentence.

In the next section, we will discuss the most straightforward approach to
addressing this issue of data sparsity.

\section{$n$-Gram Language Model}
\label{sec:n_gram_lm}

The fact that the issue of data sparsity worsens as the maximum length of
sentences grows hints us a straightforward approach to addressing this: {\em
limit the maximum length of phrases/sentences we estimate a probability on}.
This idea is a foundation on which a so-called $n$-gram language model is based.

In the $n$-gram language model, we first rewrite the probability of a given
sentence $S$ from Eq.~\eqref{eq:sentence_prob} into
\begin{align}
    \label{eq:unidir_sentence}
    p(S) = p(w_1, w_2, \ldots, w_T) = p(w_1)p(w_2|w_1)\cdots
    \underbrace{p(w_k|w_{<k})}_{(a)}
    \cdots p(w_T|w_{<T}),
\end{align}
where $w_{<k}$ denotes all the symbols before the $k$-th symbol $w_k$. From
this, the $n$-gram language model makes an important assumption that each
conditional probability (Eq.~\eqref{eq:unidir_sentence}~(a)) is only conditioned
on the $n-1$ preceding symbols only, meaning
\begin{align*}
    p(w_k | w_{<k}) \approx p(w_k | w_{k-n}, w_{k-n+1}, \ldots, w_{k-1}).
\end{align*}
This results in 
\begin{align*}
    p(S) \approx \prod_{t=1}^T p(w_t | w_{t-n}, \ldots, w_{t-1}).
\end{align*}

What does this mean? Under this assumption we are saying that any symbol in a
sentence is {\em predictable} based on the $n-1$ preceding symbols. This is in
fact a quite reasonable assumption in many languages. For instance, let us
consider a phrase ``I am from''. Even without any more context information
surrounding this phrase, such as surrounding words and the identity of a
speaker, we know that the word following this phrase will be likely a name of
place or country. In other words, the probability of a name of place or country
given the three preceding words ``I am from'' is higher than that of any other
words. 

But, of course, this assumption does not always work. For instance, consider a
phrase ``In Korea, more than half of all the residents speak
$\underbrace{\text{Korean}}_{(a)}$.'' Let us focus on the last word ``Korean''
(marked with (a).) We immediately see that it will be useful to condition its
conditional probability on the second word ``Korea''. Why is this so? Because
the conditional probability of ``Korean'' following ``speak'' should
significantly increase over all the other words (that correspond to other
languages) knowing the fact that the sentence is talking about the residents of
``Korea''. This requires the conditional distribution to be conditioned on at
least 10 words (``,'' is considered a separate word,) and this certainly will
not be captured by $n$-gram language model with $n < 9$.  

From these examples it is clear that there's a natural trade-off between the
quality of probability estimate and statistical efficiency based on the choice
of $n$ in $n$-gram language modeling. The higher $n$ the longer context the
conditional distribution has, leading to a better model/estimate (second
example,) however resulting in a situation of more sever data sparsity (see
Sec.~\ref{sec:data_sparsity}.) On the other hand, the lower $n$ leads to the
worse language modeling (second example), but this will avoid the issue of data
sparsity. 

\paragraph{$n$-gram Probability Estimation}

We can estimate the $n$-gram conditional probability $p(w_k | w_{k-n}, \ldots,
w_{k-1})$ from the training corpus.
Since it is a
{\em conditional} probability, we need to rewrite it according to the definition
of the conditional probability:
\begin{align}
    \label{eq:n_gram_prob}
    p(w_k | w_{k-n}, \ldots, w_{k-1}) = \frac{p(
    w_{k-n}, \ldots, w_{k-1}, w_k)}
    {p(w_{k-n}, \ldots, w_{k-1})}
\end{align}
This rewrite implies that the $n$-gram probability is equivalent to counting
the occurrences of the $n$-gram $(w_{k-n}, \ldots, w_{k})$
among all $n$-grams starting with $(w_{k-n}, \ldots, w_{k-1})$.

Let us consider the denominator first. The denominator can be computed by the
marginalizing the $k$-th word ($w'$ below):
\begin{align}
    \label{eq:n_gram_marginal}
    p(w_{k-n}, \ldots, w_{k-1}) = \sum_{w'\in V} p(w_{k-n}, \ldots, w_{k-1}, w').
\end{align}
From Eq.~\eqref{eq:sentence_mle}, we know how to estimate $p(w_{k-n}, \ldots,
w_{k-1}, w')$:
\begin{align}
    \label{eq:n_gram_expand}
    p(w_{k-n}, \ldots, w_{k-1}, w') \approx \frac{c(w_{k-n}, \ldots, w_{k-1},
    w')}{N_n},
\end{align}
where $c(\cdot)$ is the number of occurrences of the given $n$-gram in the
training corpus, and $N_n$ is the number of all $n$-grams in the training
corpus.

Now let's plug Eq.~\eqref{eq:n_gram_expand} into
Eqs.~\eqref{eq:n_gram_prob}--\eqref{eq:n_gram_marginal}:
\begin{align}
    \label{eq:n_gram_mle}
    p(w_k | w_{k-n}, \ldots, w_{k-1}) = \frac{
        \cancel{\frac{1}{N_n}}
        c(w_{k-n}, \ldots, w_{k-1}, w_k)}
    {
        \cancel{\frac{1}{N_n}} \sum_{w' \in V} c(w_{k-n}, \ldots, w_{k-1},
    w')
}
\end{align}

\subsection{Smoothing and Back-Off}

{\em Note that I am missing many references this section, as I am writing this
    on my travel. I will fill in missing references once I'm back from my
travel.}

The biggest issue of having an $n$-gram that never occurs in the training corpus
is that any sentence containing the $n$-gram will be given a zero probability
regardless of how likely all the other $n$-grams are. Let us continue with the
example of ``I like llama''. With an $n$-gram language model built using all the
books in Google Books, the following, totally valid sentence\footnote{
    This is not strictly true, as I should put ``a'' in front of the llama.
}
will be given a
zero probability:
\begin{itemize}
    \itemsep 0em
    \item ``I like llama which is a domesticated South American
        camelid.\footnote{
            The description of a llama taken from Wikipedia:
            \url{https://en.wikipedia.org/wiki/Llama}
        }
\end{itemize}
Why is this so? Because the probability of this sentence is given as a product
of all possible trigrams:
\begin{align*}
    p(&\text{``I'', ``like'', ``llama'', ``which'', ``is'', ``a'',
        ``domesticated'', ``South'',
    ``American'', ``camelid''}) \\
    =& p(\text{``I''}) p(\text{``like''}|\text{``I''})
    \underbrace{p(\text{``llama''}|\text{``I''}, \text{``like''})}_{=0} \cdots
    p(\text{``camelid''}|\text{``South''}, \text{``American''}) \\
    =& 0
\end{align*}

One may mistakenly believe that we can simply increase the size of corpus
(collecting even more data) to avoid this issue. However, remember that ``{\it
data sparsity is almost always an issue in statistical modeling}''
\cite{chen1996empirical}, which means that more data call for better statistical
models with often more parameters leading to the issue of data sparsity.

One way to alleviate this problem is to assign a small probability to all {\em
unseen} $n$-grams. At least, in this case, we will assign some small, non-zero
probability to any sentence, thereby avoiding a valid, but zero-probability
sentence under the $n$-gram language model. One simplest implementation of this
approach is to assume that each and every $n$-gram occurs at least $\alpha$
times and any occurrence in the training corpus is in addition to this
background occurrence. 

In this case, the estimate of an $n$-gram becomes
\begin{align*}
    p(w_k | w_{k-n}, \ldots, w_{k-1}) =&
    \frac{\alpha + c(w_{k-n}, w_{k-n+1}, \ldots, w_k)}{\sum_{w' \in V} (\alpha
    + c(w_{k-n}, w_{k-n+1}, \ldots, w'))} 
    \\
    =&
    \frac{\alpha + c(w_{k-n}, w_{k-n+1}, \ldots, w_k)}{\alpha |V| + \sum_{w' \in V}
    c(w_{k-n}, w_{k-n+1}, \ldots, w')},
\end{align*}
where $c(w_{k-n}, w_{k-n+1}, \ldots, w_k)$ is the number of occurrences of the
given $n$-gram in the training corpus. $c(w_{k-n}, w_{k-n+1}, \ldots, w')$ is
the number of occurrences of the given $n$-gram if the last word $w_k$ is
substituted with a word $w'$ from the vocabulary $V$. $\alpha$ is often set to
be a scalar such that $0 < \alpha \leq 1$. See the difference from the original
estimate in Eq.~\eqref{eq:n_gram_mle}.

It is quite easy to see that this is a quite horrible estimator: how does it
make sense to say that every unseen $n$-gram occurs with the same frequency?
Also, knowing that this is a horrible approach, what can we do about this?

One possibility is to smooth the $n$-gram probability by interpolating between
the estimate of the $n$-gram probability in Eq.~\eqref{eq:n_gram_mle} and the
estimate of the $(n-1)$-gram probability. This can written down as
\begin{align}
    p^S(w_k | w_{k-n}, \ldots, w_{k-1}) = &
    \lambda(w_{k-n}, \ldots, w_{k-1}) p(w_k | w_{k-n}, \ldots, w_{k-1}) 
    \nonumber \\
    \label{eq:n_gram_smooth}
    & + (1 - \lambda(w_{k-n}, \ldots, w_{k-1})) p^S(w_k | w_{k-n+1}, \ldots,
    w_{k-1}).
\end{align}
This implies that the $n$-gram (smoothed) probability is computed recursively by
the lower-order $n$-gram probabilities. This is clearly an effective strategy,
considering that falling off to the lower-order $n$-grams contains at least some
information of the original $n$-gram, unlike the previous approach of adding a
scalar $\alpha$ to every possible $n$-gram.

Now a big question here is how the interpolation coefficient $\lambda$ is
computed. The simplest approach we can think of is to fit it to the data as
well. However, the situation is not that easy, as using the same training
corpus, which was used to estimate $p(w_k | w_{k-n}, \ldots, w_{k-1})$ according
to Eq.~\eqref{eq:n_gram_mle}, will lead to a degenerate case. What is this
degenerate case? If the same corpus is used to fit both the non-smoothed
$n$-gram probability and $\lambda$'s, the optimal solution is to simply set all
$\lambda$'s to $1$, as that will assign the high probabilities to all the
$n$-grams. Therefore, one needs to use a separate corpus to fit $\lambda$'s.

More generally, we may rewrite Eq.~\eqref{eq:n_gram_smooth} as
\begin{align}
    \label{eq:n_gram_smoothing_general}
    p^S(w_k | w_{k-n}, \ldots, w_{k-1}) = \left\{ 
        \begin{array}{l}
            \alpha(w_k | w_{k-n}, \ldots, w_{k-1}), \text{ if } c(w_{k-n},
            \ldots, w_{k-1}, w_k) > 0 \\
            \gamma(w_{k-n+1}, \ldots, w_{k}) p^S(w_{k}|w_{k-n+1}, \ldots,
            w_{k-1}), \text{ otherwise}
        \end{array}
        \right.
\end{align}
following the notation introduced in \cite{kneser1995improved}. Specific choices
of $\alpha$ and $\gamma$ lead to a number of different smoothing techniques. For
an extensive list of these smoothing techniques, see \cite{chen1996empirical}.

Before ending this section on smoothing techniques for $n$-gram language
modeling, let me briefly describe one of the most widely used smoothing
technique, called the modified Kneser-Ney smoothing (KN smoothing), described in
\cite{chen1996empirical}. This modified KN smoothing is efficiently implemented
in the open-source software package called KenLM~\cite{Heafield-estimate}.

First, let us define some quantities. We will use $n_k$ to denote the total
number of $n$-grams that occur exactly $k$ times in the training corpus. With
this, we define the following so-called discounting factors:
\begin{align*}
    Y =& \frac{n_1}{n_1 + 2 n_2} \\
    D_1 =& 1 - 2 Y \frac{n_2}{n_1} \\
    D_2 =& 2 - 3 Y \frac{n_3}{n_2} \\
    D_{3+} =& 3 - 4 Y \frac{n_4}{n_3}.
\end{align*}
Also, let us define the following quantities describing the number of all
possible words following a given $n$-gram with a specified frequency $l$:
\begin{align*}
    N_l(w_{k-n}, \ldots, w_{k-1}) = |\{ c(w_{k-n}, \ldots, w_{k-1}, w_k) = l \}|
\end{align*}

The modified KN smoothing then defines $\alpha$ in
Eq.~\eqref{eq:n_gram_smoothing_general} to be
\begin{align*}
    \alpha(w_k | w_{k-n}, \ldots, w_{k-1}) =
    \frac{
        c(w_{k-n}, \ldots, w_{k-1}, w_k) - D(c(w_{k-n}, \ldots, w_{k-1}, w_k))
    }{
        \sum_{w' \in V} c(w_{k-n}, \ldots, w_{k-1}, w')
    },
\end{align*}
where $D$ is
\begin{align*}
    D(c) = \left\{
        \begin{array}{l l}
            0,&\text{ if }c=0 \\
            D_1,&\text{ if }c=1 \\
            D_2,&\text{ if }c=2 \\
            D_{3+},&\text{ if }c\geq 3 \\
        \end{array}
        \right.
\end{align*}
And, $\gamma$ is defined as
\begin{align*}
    \gamma(w_{k-n}, \ldots, w_{k-1})  = 
    \frac{
        D_1 N_1(w_{k-n}, \ldots, w_{k-1}) 
        + D_2 N_2(w_{k-n}, \ldots, w_{k-1})
        + D_{3+} N_{3+}(w_{k-n}, \ldots, w_{k-1})
    }{
        \sum_{w' \in V} c(w_{k-n}, \ldots, w_{k-1}, w')
    }.
\end{align*}

For details on how this modified KN smoothing has been designed, see
\cite{chen1996empirical}.



\subsection{Lack of Generalization}

Although $n$-gram language modelling works like a charm in many cases. This is
still not totally satisfactory, because of the lack of {\em generalization}.
What do I mean by {\em generalization} here?

Consider an example where three trigrams\footnote{
    Is ``trigram'' a proper term? Certainly not, but it is widely accepted by
    the whole community of natural language processing researchers. Here's an
    interesting discussion on how $n$-grams should be referred to as, from
    \cite{manning1999foundations}: ``{\it 
        these
        alternatives are usually referred to as a bigram, a trigram, and a
        four-gram model, respectively. Revealing this will surely be enough to
        cause any Classicists who are reading this book to stop, and to leave
        the field to uneducated engineering sorts ... with the declining levels
        of education in recent decades ... some people do make an attempt at
        appearing educated by saying quadgram
    }'' 
} were observed from a training corpus: ``chases a cat'', ``chases a dog'' and
``chases a rabbit''. There is a clear pattern here. The pattern is that it is
highly likely that ``chases a'' will be followed by an animal. 

How do we know this? This is a trivial example of humans' generalization
ability. We have noticed a higher-level concept, in this case an animal, from
observing words such as ``cat'', ``dog'' and ``rabbit'', and based on this
concept, we generalize this knowledge (that ``chases a'' is followed by an
animal) to unseen trigrams in the form of ``chases a [animal]''.

This however does not happen with $n$-gram language model. As an example, let's
consider a trigram ``chases a llama''. Unless this specific trigram occurred
more than once in the training corpus, the conditional probability given by 
$n$-gram language modeling will be zero.\footnote{
    Here we assume that no smoothing or backoff is used. However, even when
    these techniques are used, we cannot be satisfied, since the probability
    assigned to this trigram will be at best reasonable up to the point that the
    $n$-gram language model is giving as high probability as the bigram ``chases
    a''. In other words, we do not get any generalization based on the fact that
    a ``llama'' is an animal similar to a ``cat'', ``dog'' or ``rabbit''.
} This issue is closely related to data sparsity, but the main difference is
that it is not the lack of data, or $n$-grams, but the lack of world knowledge.
In other words, there {\em exist} relevant $n$-grams in the training corpus, but
$n$-gram language modelling is not able to exploit these.

At this point, it almost seems trivial to address this issue by incorporating
existing knowledge into language modelling. For instance, one can think of using
a dictionary to find the definition of a word in interest (continuing on from
the previous example, the definition of ``llama'') and letting the language
model notice that ``llama'' is a ``{\em a domesticated pack animal of the camel
family found in the Andes, valued for its soft woolly fleece}.'' Based on this,
the language model should figure out that the probability of ``chases a llama''
should be similar to ``chases a cat'', ``chases a dog'' or ``chases a rabbit''
because all ``cat'', ``dog'' and ``rabbit'' are animals according to the
dictionary.

This is however not satisfactory for us. First, those definitions are yet
another natural language text, and letting the model understand it becomes
equivalent to natural language understanding (which is the end-goal of this
whole course!) Second, a dictionary or any human-curated knowledge base is
an inherently limited resource. These are limited in the sense that they are
often static (not changing rapidly to reflect the changes in language use) and
are often too generic, potentially not capturing any domain-specific knowledge. 

In the next section, I will describe an approach purely based on statistics of
natural language that is able to alleviate this lack of generalization.

\section{Neural Language Model}
\label{sec:nlm}

One thing we notice from $n$-gram language modelling is that this boils down to
computing the conditional distribution of a next word $w_k$ given $n-1$
preceding words $w_{k-n}, \ldots, w_{k-1}$. In other words, the goal of $n$-gram
language modeling is to find a function that takes as input $n-1$ words and
returns a conditional probability of a next word:
\begin{align*}
    p(w_k | w_{k-n}, \ldots, w_{k-1}) = f_{\TT}^{w_k} (w_{k-n}, \ldots,
    w_{k-1}).
\end{align*}
This is almost exactly what we have learned in
Chapter~\ref{chap:function_approx}.

First, we should define the input to this language modelling function. Clearly
the input will be a sequence of $n-1$ words, but the question is how each of
these words will be represented. Since our goal is to put the least amount of
prior knowledge, we want to represent each word such that each and every word in
the vocabulary is equi-distant away from the others. One encoding scheme that
achieves this goal is $1$-of-$K$ coding. 

In this $1$-of-$K$ coding scheme, each word $i$ in the vocabulary $V$ is
represented as a binary vector $\vw_i$ whose sum equals $1$. To denote the $i$-th
word with the vector $\vw_i$, we set the $i$-th element of the vector $\vw_i$ to be
$1$ (and consequently all the other elements are set to zero.) Mathematically,
\begin{align}
    \label{eq:one_hot_vector}
    \vw_i = [0, 0, \ldots, \underbrace{1}_{\text{$i$-th element}}, \ldots,
    0]^{\top} \in \{0,1\}^{|V|}
\end{align}
This kind of vector is often called a one-hot vector.

It is easy to see that this encoding scheme perfectly fits our goal of having
minimal prior, because
\begin{align*}
    |\vw_i - \vw_j| = \left\{
        \begin{array}{l l}
            1,&\text{ if }i\neq j \\
            0,&\text{ otherwise}
        \end{array}
        \right.
\end{align*}

Now the input to our function is a sequence of $n-1$ such vectors, which I will
denote by $(\vw^1, \vw^2, \ldots, \vw^{n-1})$. As we will use a neural network
as a function approximator here,\footnote{
    Obviously, this does not have to be true, but at the end of the day, it is
    unclear if there is any parametric function approximation other than neural
    networks.
} 
these vectors will be multiplied with a weight matrix $\mE$. After this, we get
a sequence of continuous vectors $(\vp^1, \vp^2, \ldots, \vp^{n-1})$, where
\begin{align}
    \label{eq:nlm_first_layer}
    \vp^j = \mE^\top \vw^j
\end{align}
and $\mE \in \RR^{|V| \times d}$. 

Before continuing to build this function, let us see what it means to multiply
the transpose of a matrix with an one-hot vector from left. Since only one of
the elements of the one-hot vector is non-zero, all the rows of the matrix will
be ignored except for the row corresponding to the index of the non-zero element
of the one-hot vector. This row is multiplied by $1$, which simply gives us the
same row as the result of this whole matrix--vector multiplication. In short,
the multiplication of the transpose of a matrix with an one-hot vector is
equivalent to {\em slicing out a single row from the matrix}.

In other words, let 
\begin{align}
    \label{eq:word_emb}
    \mE = \left[ 
        \begin{array}{l}
            \ve_1 \\
            \ve_2 \\
            \vdots \\
            \ve_{|V|}
        \end{array}
    \right],
\end{align}
where $\ve_i \in \RR^d$. Then,
\begin{align*}
    \mE^\top \vw_i = \ve_i.
\end{align*}

This view has two consequences. First, in practice, it will be much more
efficient computationally to implement this multiplication as a simple table
look-up. For instance, in Python with NumPy, do
\begin{verbatim}
    p = E[i,:]
\end{verbatim}
instead of
\begin{verbatim}
    p = numpy.dot(E.T, w_i)
\end{verbatim}

Second, from this perspective, we can see each row of the matrix $\mE$ as a
continuous-space representation of a corresponding word. $\ve_i$ will be a
vector representation of the $i$-th word in the vocabulary $V$. This
representation is often called a {\it word embedding} and should reflect the
underlying meaning of the word. We will discuss this further shortly.

Closely following \cite{bengio2006neural}, we will simply concatenate the
continuous-space representations of the input words such that
\begin{align*}
    \vp = \left[ \vp^1; \vp^2; \ldots; \vp^{n-1}\right]^\top
\end{align*}
This vector $\vp$ is a representation of $n-1$ input words in a continuous
vector space and often referred to as a {\em context vector}.

This context vector is fed through a composition of nonlinear feature extraction
layers. We can for instance apply the simple transformation layer from
Eq.~\eqref{eq:layer} such that
\begin{align}
    \label{eq:nlm_context}
    \vh = \tanh(\mW \vp + \vb),
\end{align}
where $\mW$ and $\vb$ are the parameters. 

Once a set of nonlinear layers has been applied to the context vector, it's time
to compute the output probability distribution. In this case of language
modelling, the distribution outputted by the function is a categorical
distribution. We discussed how we can build a function to return a categorical
distribution already in Sec.~\ref{sec:other_dist}. 

As a recap, a categorical distribution defines a probability of one event
happening among $K$ discrete events. The probability of the $k$-th event
happening is often denoted as $\mu_k$, and
\begin{align*}
    \sum_{k=1}^K \mu_k = 1.
\end{align*}
Therefore, the function needs to return a $K$-dimensional vector $[\mu_1,\mu_2,
\ldots, \mu_K ]$. In this case of language modelling, $K=|V|$ and $\mu_i$
corresponds to the probability of the $i$-th word in the vocabulary for the next
word.

As discussed earlier in Sec.~\ref{sec:other_dist}, we can use softmax to compute
each of those output probabilities:
\begin{align}
    \label{eq:nlm_softmax}
    p(w_n = k | w_1, w_2, \ldots, w_{n-1}) = \mu_k = \frac{\exp(\vu_k^\top \vh +
    c_k)}{\sum_{k'=1}^{|V|} \exp(\vu_{k'}^\top \vh + c_{k'})},
\end{align}
where $\vu_k \in \RR^{\dim(\vh)}$. 

\begin{figure}[t]
    \centering
    \begin{minipage}[!b]{0.49\textwidth}
        \centering
        \includegraphics[width=0.9\columnwidth]{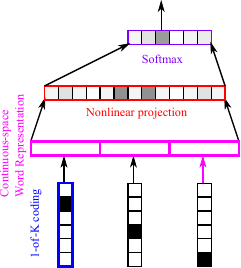}
    \end{minipage}
    \begin{minipage}[!b]{0.49\textwidth}
        \centering
        \includegraphics[width=0.9\columnwidth]{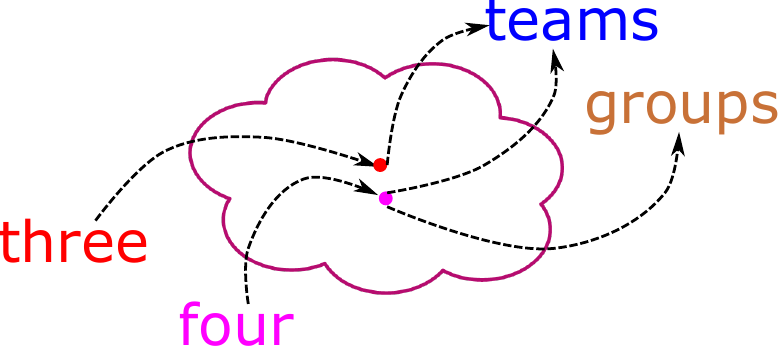}
    \end{minipage}
    \begin{minipage}[!b]{0.49\textwidth}
        \centering
        (a)
    \end{minipage}
    \begin{minipage}[!b]{0.49\textwidth}
        \centering
        (b)
    \end{minipage}
    \caption{(a) Schematics of neural language model. (b) Example of how neural language model generalizes to an unseen
    $n$-gram.}
    \label{fig:nlm}
\end{figure}

This whole function is called a neural language model.  See
Fig.~\ref{fig:nlm}~(a)
for the graphical illustration of neural language model.

\subsection{How does Neural Language Model Generalize to Unseen $n$-Grams? --
Distributional Hypothesis}
\label{sec:distributional}

Now that we have described neural language model, let us take a look into what
happens inside. Especially, we will focus on how the model generalizes to unseen
$n$-grams.

The previously described neural language model can be thought of as a composite
of two function ($g \circ f$). The first stage $f$ projects a sequence of
context words, or preceding $n-1$ words to a continuous vector space:
\begin{align*}
    f: \left\{ 0, 1\right\}^{|V| \times n-1} \to \RR^d
\end{align*}
We will call the resulting vector $\vh$ a {\em context vector}.
The second stage $g$ maps this continuous vector $\vh$ to the target word
probability, by applying affine transformation to the vector $\vh$ followed by
softmax normalization. 

Let us look more closely at what $g$ does in Eq.~\eqref{eq:nlm_softmax}. If we
ignore the effect of the bias $c_k$ for now, we can clearly see that the
probability of the $k$-th word in the vocabulary is large when the output vector
$\vu_k$ (or the $k$-th row of the output matrix $\mU$) is well aligned with the
context vector $\vh$. In other words, the probability of the next word being the
$k$-th word in the vocabulary is roughly proportional to the inner product
between the context vector $\vh$ and the corresponding target word vector
$\vu_k$.

Now let us consider two context vectors $\vh_j$ and $\vh_k$. These contexts are
followed by a similar set of words, meaning that the conditional distributions
of the next word are similar to each other. Although these distributions are
defined over all possibility target words, let us look at the probabilities of
only one of the target words $w_l$: 
\begin{align*}
    p_j^l =& p(w_l | \vh_j) = \frac{1}{Z_j} \exp\left( \vw_l^\top \vh_j \right), \\
    p_k^l =& p(w_l | \vh_k) = \frac{1}{Z_k} \exp\left( \vw_l^\top \vh_k \right).
\end{align*}
The ratio between $p_j^l$ and $p_k^l$ is then\footnote{
    Note that both $p_j^l$ and $p_k^l$ are positive due to our use of {\em softmax}.
}
\begin{align*}
    \frac{p_j^l}{p_k^l} = \frac{Z_k}{Z_j} \exp\left( \vw_l^\top (\vh_j -
    \vh_k)\right).
\end{align*}
From this, we can clearly see that in order for the ratio $\tfrac{p_j^l}{p_k^l}$
to be $1$, i.e., $p_j^l = p_k^l$, 
\begin{align}
    \label{eq:diff}
    \vw_l^\top \left(\vh_j - \vh_k\right) = 0.
\end{align}

Now let us assume that $\vw_l$ is not an all-zero vector, as otherwise it will
be too dull a case. In this case, the way to achieve the equality in
Eq.~\eqref{eq:diff} is to drive the context vectors $\vh_j$ and $\vh_k$ to each
other. In other words, the context vectors must be similar to each other (in
terms of Euclidean distance) in order to result in similar conditional
distributions of the next word.

What does this mean? This means that the neural language model must project
$(n-1)$-grams that are followed by the same word to nearby points in the context
vector space, while keeping the other $n$-grams away from that neighbourhood.
This is necessary in order to give a similar probability to the same word. If
two $(n-1)$-grams, which are followed by the same word in the training corpus,
are projected to far away points in the context vector space, it naturally
follows from this argument that the probability over the next word will differ
substantially, resulting in a bad language model.

Let us consider an extreme example, where we do bigram modeling with the
training corpus comprising only three sentences:
\begin{itemize}
    \itemsep 0em
    \item There are {\bf three teams} left for the qualification.
    \item {\bf four teams} have passed the first round.
    \item {\bf four groups} are playing in the field.
\end{itemize}
We will focus on the bold-faced phrases; ``three teams'', ``four teams'' and
``four group''. The first word of each of these bigrams is a context word, and
neural language model is asked to compute the probability of the word following
the context word.

It is important to notice that neural language model {\em must} project
``three'' and ``four'' to nearby points in the context space (see
Eq.~\eqref{eq:nlm_context}.) This is because the context vectors from these two
words need to give a similar probability to the word ``teams''. This naturally
follows from our discussion earlier on how dot product preserves the ordering in
the space. And, from these two context vectors (which are close to each other),
the model assigns similar probabilities to ``teams'' and ``groups'', because
they occur in the training corpus. In other words, the target word vector
$\vu_{\text{teams}}$ and $\vu_{\text{groups}}$ will also be similar to each
other, because otherwise the probability of ``teams'' given ``four''
($p(\text{teams} | \text{four})$) and
``groups'' given ``four'' ($p(\text{groups} | \text{four})$) will be very
different despite the fact that they occurred equally likely in the training
corpus.

Now, let's assume the case where we use the neural language model trained on
this tiny training corpus to assign a probability to an unseen bigram ``three
groups''. The neural language model will project the context word ``three'' to a
point in the context space close to the point of ``four''. From this context
vector, the neural language model will {\em have to} assign a high probability
to the word ``groups'', because the context vector $\vh_{\text{three}}$ and the
target word vector $\vu_{\text{groups}}$ well align. Thereby, even without ever
seeing the bigram ``three groups'', the neural language model can assign a
reasonable probability. See Fig.~\ref{fig:nlm}~(b) for graphical
illustration.

What this example shows is that neural language model automatically learns the
similarity among different context words (via context vectors $\vh$), and also among
different target words (via target word vectors $\vu_k$), by exploiting {\em
co-occurrences} of words. In this example, the neural language model learned
that ``four'' and ``three'' are similar from the fact that both of them occur
together with ``teams''. Similarly, in the target side, the neural language
model was able to capture the similarity between ``teams'' and ``groups'' by
noticing that they both follow a common word ``four''.

This is a clear, real-world demonstration of the so-called distributional
hypothesis. Distributional hypothesis states that ``words which are similar in
meaning appear in similar distributional contexts'' \cite{firth1957}. By
observing which words a given word co-occurs together, it is possible to peek
into the word's underlying meaning. Of course, this is only a partial
picture\footnote{
    We will discuss why this is only a partial picture later on.
}
into the underlying meaning of each word, or as a matter of fact a phrase, but
surely still a very interesting property that is being naturally exploited by
neural language model.

In neural language model, the most direct way to observe the effect of this
distributional hypothesis/structure is to investigate the first layer's weight
matrix $\mE$ in Eq.~\eqref{eq:word_emb}. This weight matrix can be considered as
a set of dense vectors of the words in the input vocabulary $\left\{ \ve_1,
\ve_2, \ldots, \ve_{|V|}\right\}$, and any visualization technique, such as
principal component analysis (PCA) or t-SNE~\cite{VanDerMaaten08}, can be used
to project each high-dimensional word vector into a lower-dimensional space
(often 2-D or 3-D).

\subsection{Continuous Bag-of-Words Language Model: \\ Maximum Pseudo--Likelihood
Approach}
\label{sec:cbow}

This is about time someone asks a question why we are only considering the {\em
preceding} words when doing language modelling. Is it a good assumption that the
conditional distribution over a word is only dependent on preceding words? 

In fact, we do not have to do so. We can certainly model a natural language
sentence such that each word in a sentence is conditioned on $2n$ surrounding
words ($n$ words to the left and $n$ words to the right.) In this case, we get a
{\em Markov random field (MRF) language model}~\cite{jernite2015fast}.

\begin{figure}[ht]
    \centering
    \includegraphics[width=\textwidth]{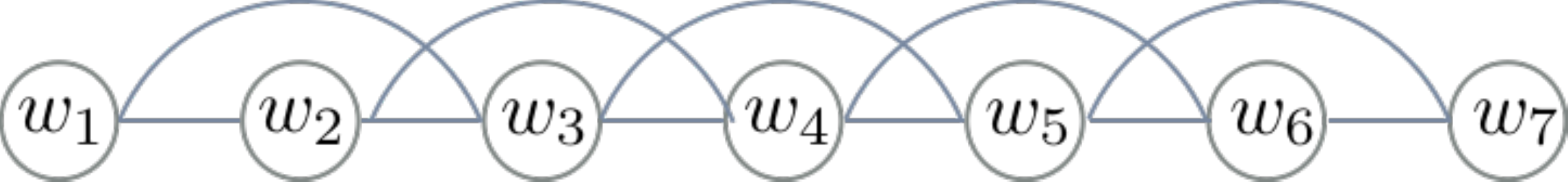}
    \caption{An example Markov random field language model (MRF-LM) with the
    order $n=1$.}
    \label{fig:mrf_lm}
\end{figure}

In a Markov random field (MRF) language model (MRF-LM), we say each word in a
given sentence is a random variable $w_i$. We connect each word with its $2n$
surrounding words with {\em undirected} edges, and these edges represent the
conditional dependency structure of the whole MRF-LM. An example of an MRF-LM
with $n=1$ is shown in Fig.~\ref{fig:mrf_lm}.

A probability over a Markov random field is defined as a product of clique
potentials. A potential is defined for each clique as a positive function whose
input is the values of the random variables in the clique. In the case of
MRF-LM, we will assign $1$ as a potential to every clique except for cliques of
two random variables (in other words, we only use pairwise potentials only.) The
pairwise potential between the words $i$ and $j$ is defined as
\begin{align*}
    \phi(\vw^i, \vw^j) = \exp\left( (\mE^\top \vw^{i})^\top  \mE^\top \vw^j\right) = 
    \exp\left( \ve_{w^i}^\top \ve_{w^j} \right),
\end{align*}
where $\mE$ is from Eq.~\eqref{eq:word_emb}, and $\vw^i$ is the one-hot
vector of the $i$-th word. One must note that this is one possible
implementation of the pairwise potential, and there may be other possibilities,
such as to replace the dot product between the word vectors ($\ve_{w_i}^\top
\ve_{w_j}$) with a deeper network.

With this pairwise potential, the probability over the whole sentence is defined
as
\begin{align*}
    p(w_1, w_2, \ldots, w_T) = \frac{1}{Z} \prod_{t=1}^{T-n} \prod_{j=t}^{t+n}
    \phi(\vw^t, \vw^j) = \frac{1}{Z} \exp\left( 
        \sum_{t=1}^{T-n} \ve_{w^t}^\top \ve_{w^j}
    \right),
\end{align*}
where $Z$ is the normalization constant. This normalization constant makes the
product of the potentials to be a probability and often is at the core of
computational intractability in Markov random fields.

\begin{figure}[ht]
    \centering
    \includegraphics[width=\textwidth]{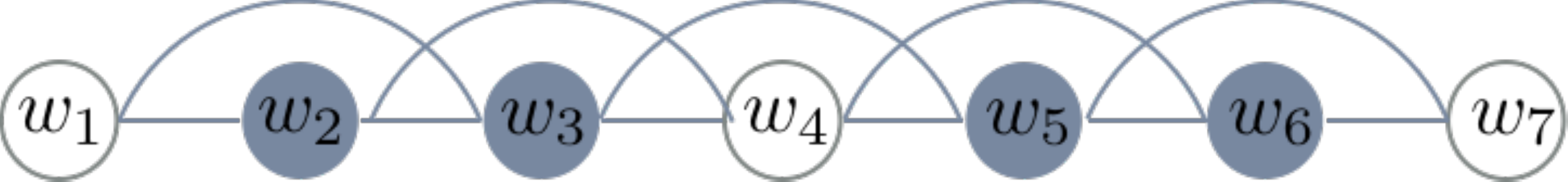}
    \caption{Gray nodes indicate the Markov blank of the fourth word.}
    \label{fig:mrf_lm_mblanket}
\end{figure}

Although compute the full sentence probability is intractable in this MRF-LM, it
is quite straightforward to compute the conditional probability of each word
$w^i$ given all the other words. When computing the conditional probability, we
must first notice that the conditional probability of $w^i$ {\em only} depends
on the values of other words included in its {\em Markov blanket.} In the case
of Markov random fields, the Markov blanket of a random variable is defined as a
set of all {\em immediate neighbours}, and it implies that the conditional
probability of $w^i$ is dependent only on $n$ preceding words and the $n$
following words. See Fig.~\ref{fig:mrf_lm_mblanket} for an example.

Keeping this in mind, we can easily see that
\begin{align*}
    p(w^i|w^{i-n}, \ldots, w^{i-1}, w^{i+1}, \ldots, w^{i+n})
    = \frac{1}{Z'} \exp\left(\ve_{w^i}^\top \left( 
            \sum_{k=1}^n 
            \ve_{w^{i-k}}
            +
            \sum_{k=1}^n 
            \ve_{w^{i+k}}
    \right)\right),
\end{align*}
where $Z'$ is a normalization constant computed by
\begin{align*}
    Z' = \sum_{v \in V} \exp\left( \ve_v^\top \left( 
            \sum_{k=1}^n 
            \ve_{w^{i-k}}
            +
            \sum_{k=1}^n 
            \ve_{w^{i+k}}
    \right)\right).
\end{align*}

\begin{figure}[ht]
    \centering
    \includegraphics[width=0.5\textwidth]{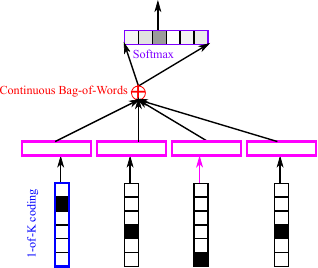}
    \caption{Continuous Bag-of-Words model approximates the conditional
    distribution over the $j$-th word $w_j$ under the MRF-LM.}
    \label{fig:cbow}
\end{figure}

Do you see a stark similarity to neural language model we discussed earlier?
This conditional probability is a shallow neural network with a single {\em
linear} hidden layer whose input are the context words ($n$ preceding and $n$
following words) and the output is the conditional distribution of the center
word $w_i$. We will talk about this shortly in more depth. See
Fig.~\ref{fig:cbow} for graphical illustration.

Now we know that it is often difficult to compute the full sentence probability
$p(w^1, \ldots, w^T)$ due to the intractable normalization constant $Z$. We
however know how to compute the conditional probabilities (for all words) quite
tractably. The former fact implies that it is perhaps not the best idea to
maximize log-likelihood to train this model.\footnote{
    However this is not to say maximum likelihood in this case is impossible.
    There are different ways to approximate the full sentence probability under
    this model. See \cite{jernite2015fast} for one such approach.
} The latter however sheds a bit of light, because we can train a model to
maximize pseudo--likelihood \cite{besag1975statistical} instead.\footnote{
    See the note by Amir Globerson (later modified by David Sontag) available at
    \url{http://cs.nyu.edu/~dsontag/courses/inference14/slides/pseudolikelihood_notes.pdf}.
}

Pseudo--likelihood of the MRF-LM is defined as
\begin{align}
    \label{eq:pl_cbow}
    \log \PL = \sum_{i=1}^T \log p(w^i | w^{i-n}, \ldots, w^{i-1}, w^{i+1}, \ldots,
    w^{i+n}).
\end{align}
Maximizing this pseudo--likelihood is equivalent to training a neural network in
Fig.~\ref{fig:cbow} which approximates each conditional distribution $p(w^i |
w^{i-n}, \ldots, w^{i-1}, w^{i+1}, \ldots, w^{i+n})$ to give a higher
probability to the ground-truth center word in the training corpus.

Unfortunately, even after training the model by maximizing the pseudo--likelihood
in Eq.~\eqref{eq:pl_cbow}, we do not have a good way to compute the full
sentence probability under this model. Under certain conditions maximizing
pseudo--likelihood indeed converges to the maximum likelihood solution, but this
does not mean that we can use the product of all the conditionals as a
replacement of the full sentence probability. However, this does not mean that
we cannot use this MRF-LM as a language model, since given a fixed model, the
pseudo--probability (the product of all the conditionals) can score different
sentences. 

This is in contrast to the neural language model we discussed earlier in
Sec.~\ref{sec:nlm}. In the case of neural language model, we were able to
compute the probability of a given sentence by computing the conditional
probability of each word, reading from left until the end of the sentence. This
is perhaps one of the reasons why the MRF-LM is not often used in practice as a
language model. Then, you must ask why I even bothered to explain this MRF-LM in
the first place.

This approach, which was proposed in \cite{mikolov2013efficient} as a
continuous bag-of-words (CBoW) model,\footnote{
    One difference between the model we derived in this section starting from
    the MRF-LM and the one proposed in \cite{mikolov2013efficient} is that in
    our derivation, the neural network shares a single weight matrix $\mE$ for
    both the input and output layers.
} was found to exhibit an interesting property. That is, the word embedding
matrix $\mE$ learned as a part of this CBoW model very well reflects underlying
structures of words, and this has become one of the darling models by natural
language processing researchers in recent years. We will discuss further in the
next section.

\paragraph{Skip-Gram and Implicit Matrix Factorization}

In \cite{mikolov2013efficient}, another model, called skip-gram, is proposed.
The skip-gram model is built by flipping the continuous bag-of-words model.
Instead of trying to predict the middle word given $2n$ surrounding words, the
skip-gram model tries to predict randomly chosen one of the $2n$ surrounding
words given the middle word. From this description alone, it is quite clear that
this skip-gram model is not going to be great as a language model. However, it
turned out that the word vectors obtained by training a skip-gram model were as
good as those obtained by either a continuous bag-of-words model or any other
neural language model. Of course, it is debatable which criterion be used to
determine the goodness of word vectors, but in many of the existing so-called
``intrinsic'' evaluations, those obtained from a skip-gram model have been shown
to excel.

The authors of \cite{Omer2014} recently showed that training a skip-gram model
with negative sampling (see \cite{mikolov2013efficient}) is equivalent to
factorizing a positive point-wise mutual information matrix (PPMI) into two
lower-dimensional matrices. The left lower-dimensional matrix corresponds to the
input word embedding matrix $\mE$ in a skip-gram model. In other words, training
a skip-gram model {\em implicitly} factorizes a PPMI matrix. 

Their work drew a nice connection between the existing works on distributional word
representations from natural language processing, or even computational
linguistics and these more recent neural approaches. I will not go into any
further detail in this course, but I encourage readers to read \cite{Omer2014}.

\subsection{Semi-Supervised Learning with Pretrained Word Embeddings}
\label{sec:semi_emb}

One thing I want to emphasize in these language models, including $n$-gram
language model, neural language model and continuous bag-of-words model, is that
they are purely {\em unsupervised}, meaning that all we need is a large corpus
of unannotated text. This is one thing that makes this statistical approach to
language modelling much more appealing than any other approach based on
linguistic structures (see Sec.~\ref{sec:linguistic_lm} for a brief discussion.) 

When it comes to neural language model and continuous bag-of-words model, we now
know that these networks learn continuous vector representations of input words,
target words and the context phrase ($\vh$ from Eq.~\eqref{eq:nlm_context}.) We
also discussed how these vector representations encode similarities among
different linguistic units, be it a word or a phrase. 

What this implies is that once we train this type of language model on a large,
or effectively infinite,\footnote{
    Why? Because of almost universal broadband access to the Internet!
}
corpus of {\em unlabelled} text, we get good vectors for those linguistic units
for free. Among these, word vectors, the rows of the input weight matrix $\mE$
in Eq.~\eqref{eq:word_emb}, have been extensively used in many natural language
processing applications in recent years since
\cite{turian2010word,collobert2011natural,mikolov2013efficient}.

Let us consider an extreme example of classifying each English word as either
``positive'' or ``negative''. For instance, ``happy'' is positive, while ``sad''
is negative. A training set of 2 examples--1 positive and 1 negative words-- is
given. How would one build a classifier?\footnote{
    Although the setting of 2 training examples is extreme, but the task itself
    turned out to be not-so-extreme. In fact, there is multiple dictionaries of
    words' sentiment maintained. For instance, check
    \url{http://sentiwordnet.isti.cnr.it/search.php?q=llama}.
}

There are two issues here. First, it is unclear how we should represent the
input, in this case a word. A good reader who has read this note so far will be
clearly ready to use an one-hot vector and use a softmax layer in the output,
and I commend you for that. However, this still does not solve a more serious
issue which is that we have only {\em two} training examples!  All the word
vectors, save for two vectors corresponding to the words in the training set,
will not be updated at all.

One way to overcome these two issues is to make somewhat strong, but reasonable
assumption that {\em similar} input will have similar sentiments. This
assumption is at the heart of semi-supervised learning~\cite{Chapelle2006}. It
says that high-dimensional data points in effect lies on a lower-dimensional
manifold, and the target values of the points on this manifold change smoothly.
Under this assumption, if we can well model this lower-dimensional data manifold
using unlabelled training examples, we can train a good classifier\footnote{
    What do I mean by a good classifier? A good classifier is a classifier that
    classifies {\em unseen} test examples well. See
    Sec.~\ref{sec:model_selection}.
} 

And, guess what? We have access to this lower-dimensional manifold, which is
represented by the set of {\em pretrained} word vectors $\mE$. Believing that
similar words have similar sentiment and that these pretrained word vectors
indeed well reflect similarities among words, let me build a simple nearest
neighbour (NN) classifier which uses the pretrained word vectors:
\begin{align*}
    \text{NN}(w) = \left\{ \begin{array}{l l}
            \text{positive},&\text{ if } \text{cos}(\ve_w,\ve_{\text{happy}}) > 
            \text{cos}(\ve_w,\ve_{\text{bad}}) \\
            \text{negative},&\text{ otherwise}
        \end{array}
    \right.,
\end{align*}
where $\text{cos}(\cdot, \cdot)$ is a cosine similarity defined as
\begin{align*}
    \text{cos}(\ve_i, \ve_j) = \frac{\ve_i^\top \ve_j}{\|\ve_i\| \|\ve_j\|}.
\end{align*}

This use of a term ``similarity'' almost makes this set of pretrained word
vectors look like some kind of magical wand that can solve everything.\footnote{
    For future reference, I must say there were many papers claiming that the
    pretrained word vectors are indeed magic wands at three top-tier natural
    language processing conferences (ACL, EMNLP, NAACL) in 2014 and 2015.
} This is however not true, and using pretrained word vectors must be done with
caution. 

Why should we be careful in using these pretrained word vectors? We must
remember that these word vectors were obtained by training a neural network to
maximize a certain objective, or to minimize a certain cost function. This means
that these word vectors capture certain aspects of words' underlying structures
that are necessary to achieve the training objective, and that there is no
reason for these word vectors to capture any other properties of the words that
are not necessary for maximizing the training objective.  In other words,
``similarity'' among multiple words has many different aspects, and these word
vectors will capture only a few of these many aspects. Which few aspects will be
determined by the choice of training objective.

The hope is that language modelling is a good training objective that will
encourage the word vectors to capture as many aspects of similarity as
possible.\footnote{
    Some may ask how a single vector, which is a point in a space, can capture
    multiple aspects of similarity. This is possible because these word vectors
    are high-dimensional.
} But, is this true in general? 

Let's consider an example of words describing emotions, such as ``happy'',
``sad'' and ``angry'', in the context of a continuous bag-of-words model. These
emotion-describing words often follow some forms of a verb ``feel'', such as
``feel'', ``feels'', ``felt'' and ``feeling''. This means that those
emotion-describing words will have to be projected nearby in the context space
in order to give a high probability to those forms of ``feel'' as a middle word.
This is understandable and agrees quite well with our intuition. All those
emotion-describing words are similar to each other in the sense that they all
describe emotion. But, wait, this aspect of similarity is not going to help
sentiment classification of words. In fact, this aspect of similarity will {\em
hurt} the sentiment classifier, because a positive word ``happy'' will be close
to negative words ``sad'' and ``angry'' in this word vector space!

The lesson here is that when you are solving a language-related task with very
little data, it is a good idea to consider using a set of pretrained word
vectors from neural language models. However, you must do so in caution, and
perhaps try to pretrain your own word vectors by training a neural network to
maximize a certain objective that better suits your final task.

But, then, what other training objectives are there? We will get to that later.

\section{Recurrent Language Model}
\label{sec:rlm}

Neural language model indeed avoids the lack of generalization in the
conventional $n$-gram language modeling. It still assumes the $n$-th order
Markov property, meaning that it looks only as far back into the past as $n-1$
words. In Sec.~\ref{sec:n_gram_lm}, I gave an example of ``In Korea, more than
half of all the residents speak Korean''. In this example, the conditional
distribution over the last word in the sentence clearly will be better estimated
if it is conditioned on the second word of the sentence which is more than 10
words back in the past.

Let us recall what we learned in Sec.~\ref{sec:rnn_x_y}. There, we learn how to
build a recurrent neural network to read a variable-length sequence and return a
variable-length output sequence. An example we considered back then was a task
of part-of-speech tagging, where the input is a sentence such as
\begin{align*}
    x = (\text{Children}, \text{eat}, \text{sweet}, \text{candy}),
\end{align*}
and the target output is a sequence of part-of-speech tags such as
\begin{align*}
    y = (\text{noun}, \text{verb}, \text{adjective}, \text{noun}).
\end{align*}

\begin{figure}[t]
    \centering
    \begin{minipage}{0.44\textwidth}
        \centering
        \includegraphics[width=0.95\columnwidth]{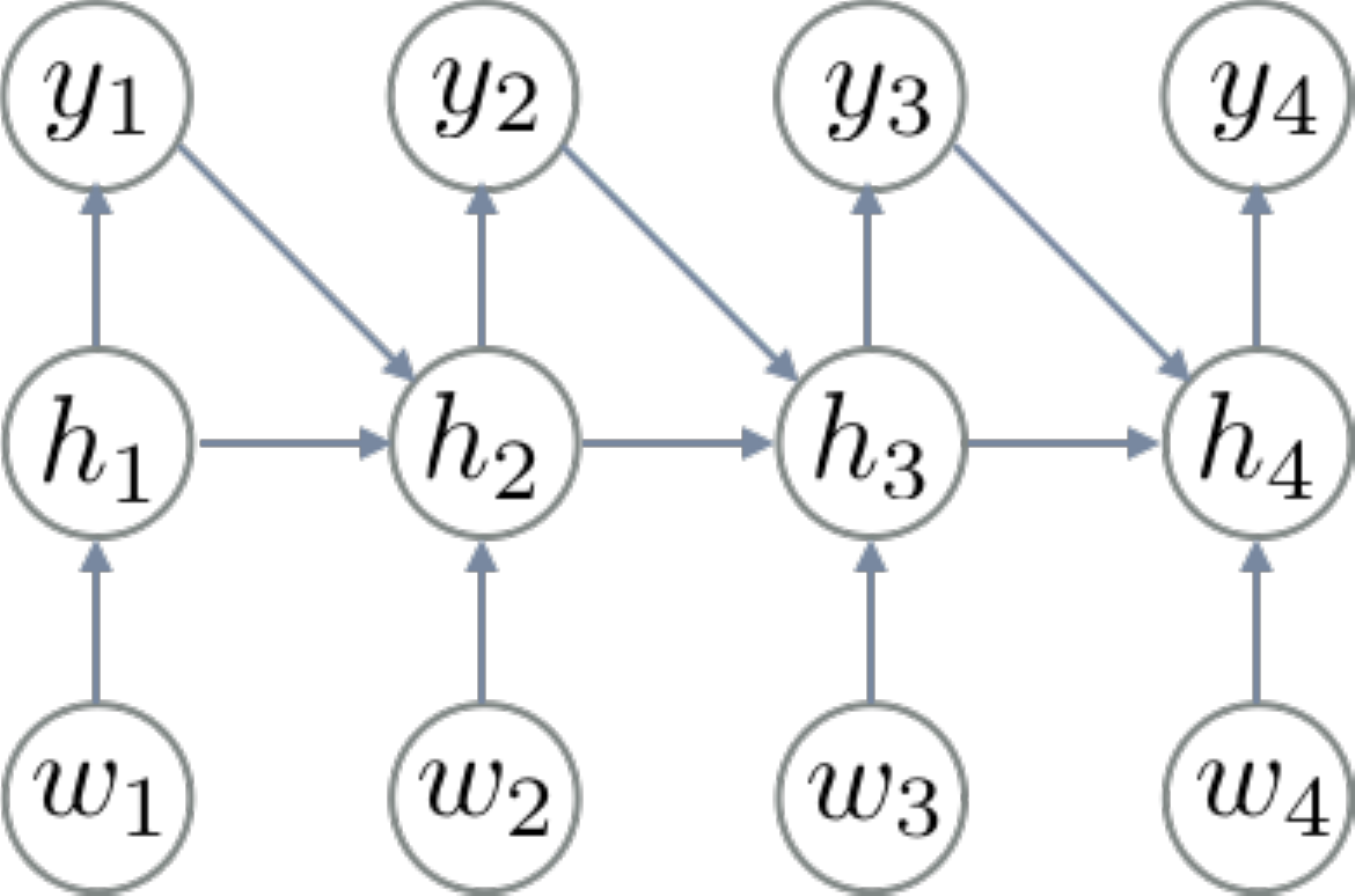}
    \end{minipage}
    \hfill
    \begin{minipage}{0.55\textwidth}
        \centering
        \includegraphics[width=0.95\columnwidth]{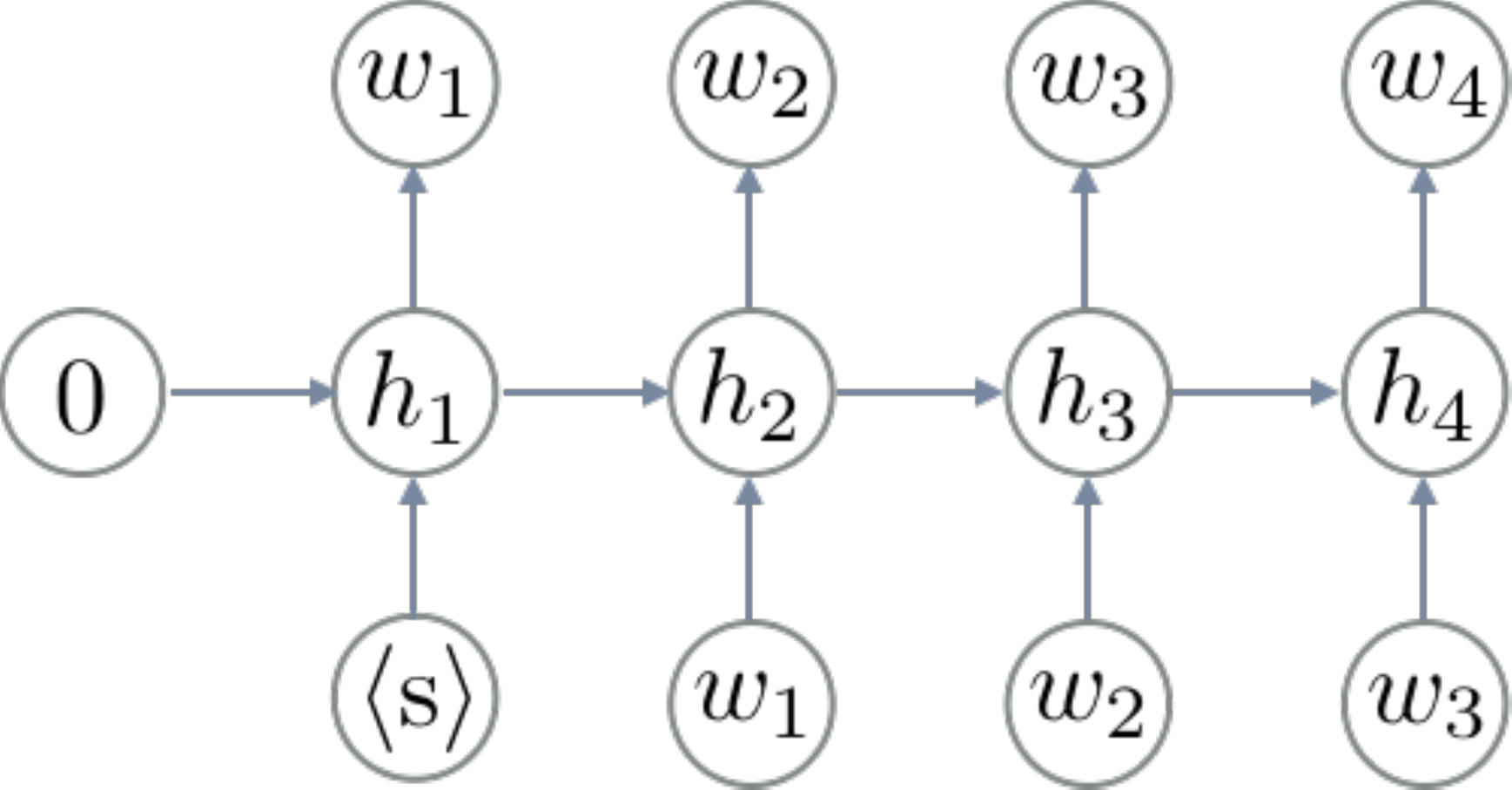}
    \end{minipage}

    \begin{minipage}{0.44\textwidth}
        \centering
        (a)
    \end{minipage}
    \hfill
    \begin{minipage}{0.55\textwidth}
        \centering
        (b)
    \end{minipage}

    \caption{
        (a) A recurrent neural network from Sec.~\ref{sec:rnn_x_y}. (b) A
        recurrent neural network language model.
    }
    \label{fig:rnn_lm}
\end{figure}

In order to make less of an assumption on the conditional independence of the
predicted tags, we made a small adjustment such that the prediction $Y_{t}$ at
each timestep was fed back into the recurrent neural network in the next
timestep together with the input $X_{t+1}$. See Fig.~\ref{fig:rnn_lm}~(a) for
graphical illustration.

Why am I talking about this again, after saying that the task of part-of-speech
tagging is not even going to be considered as a valid topic for the final
project? Because the very same model for part-of-speech tagging will be turned
into the very {\em recurrent neural network language model} in this section.

Let us start by considering a single conditional distribution, marked (a) below,
from the full sentence probability:
\begin{align*}
    p(w^1, w^2, \ldots, w^T) = \prod_{t=1}^T \underbrace{p(w^t|w^1, \ldots,
    w^{t-1})}_{(a)}.
\end{align*}
This conditional probability can be approximated by a neural network, as we've
been doing over and over again throughout this course, that takes as input
$(w^1, \ldots, w^{t-1})$ and returns the probability over all possible words in
the vocabulary $V$.  This is not unlike neural language model we discussed
earlier in Sec.~\ref{sec:nlm}, except that the input is now a variable-length
sequence.

In this case, we can use a recurrent neural network which is capable of
summarizing/memorizing a variable-length input sequence.  A recurrent neural
network summarizes a given input sequence $(w^1, \ldots, w^{t-1})$ into a memory
state $\vh^{t-1}$:
\begin{align}
    \label{eq:rnnlm_iteration}
    \vh^{t'} = \left\{
        \begin{array}{l l}
            0,&\text{ if }t'=0 \\
            f(\ve_{w^{t'}}, \vh^{t'-1}),&\text{ otherwise}
        \end{array}
        \right.,
\end{align}
where $t'$ runs from $0$ to $t-1$.  $f$ is a recurrent function which can be any
of a naive transition function from Eq.~\eqref{eq:rnn_layer}, a gated recurrent
unit or a long short-term memory unit from Sec.~\ref{sec:gru}. $\ve_{w^{t'}}$ is
a word vector corresponding to the word $w^{t'}$.

This summary $\vh^{t-1}$ is affine-transformed followed by a softmax nonlinear
function to compute the conditional probability of $w^t$. Hopefully, everyone
remembers how it is done. As in Eq.~\eqref{eq:rnn_y_h}, 
\begin{align*}
    \vmu = \softmax(\mV \vh^{t-1}),
\end{align*}
where $\vmu$ is a vector of probabilities of all the words in the vocabulary.

One thing to notice here is that the iteration procedure in
Eq.~\eqref{eq:rnnlm_iteration} computes a sequence of every memory state vector
$\vh^t$ by simply reading the input sentence {\em once}. In other words, we can
let the recurrent neural network read one word $w^t$ at a time, update the
memory state $\vh^t$ and compute the conditional probability of the {\em next}
word $p(w^{t+1}|w^{\leq t})$. 

This procedure is illustrated in Fig.~\ref{fig:rnn_lm}~(b).\footnote{
    In the figure, you should notice the beginning-of-the-sentence symbol
    $\left< \text{s} \right>$. This is necessary in order to use the very same
    recurrent function $f$ to compute the conditional probability of the first
    word in the input sentence.
}
This language model is called a {\em recurrent neural network language
model} (RNN-LM, \cite{mikolov2010recurrent}). 

\begin{figure}[t]
    \centering
    \includegraphics[width=0.6\textwidth]{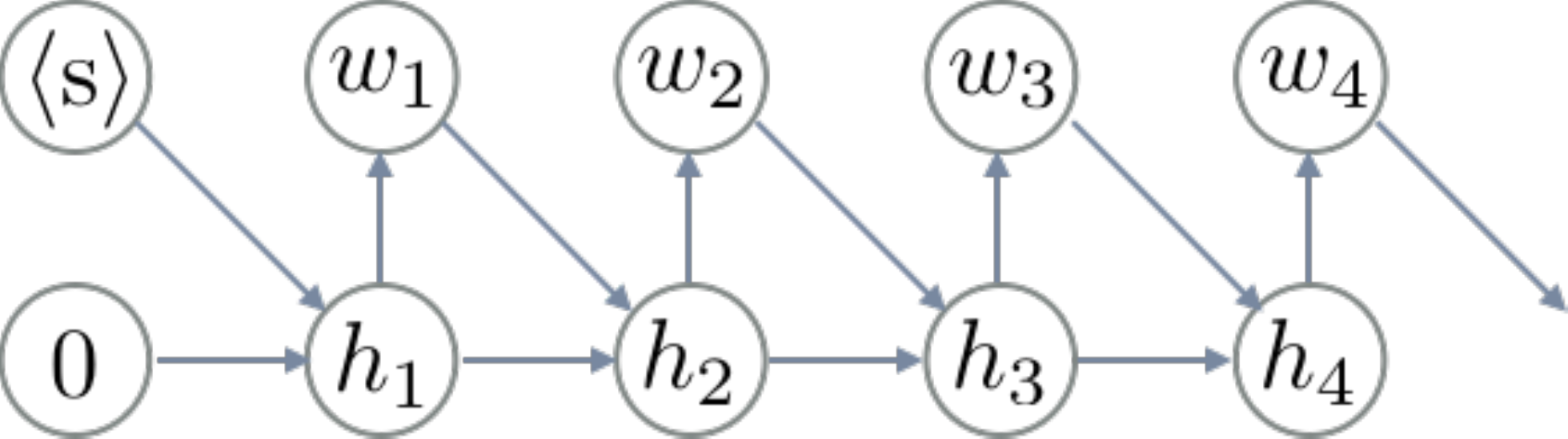}

    \caption{
        A recurrent neural network language model
    }
    \label{fig:rnn_lm2}
\end{figure}

But, wait, from looking at Figs.~\ref{fig:rnn_lm}~(a)--(b), there is a clear
difference between the recurrent neural networks for part-of-speech tagging and
language model. That is, there is no feedback connection from the output of the
previous time step back into the recurrent neural network in the RNN-LM.  This
is simply an illusion from the limitation in the graphical illustration, because
the input $w^{t+1}$ in the next time step is in fact the output $w^{t+1}$ at the
current time step. This becomes clearer by drawing the same figure in a slightly
different way, as in Fig.~\ref{fig:rnn_lm2}.

\section{How do $n$-gram language model, neural language model and RNN-LM
compare?}

Now the question is which one of these language models we should use in practice.
In order to answer this, we must first discuss the metric most commonly used for
evaluating language models.

The most commonly used metric is a {\em perplexity}. In the context of language
modelling, the perplexity $\PPL$ of a model $\MM$ is computed by
\begin{align}
    \label{eq:ppl}
    \PPL = {b}^{-\frac{1}{N} \sum_{n=1}^N \log_{b} p_{\MM}(w_n | w_{<n})},
\end{align}
where $N$ is the number of all the words in the validation/test corpus, and $b$
is some constant that is often $2$ or $10$ in practice. 

What is this perplexed metric? I totally agree with you on this one. Of course,
there is a quite well principled way to explain what this perplexity is based on
information theory. This is however not necessary for us to understand this
metric called perplexity. 

As the exponential function (with base $b$ in the case of perplexity in
Eq.~\eqref{eq:ppl}) is a monotonically increasing function, we see that the
ordering of different language models based on the perplexity will not change
even if we only consider the exponent: 
\begin{align*}
    -\frac{1}{N} \sum_{n=1}^N \log_{b} p_{\MM}(w_n | w_{<n}).
\end{align*}
Furthermore, assuming that $b > 1$, we can simply replace $\log_b$ with $\log$
(natural logarithm) without changing the order of different language models:
\begin{align*}
    -\frac{1}{N} \sum_{n=1}^N \log p_{\MM}(w_n | w_{<n}).
\end{align*}
Now, this looks awfully similar to the cost function, or negative
log-likelihood, we minimize in order to train a neural network (see
Chapter~\ref{chap:function_approx}.)

Let's take a look at a single term inside the summation above:
\begin{align*}
    \log p_{\MM}(w_n | w_{<n}).
\end{align*}
This is simply measuring how high a probability the language model $\MM$ is
assigning to a {\em correct next word} given all the previous words. Again,
because $\log$ is a monotonically increasing function.

In summary, the (inverse) perplexity measures how high a probability the
language model $\MM$ assigns to correct next words in the test/validation corpus
{\em on average}.  Therefore, {\em a better language model is the one with a lower
perplexity}.  There is nothing so perplexing about the perplexity, once we start
viewing it from this perspective.

We are now ready to compare different language models, or to be more precise,
three different classes of language models--count-based $n$-gram language model,
neural $n$-gram language model and recurrent neural network language model. The
biggest challenge in doing so is that this comparison will depend on many
factors that are not easy to control. To list a few of them,
\begin{itemize}
    \itemsep 0em
    \item Language
    \item Genre/Topic of training, validation and test corpora
    \item Size of a training corpus
    \item Size of a language model
\end{itemize}

\begin{figure}[ht]
    \centering
    \includegraphics[width=0.6\textwidth]{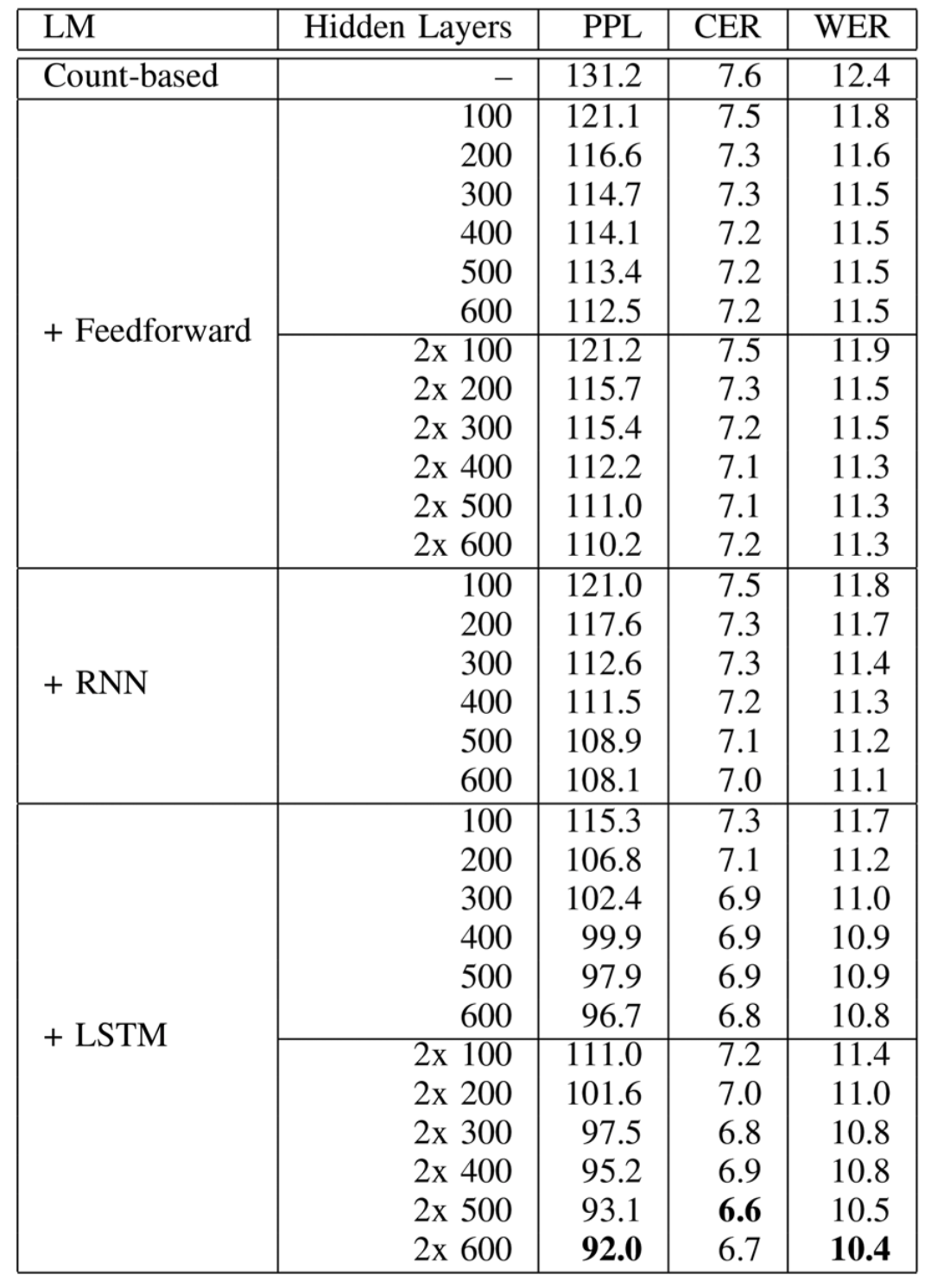}

    \caption{
        The perplexity, word error rate (WER) and character error rate (CER) of
        an automatic speech recognition system using different language models.
        Note that all the results by neural or recurrent language models are by
        interpolating these models with the count-based $n$-gram language model.
        Reprinted from \cite{sundermeyer2015feedforward}.
    }
    \label{fig:sundermeyer1}
\end{figure}

Because of this difficulty, this kind of comparison has often been done in the
context of a specific downstream application. This choice of a downstream
application often puts rough constraints on the size of available, or commonly
used, corpus, target language and reasonably accepted size of language models.
For instance, the authors of \cite{baltescu2014pragmatic} compared the
conventional $n$-gram language model and neural language model, with various
approximation techniques, with machine translation as a final task. In
\cite{sundermeyer2015feedforward}, the authors compared all the three classes of
language model in the context of automatic speech recognition.

\begin{figure}[ht]
    \centering
    \includegraphics[width=0.6\textwidth]{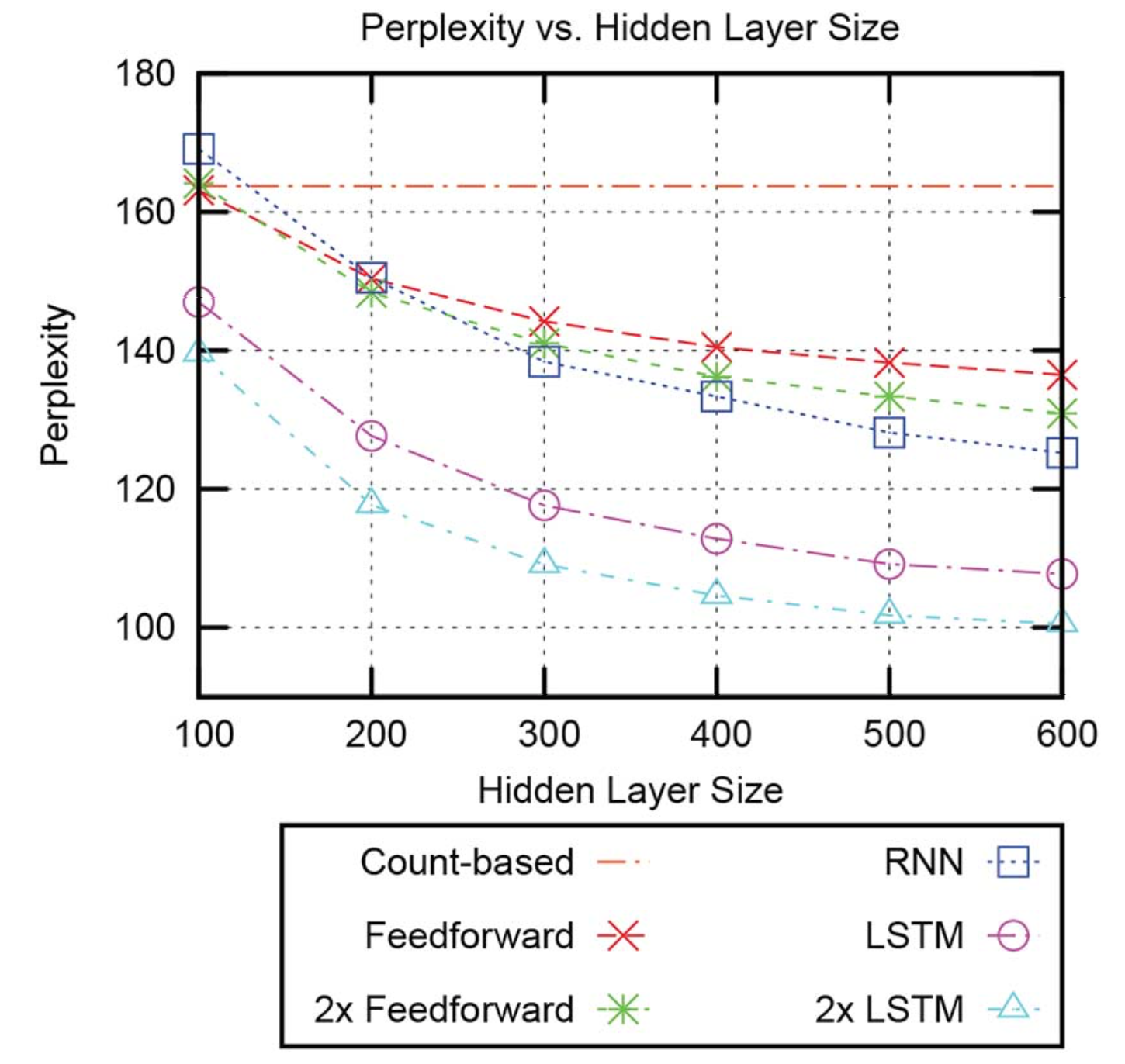}

    \caption{
        The trend of perplexity as the size of language model changes.
        Reprinted from \cite{sundermeyer2015feedforward}.
    }
    \label{fig:sundermeyer2}
\end{figure}

First, let us look at one observation made in \cite{sundermeyer2015feedforward}.
From Fig.~\ref{fig:sundermeyer1}, we can see that it is beneficial to use a
recurrent neural network language model (RNN-LM) compared to a usual neural
language model. Especially when long short-term memory units were used, the
improvement over the neural language model was significant. Furthermore, we see
that it is possible to improve these language models by simply increasing their
size. 

Similarly, in Fig.~\ref{fig:sundermeyer2} from the same paper
\cite{sundermeyer2015feedforward}, it is observed that larger language models
tend to get better/lower perplexity and that RNN-LM in general outperforms
neural language models. 

These two observations do seem to suggest that neural and recurrent language
models are better candidates as language model. However, this is not to be taken
as an evidence for choosing neural or recurrent language models. It has been
numerously observed over years that the best performance, both in terms of
perplexity and in terms of performance in the downstream applications such as
machine translation and automatic speech recognition, is achieved by combining
a count-based $n$-gram language model and a neural, or recurrent, language
model. See, for instance, \cite{schwenk2007continuous}.

This superiority of combined, or hybrid, language model suggests that the
count-based, or conventional, $n$-gram language model, neural language model and
recurrent neural network language model are capturing underlying structures of
natural language sentences that are complement to each other. However, it is not
crystal clear how these captured structures differ from each other.

\chapter{Neural Machine Translation}
\label{chap:nmt}

Finally, we have come to the point in this course where we discuss an {\em
actual} natural language task. In this chapter, we will discuss how translation
from one language to another can be done with statistical methods, more
specifically neural networks.

\section{Statistical Approach to Machine Translation}

Let's first think of what it means to translate one sentence $X$ in a source
language to an equivalent sentence $Y$ in a target language which is different
from the source language. A process of translation is a function that takes as
input the source sentence $X$ and returns a correct translation $Y$, and it is
clear that there may be more than one correct translations. The latter fact
implies that this function of translation should return not a single correct
translation, but a probability distribution that assigns high probabilities to
more than one likely translations.

Now, let us write it in a more formal way. First, the input is a sequence of
words
\[
    X = (x_1, x_2, \ldots, x_{T_x}),
\]
where $T_x$ is the length of the source sentence. A target sentence is 
\[
Y = (y_1, y_2, \ldots, y_{T_y}).
\]
Similarly, $T_y$ is the length of the target sentence.

The translation function $f$ then reads the input sequence $X$ and computes the
probability over target sentences. In other words,
\begin{align}
    \label{eq:translation_f}
    f: V_x^+ \to C_{|V_y|-1}^+
\end{align}
where $V_x$ is a source vocabulary, and $V_x^+$ is a set of all possible source
sentences of any length $T_x>0$. $V_y$ is a target vocabulary, and $C_k$ is a
standard $k$-simplex. 

What is a standard $k$-simplex? It is a set defined by
\begin{align*}
    C_k = \left\{ 
        (t_0, \ldots, t_k) \in \RR^{k+1} 
        \left|
        \sum_{i=1}^k t_k = 1 
        \text{ and }
        t_i \geq 0 \text{ for all } i
        \right.
    \right\}.
\end{align*}
In short, this set contains all possible settings for categorical distributions
of $k+1$ possible outcomes. This means that the translation function $f$ returns
a probability distribution $P(Y|X)$ over all possible translations of length
$T_y>1$. 

Given a source sentence $X$, this translation function $f$ returns the
conditional probability of a translation $Y$: $P(Y | X)$. Let us rewrite this
conditional probability according to what we have discussed in
Chapter~\ref{chap:nlm}:
\begin{align}
    \label{eq:condlm}
    P(Y | X) = \prod_{t=1}^{T_y} 
    \underbrace{P(y_t | y_{1}, \ldots, y_{t-1},
        \underbrace{X}_{\text{conditional}}
)}_{\text{language modelling}}
\end{align}
Looking at it in this way, it is clear that this is nothing but {\em conditional
language modelling}.  This means that we can use any of the techniques we have
used earlier in Chapter~\ref{chap:nlm} for statistical machine translation.

Training can be trivially done by maximizing the log-likelihood or equivalently
minimizing the {\em negative} log-likelihood (see
Sec.~\ref{sec:distribution_approx}):
\begin{align}
    \label{eq:nmt_cost}
    \tilde{C}(\TT) = -\frac{1}{N} \sum_{n=1}^N \sum_{t=1}^{T_y} \log p(y^n_t |
    y^n_{<t}, X^n),
\end{align}
given a training set 
\begin{align}
    \label{eq:data}
    D = \left\{ (X^1, Y^1), (X^2, Y^2), \ldots, (X^N, Y^N)\right\}
\end{align}
consisting of $N$ training {\em pairs}.

\begin{figure}[t]
    \centering
    \includegraphics[width=0.8\textwidth]{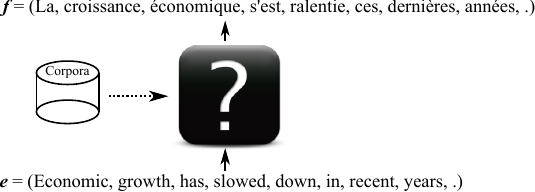}
    \caption{Graphical illustration of statistical machine translation}
    \label{fig:smt}
\end{figure}

All these look extremely straightforward and do not deviate too much from what
we have learned so far in this course. A big picture on this process translation
is shown in Fig.~\ref{fig:smt}.  More specifically, building a statistical
machine translation model is simple, because we have learned how to 
\begin{enumerate}
    \itemsep 0em
    \item Assign a probability to a sentence in
        Sec.~\ref{sec:lm}.
    \item Handle variable-length sequences with recurrent
        neural networks in Sec.~\ref{sec:rnn}.
    \item Compute the gradient of an empirical cost function $\tilde{C}$ with
        respect to the parameters $\TT$ of a recurrent neural network in
        Sec.~\ref{sec:rnn_backprop} and Sec.~\ref{sec:autodiff}.
    \item Use stochastic gradient descent to minimize the cost function in
        Sec.~\ref{sec:sgd}.
\end{enumerate}

Of course, simply knowing all these does not get you a working neural network
that translates from one language to another. We will discuss in detail how we
can build such a neural network in the next section. Before going to the next
section, we must first discuss two issues; (1) where do we get training data?
(2) how do we evaluate machine translation systems?

\subsection{Parallel Corpora: Training Data for Machine Translation}
\label{sec:parallel_corpora}

First, let us consider again what the problem we're trying to solve here. It is
machine translation, and from the description in the previous section and from
Eqs.~\eqref{eq:translation_f}--\eqref{eq:condlm}, it is a {\em
sentence-to-sentence} translation task. We approach this problem by building a
model that takes as input a source sentence $S$ and computes the probability
$P(Y|X)$ of a target sentence $Y$, equivalently a translation. In order for this
model to translate, we must train it with a training set of pairs of a source
sentence and its correct translation.

The very first problem we run into is where we can find this training set which
is often called a {\em parallel corpus}. It is not easy to think of documents
which have been translated into multiple languages. Let's take for instance all
the books that are being translated each year. According to \cite{post2011},
approximately 3\% of titles published each year in English are
translations from another language.\footnote{
    ``According to the information Bowker released in October of 2005, in 2004 there
    were 375,000 new books published in English.'' .. ``Of that total, approx.
    14,440 were new translations, which is slightly more than 3\% of all books
    published.'' \cite{post2011}.
} 
A few international news agencies publish some of their news articles in
multiple languages. For instance, AFP publishes 1,500 stories in French, 700
stories in English, 400 stories in Spanish, 250 stories in Arabic, 200 stories
in German and 150 stories in Portuguese each day, and there are some overlapping
stories across these six languages.\footnote{
    \url{http://www.afp.com/en/products/services/text}
} 
Online commerce sites, such as eBay, often list their products in international
sites with their descriptions in multiple languages.\footnote{
    \url{http://sellercentre.ebay.co.uk/international-selling-tools}
}

Unfortunately these sources of multiple languages of the same content are not
suitable for our purpose. Why is this so? Most importantly, they are often
copy-righted and sold for personal use only. We cannot buy more than 14,400
books in order to train a translation model. We will likely go broke before
completing the purchase, and even if so, it is unclear whether it is acceptable
under copyright to use these text to build a translation model. Because we are
mixing multiple sources of which each is protected under copyright, is the
translation model trained from a mix of all these materials considered a
derivative work?\footnote{
    \url{http://copyright.gov/circs/circ14.pdf}
}

This issue is nothing new, and has been there since the very first statistical
machine translation system was proposed in \cite{brown1990statistical}.
Fortunately, it turned out that there are a number of legitimate sources where
we can get documents translated in more than one languages, often very
faithfully to their content. These sources are parliamentary proceedings of
bilingual, or multilingual countries.

Brown et al.~\cite{brown1990statistical} used the proceedings from the Canadian
parliament, which are by law kept in both French and English. All of these
proceedings are digitally available and called {\em Hansards}. You can check it
yourself online at \url{http://www.parl.gc.ca/}, and here's an excerpt from the
Prayers of the 2nd Session, 41st Parliament, Issue 152:\footnote{
    This is one political lesson here: {\em Canada is still headed by the Queen of
    the United Kingdom}.
}
\begin{itemize}
    \itemsep 0em
    \item {\bf French}: ``ELIZABETH DEUX, par la Grâce de Dieu, REINE du
        Royaume-Uni, du Canada et de ses autres royaumes et territoires, Chef du
        Commonwealth, D\'efenseur de la Foi.''
    \item {\bf English}: ``ELIZABETH THE SECOND, by the Grace of God of the
        United Kingdom, Canada and Her other Realms and Territories QUEEN, Head
        of the Commonwealth, Defender of the Faith.''
\end{itemize}
Every single word spoken in the Canadian parliament is translated either into
French or into English. A more recent version of Hansards preprocessed for
research can be found at
\url{http://www.isi.edu/natural-language/download/hansard/}.

Similarly, the European parliament used to provided the parliamentary
proceedings in all 23 official languages.\footnote{
    Unfortunately, the European parliament decided to stop translating its
    proceedings into all 23 official languages on 21 Nov 2011 as an effort
    toward budget cut. See
    \url{http://www.euractiv.com/culture/parliament-cuts-translation-budg-news-516201}.
} This is a unique data in the sense that each and every sentence is translated
into either 11 or 26 official languages. For instance, here is one example
\cite{koehn2005europarl}:
\begin{itemize}
    \itemsep 0em
    \item {\bf Danish}: det er næsten en personlig rekord for mig dette
        efter{\aa}r.
    \item {\bf German}: das ist f\"ur mich fast pers\"onlicher rekord in diesem
        herbst .
    \item {\bf Greek}: (omitted)
    \item {\bf English}: that is almost a personal record for me this autumn !
    \item {\bf Spanish}: es la mejor marca que he alcanzado este oto\~no .
    \item {\bf Finnish}: se on melkein minun enn\"atykseni t\"an\"a syksyn\"a !
    \item {\bf French}: c ' est pratiquement un record personnel pour moi , cet
        automne !
    \item {\bf Italian}: e ' quasi il mio record personale dell ' autunno .
    \item {\bf Dutch}: dit is haast een persoonlijk record deze herfst .
    \item {\bf Portuguese}: \'e quase o meu recorde pessoal deste semestre !
    \item {\bf Swedish}: det \"ar n\"astan personligt rekord f\"or mig denna h\"ost !
\end{itemize}

The European proceedings has been an invaluable resource for machine translation
research. At least, the existing multilingual proceedings (up to 2011) can be
still used, and it is known in the field as the ``Europarl'' corpus
\cite{koehn2005europarl} and can be downloaded from
\url{http://www.statmt.org/europarl/}. 

These proceedings-based parallel corpora have two distinct advantages. First,
in many cases, the sentences in those corpora are well-formed, and their
translations are done by professionals, meaning the quality of the corpora is
guaranteed. Second, surprisingly, the topics discussed in those proceedings are
quite diverse. Clearly the members of the parliament do not often chitchat too
often, but they do discuss a diverse set of topics. Here's one such example from
the Europarl corpus:
\begin{itemize}
    \itemsep 0em
    \item {\bf English}: 
        Although there are now two Finnish channels and one Portuguese one, there is
        still no Dutch channel, which is what I had requested because Dutch people here
        like to be able to follow the news too when we are sent to this place of exile
        every month.
    \item {\bf French}: 
        Il y a bien deux chaînes finnoises et une chaîne portugaise, mais il n'y a
        toujours aucune chaîne néerlandaise. Pourtant je vous avais demand\'e une chaîne
        n\'eerlandaise, car les Néerlandais aussi désirent pouvoir suivre les actualités
        chaque mois lorsqu'ils sont envoyés en cette terre d'exil.
\end{itemize}

One apparent limitation is that these proceedings cover only a handful of
languages in the world, mostly west European languages. This is not desirable.
Why? According to Ethnologue (2014)\footnote{
    \url{http://www.ethnologue.com/world}
}, the top-five most spoken languages in the world are
\begin{enumerate}
    \itemsep 0em
    \item Chinese: approx. 1.2 billion
    \item Spanish: approx. 414 million
    \item English: approx. 335 million
    \item Hindi: approx. 260 million
    \item Arabic: approx. 237 million
\end{enumerate}
There are only two European languages in this list.

So, then, where can we get all data for all these non-European languages? There
are a number of resources you can use, and let me list a few of them here:

You can find the translated subtitle of the TED talks at the Web Inventory of
Transcribed and Translated Talks (WIT$^3$, \url{https://wit3.fbk.eu/})
\cite{cettoloEtAl:EAMT2012}. It is a quite small corpus, but includes 104
languages. For Russian--English data, Yandex released a parallel corpus of one
million sentence pairs. You can get it at
\url{https://translate.yandex.ru/corpus?lang=en}. You can continue with other
languages by googling very hard, but eventually you run into a hard wall.

This hard wall is not only the lack of any resource, but also lack of enough
resource. For instance, I quickly googled for Korean--English parallel corpora
and found the following resources:
\begin{itemize}
    \item SWRC English-Korean multilingual corpus: 60,000 sentence pairs
        \url{http://semanticweb.kaist.ac.kr/home/index.php/Corpus10}
    \item Jungyeul's English-Korean parallel corpus: 94,123 sentence pairs
        \url{https://github.com/jungyeul/korean-parallel-corpora}
\end{itemize}
This is just not large enough. 

One way to avoid this or mitigate this problem is to automatically mine parallel
corpora from the Internet. There have been quite some work in this direction as
a way to increase the size of parallel corpora
\cite{resnik2003web,zhang2006automatic}. The idea is to build an algorithm that
crawls the Internet and find a pair of corresponding pages in two different
languages. One of the largest preprocessed corpus of multiple languages from the
Internet is the Common Crawl Parallel Corpus created by Smith et
al.~\cite{smith2013dirt} available at
\url{http://www.statmt.org/wmt13/training-parallel-commoncrawl.tgz}.

\subsection{Automatic Evaluation Metric}
\label{sec:bleu}

Let's say we have trained a machine translation model on a training corpus. A
big question follows: {\em how do we evaluate this model}?

In the case of classification, evaluation is quite straightforward. All we need
to do is to classify held-out test examples with a trained classifier and see
how many examples were correctly classified. This is however not true in the
case of translation. 

There are a number of issues, but let us discuss two most important problems
here. First, there may be many correct translations given a single source
sentence. For instance, the following three sentences are the translations made
by a human translator given a single Chinese sentence \cite{papineni2002bleu}:
\begin{itemize}
    \itemsep 0em
    \item It is a guide to action that ensures that the military will forever
        heed Party commands.
    \item It is the guiding principle which guarantees the military forces
        always being under the command of the Party.
    \item It is the practical guide for the army always to heed the directions
        of the party.
\end{itemize}
They all clearly differ from each other, although they are the translations of a
single source sentence. 

Second, the quality of translation cannot be measured as either success or
failure. It is rather a smooth measure between success and failure. Let us
consider an English translation of a French sentence ``J'aime un llama, qui est
un animal mignon qui vit en Am\'erique du Sud''.\footnote{
    I would like to thank Laurent Dinh for the French translation.
}

One possible English translation of this French sentence is ``I like a llama
which is a cute animal living in South America''. Let's give this translation a
score $100$ (success). According to Google translate, the French sentence above
is ``I like a llama, a cute animal that lives in South America''. I see that
Google translate has omitted ``qui est'' from the original sentence, but the
whole meaning has well been captured. Let us give this translation a slightly
lower score of $90$. 

Then, how about ``I like a llama from South America''?  This is certainly not a
correct translation, but except for the part about a llama being cute, this
sentence does communicate most of what the original French sentence tried to
communicate. Maybe, we can give this translation a score of $50$. 

How about ``I do not like a llama which is an animal from South America''? This
translation correctly describes the characteristics of llama exactly as
described in the source sentence. However this translation incorrectly states
that I do not like a llama, when I like a llama according to the original French
sentence. What kind of score would you give this translation?

Even worse, we want an automated evaluation algorithm. We cannot look at
thousands of validation or test sentence pairs to tell how well a machine
translation model does. Even if we somehow did it for a single model, in order
to compare this translation model against others, we must do it for every single
machine translation model under comparison. We must have an automatic evaluation
metric in order to efficiently test and compare different machine translation
models.

\paragraph{BLEU} 
One of the most widely used automatic evaluation metric for assessing the
quality of translations is BLEU proposed in \cite{papineni2002bleu}. BLEU
computes the geometric mean of the modified $n$-gram precision scores multiplied
by brevity penalty. Let me describe this in detail here.

First, we define the modified $n$-gram precision $p_n$ of a translation $Y$ as
\begin{align*}
    p_n = \frac{
        \sum_{S \in C} \sum_{\text{ngram} \in S} \hat{c}(\text{ngram})
    }
    {
        \sum_{S \in C} \sum_{\text{ngram} \in S} c(\text{ngram})
    },
\end{align*}
where $C$ is a corpus of all the sentences/translations, and $S$ is a set of all
unique $n$-grams in one sentence in $C$. $c(\text{ngram})$ is the count of the
$n$-gram, and $\hat{c}(\text{ngram})$ is
\begin{align*}
    \hat{c}(\text{ngram}) = \min(c(\text{ngram}), c_{\text{ref}}(\text{ngram})).
\end{align*}
$c_{\text{ref}}(\text{ngram})$ is the count of the $n$-gram in reference
sentences.

What does this modified $n$-gram precision measure? It measures the ratio
between the number of $n$-grams in the translation and the number of those
$n$-grams actually occurred in a reference (ground-truth) translation. If there
is no $n$-gram from the translation in the reference, this modified precision
will be zero because $c_{\text{ref}}(\cdot)$ will be zero all the time. 

It is common to use the geometric average of modified 1-, 2-, 3- and 4-gram
precisions, which is computed by
\begin{align*}
    P_1^4 = \exp\left( \frac{1}{4} \sum_{n=1}^4 \log p_n \right).
\end{align*}

If we use this geometric average $P$ as it is, there is a big loophole. One can
get a high average modified precision by making as short a translation as
possible. For instance, a reference translation is
\begin{itemize}
    \item I like a llama, a cute animal that lives in South America .
\end{itemize}
and a translation we are trying to evaluate is
\begin{itemize}
    \item cute animal that lives
\end{itemize}
This is clearly a very bad translation, but the modified 1-, 2-, 3- and 4-gram
precisions will be high. The modified precisions are
\begin{align*}
    p_1 =& \frac{1+1+1+1}{1+1+1+1} = \frac{4}{4} = 1 \\
    p_2 =& \frac{1+1+1}{1+1+1} = \frac{3}{3} = 1 \\
    p_3 =& \frac{1+1}{1+1} = \frac{2}{2} = 1 \\
    p_4 =& \frac{1}{1} = \frac{1}{1} = 1.
\end{align*}
Their geometric average is then
\begin{align*}
    P_1^4 = \exp\left( \frac{1}{4} \left( 0 + 0 + 0 + 0 \right)\right) = 1
\end{align*}
which is the maximum modified precision you can get!

In order to avoid this behaviour, BLEU penalizes the geometric average of the
modified $n$-gram precisions by the ratio of the lengths between the reference
$r$ and translation $l$. This is done by first computing a brevity penalty:
\begin{align*}
    \BP = \left\{ 
        \begin{array}{l l}
            1 & \text{, if }l \geq r \\
            \exp\left( 1 - \frac{r}{l} \right) & \text{, if }l < r
        \end{array}
        \right.
\end{align*}
If the translation is longer than the reference, it uses the geometric average
of the modified $n$-gram precisions as it is. Otherwise, it will penalize it by
multiplying the average precision with a scalar less than $1$. 
In the case of the example above, the brevity penalty is $0.064$, and the final
BLEU score is $0.064$. 

\begin{figure}[ht]
    \centering
    \begin{minipage}{0.48\textwidth}
        \centering
        \includegraphics[width=0.99\columnwidth]{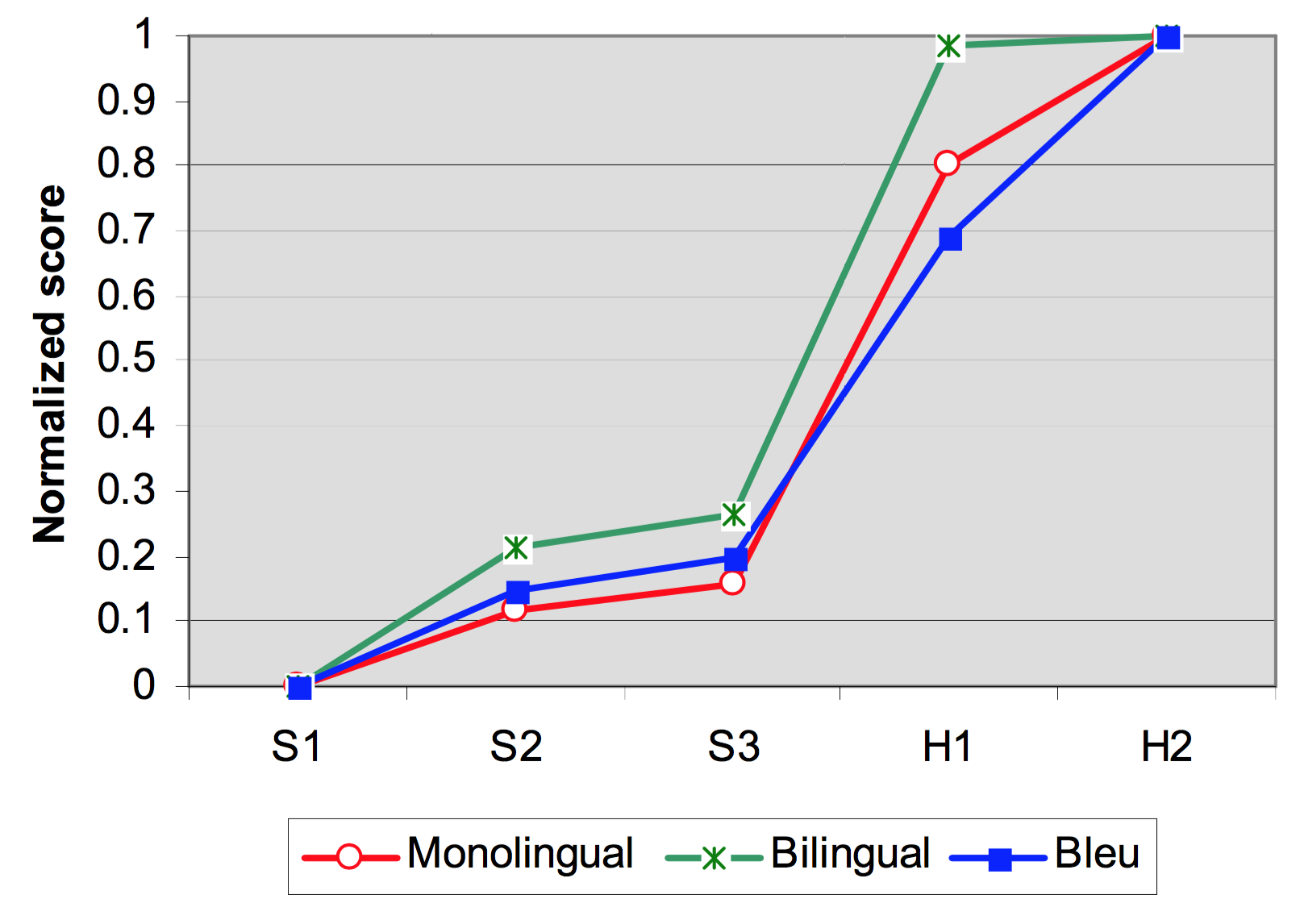}
    \end{minipage}
    \hfill
    \begin{minipage}{0.48\textwidth}
        \centering
        \includegraphics[width=0.99\columnwidth]{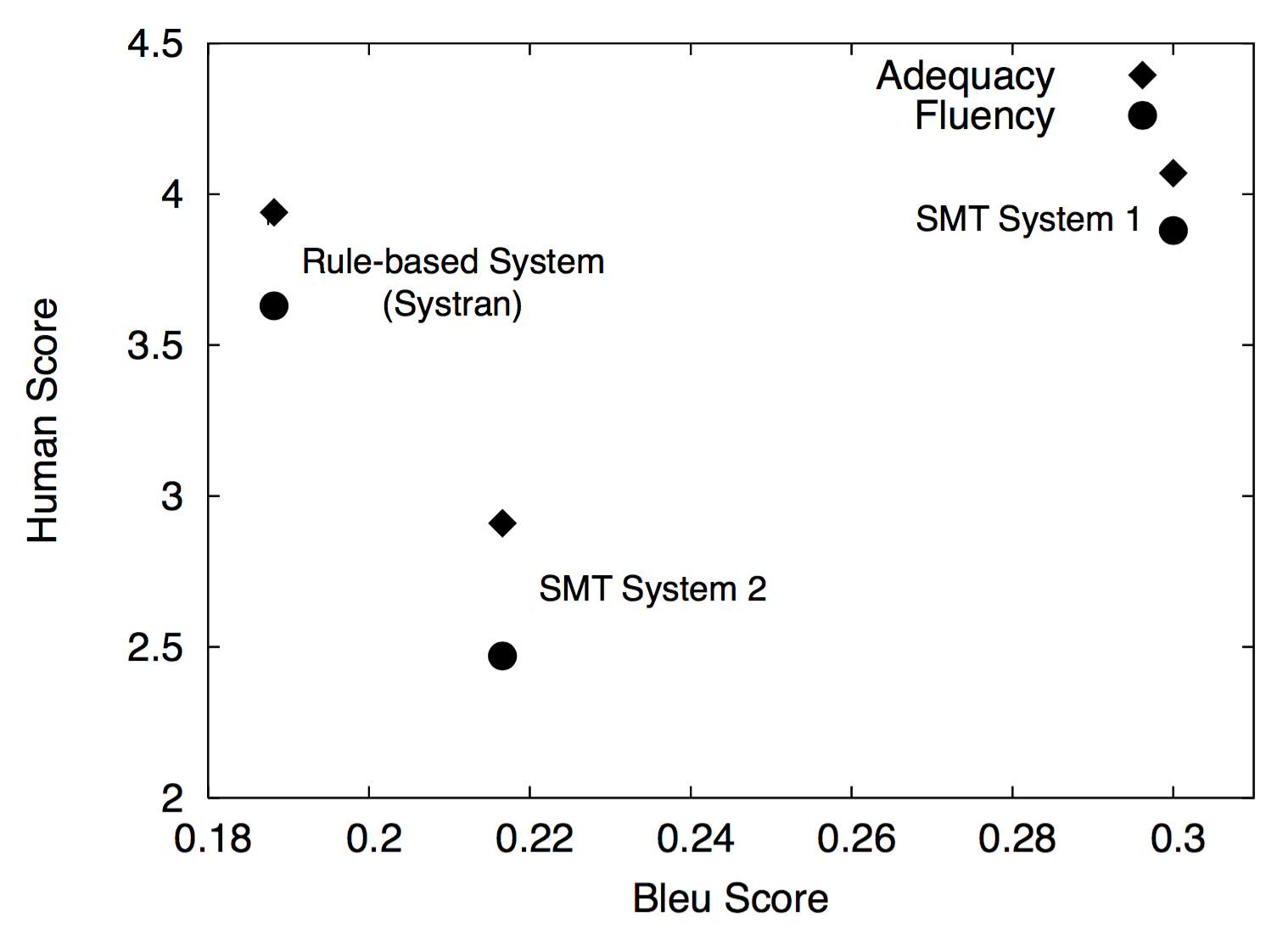}
    \end{minipage}

    \begin{minipage}{0.48\textwidth}
        \centering
        (a)
    \end{minipage}
    \hfill
    \begin{minipage}{0.48\textwidth}
        \centering
        (b)
    \end{minipage}
    \caption{(a) BLEU vs. bilingual and monolingual judgements of three machine
    translation systems (S1, S2 and S3) and two humans (H1 and H2). Reprinted
from \cite{papineni2002bleu}. (b) BLEU vs. human judgement (adequacy and fluency
separately) of three machine translation systems (two statistical and one
rule-based systems). Reprinted from \cite{callison2006re}.}
    \label{fig:bleu}
\end{figure}

The BLEU was shown to correlate well with human judgements in the original
article \cite{papineni2002bleu}. Fig.~\ref{fig:bleu}~(a) shows how BLEU correlates
with the human judgements in comparing different translation systems.

This is however not to be taken as a message saying that the BLEU is the perfect
automatic evaluation metric. It has been shown that the BLEU is only adequate in
comparing two similar machine translation systems, but not too much so in
comparing two very different systems. For instance, Callison-Burch et
al.~\cite{callison2006re} observed that the BLEU underestimates the quality of
the machine translation system that is {\em not} a phrase-based statistical
system. See Fig.~\ref{fig:bleu}~(b) for an example.

BLEU is definitely not a perfect metric, and many researchers strive to build a
better evaluation metric for machine translation systems. Some of the
alternatives available at the moment are
METEOR~\cite{denkowski:lavie:meteor-wmt:2014} and TER~\cite{snover2006study}.

\section{Neural Machine Translation: \\ Simple Encoder-Decoder Model}
\label{sec:nmt_simple}

From the previous section and  from Eq.~\ref{eq:condlm}, it is clear that we
need to model each conditional distribution inside the product as a function.
This function will take as input all the previous words in the target sentence
$Y=(y_1, \ldots, y_{t-1})$ and the whole source sentence $X=(x_1, \ldots, x_{T_x})$.
Given these inputs the function will compute the probabilities of all the words
in the target vocabulary $V_y$. In this section, I will describe an approach
that was proposed multiple times independently over 17 years in
\cite{forcada1997recursive,cho2014learning,sutskever2014sequence}.

Let us start by tackling how to handle the source sentence $X=(x_1, \ldots,
x_{T_x})$. Since this is a variable-length sequence, we can readily use a
recurrent neural network from Chapter~\ref{chap:rnn}. However, unlike the
previous examples, there is no explicit target/output in this case. All we need
is a (vector) summary of the source sentence.

We call this recurrent neural network an {\em encoder}, as it encodes the source
sentence into a (continuous vector) code. It is implemented as
\begin{align}
    \label{eq:nmt_encoder}
    \vh_t = \phi_{\enc}\left( \vh_{t-1}, \mE_x^\top \vx_t \right).
\end{align}
As usual, $\phi_{\enc}$ can be any recurrent activation function, but it is
highly recommended to use either gated recurrent units (see Sec.~\ref{sec:gru})
or long short-term memory units (see Sec.~\ref{sec:lstm}.) $\mE_x \in \RR^{|V_x|
\times d}$ is an input weight matrix containing word vectors as its rows (see
Eq.~\eqref{eq:word_emb} in Sec.~\ref{sec:nlm},) and $\vx_t$ is an one-hot vector
representation of the word $x_t$ (see Eq.~\eqref{eq:one_hot_vector} in
Sec.~\ref{sec:nlm}.) $\vh_0$ is initialized as an all-zero vector.

\begin{figure}[t]
    \centering
    \begin{minipage}{0.48\textwidth}
        \centering
        \includegraphics[width=0.98\columnwidth]{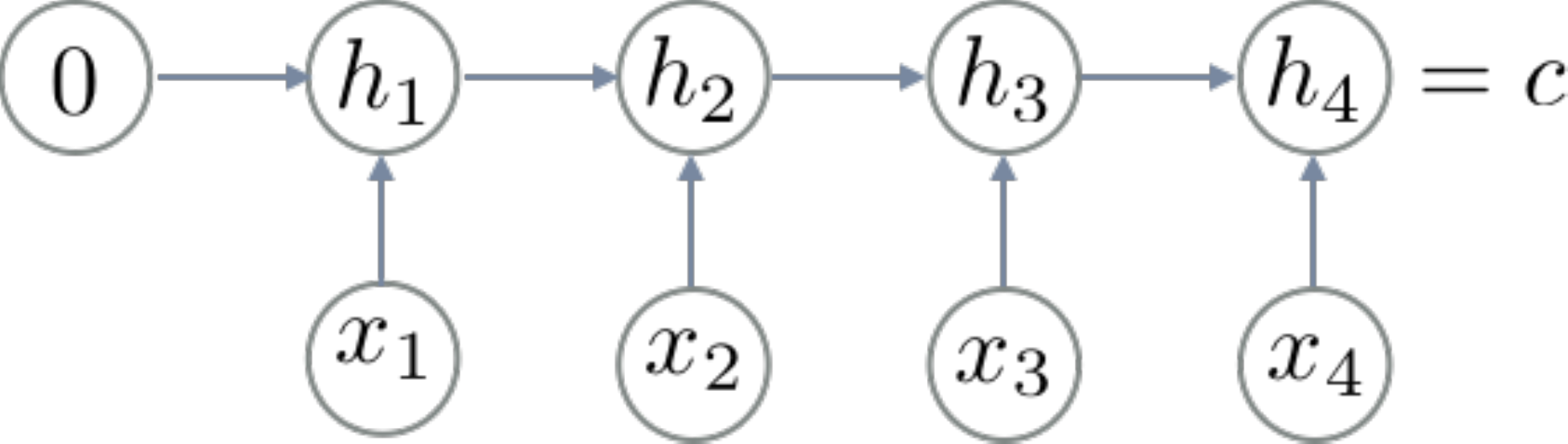}
    \end{minipage}
    \hfill
    \begin{minipage}{0.48\textwidth}
        \centering
        \includegraphics[width=0.98\columnwidth]{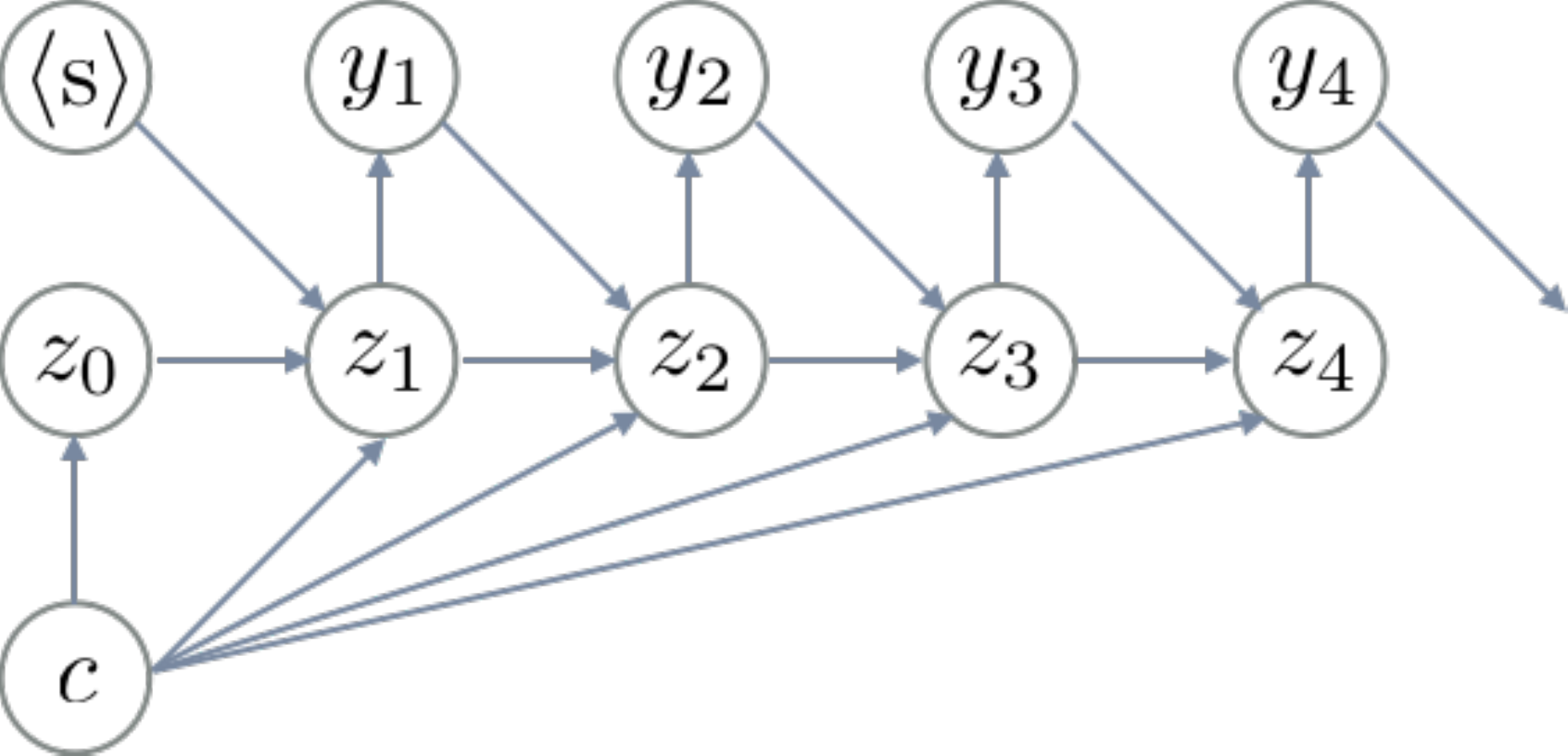}
    \end{minipage}

    \begin{minipage}{0.48\textwidth}
        \centering
        (a)
    \end{minipage}
    \hfill
    \begin{minipage}{0.48\textwidth}
        \centering
        (b)
    \end{minipage}
    \caption{(a) The encoder and (b) the decoder of a simple neural machine
    translation model}
    \label{fig:nmt}
\end{figure}

After reading the whole sentence up to $\vx_{T_x}$, the last memory state
$\vh_{T_x}$ of the encoder summarizes the whole source sentence into a single
vector, as shown in Fig.~\ref{fig:nmt_enc}~(a). Thanks to this encoder, we can now
work with a single vector instead of a whole sequence of source words. Let us
denote this vector as $\vc$ and call it a {\em context vector}.

We now need to design a {\em decoder}, again, using a recurrent neural network.
As I mentioned earlier, the decoder is really nothing but a language model,
except that it is conditioned on the source sentence $X$. What this means is
that we can build a recurrent neural network language model from
Sec.~\ref{sec:rlm} but feeding also the context vector at each time step. In
other words,
\begin{align}
    \label{eq:nmt_decoder}
    \vz_t = \phi_{\dec}\left( \vz_{t-1}, \left[ \mE_y^\top \vy_{t-1}; \vc
    \right]\right)
\end{align}
Do you see the similarity and dissimilarity to Eq.~\eqref{eq:rnnlm_iteration}
from Sec.~\ref{sec:rlm}? It's essentially same, except that the input at time $t$
is a concatenated vector of the word vector of the previous word $y_{t-1}$ and
the context vector $\vc$.

Once the decoder's memory state is updated, we can compute the probabilities of
all possible target words by
\begin{align}
    \label{eq:nmt_decoder_out}
    p(y_t = w'|y_{<t}, X) \propto \exp\left( \ve_{w'}^\top \vz_t \right),
\end{align}
where $\ve_{w'}$ is the target word vector associated the word $w'$. This is
equivalent to affine-transforming $\vz_t$ followed by a softmax function from
Eq.~\eqref{eq:softmax} from Sec.~\ref{sec:distribution_approx}.

Now, should we again initialize $\vz_0$ to be an all-zero vector? Maybe, or
maybe not. One way to view what this decoder does is that the decoder models a
{\em trajectory} in a continuous vector space, and each point in the trajectory
is $\vz_t$. Then, $\vz_0$ acts as a starting point of this trajectory, and it is
natural to initialize this starting point to be a point relevant to the source
sentence. Because we have access to the source sentence's content via $\vc$, we
can again use it to initialize $\vz_0$ as
\begin{align}
    \label{eq:nmt_init}
    \vz_0 = \phi_{\init}\left( \vc \right).
\end{align}
See Fig.~\ref{fig:nmt}~(b) for the graphical illustration of the decoder.

Although I have used $\vc$ as if it is a separate variable, this is not true.
$\vc$ is simply a shorthand notation of the last memory state of the encoder
which is a function of the whole source sentence. What does this mean? It means
that we can compute the gradient of the empirical cost function in
Eq.~\eqref{eq:nmt_cost} with respect to all the parameters of both the encoder
and decoder and maximize the cost function using stochastic gradient descent,
just like any other neural network we have learned so far in this course.

\subsection{Sampling vs. Decoding}

\paragraph{Sampling}

We are ready to compute the conditional distribution $P(Y|X)$ over all possible
translations given a source sentence. When we have a distribution, the first
thing we can try is to sample from this distribution. Often, it is not
straightforward to generate samples from a distribution, but fortunately, in
this case, we can readily generate exact samples from the distribution $P(Y|X)$. 

We simply iterate over the following steps until a token indicating the end of a
sentence ($\left< \text{eos} \right>$):
\begin{enumerate}
    \itemsep 0em
    \item Compute $\vc$ (Eq.~\eqref{eq:nmt_encoder})
    \item Initialize $\vz_0$ with $\vc$ (Eq.~\eqref{eq:nmt_init})
    \item Compute $\vz_t$ given $\vz_{t-1}$, $\vy_{t-1}$ and $\vc$
        (Eq.~\eqref{eq:nmt_decoder})
    \item Compute $p(y_t|y_{<t}, X)$ (Eq.~\eqref{eq:nmt_decoder_out})
    \item Sample $\tilde{y}_t$ from the compute distribution
    \item Repeat (3)--(5) until $\tilde{y}_t = \left< \text{eos} \right>$
\end{enumerate}

After taking these steps, we get a sample $\tilde{Y}=\left( \tilde{y}_1, \ldots,
\tilde{y}_{|\tilde{Y}|}\right)$ given a source sentence $X$. Of course, there is
no guarantee that this will be a good translation of $X$. In order to find a
good translation, meaning a translation with a high probability
$P(\tilde{Y}|X)$, we need to repeatedly sample multiple translations from
$P(Y|X)$ and choose one with the high probability.

This is not too desirable, as it is not clear how many translations we need to
sample from $P(Y|X)$ and also it will likely be computationally expensive. We
must wonder whether we can solve the following optimization problem directly:
\begin{align*}
    \tilde{Y} = \argmax_{Y} \log P(Y|X).
\end{align*}
Unfortunately, the exact solution to this requires evaluating $P(Y|X)$ for every
possible $Y$. Even if we limit our search space of $Y$ to consist of only
sentences of length up to a finite number, it will likely become too large (the
cardinality of the set grows exponentially with respect to the number of words
in a translation.) Thus, it only makes sense to solving the optimization problem
above approximately.

\paragraph{Approximate Decoding: Beamsearch}

Although it is quite clear that finding a translation $\tilde{Y}$ that maximizes
the log-probability $\log P(\tilde{Y}|X)$ is extremely expensive, we will
regardlessly try it here. 

\begin{figure}[t]
    \centering
    \begin{minipage}{0.48\textwidth}
        \centering
        \includegraphics[width=0.98\columnwidth]{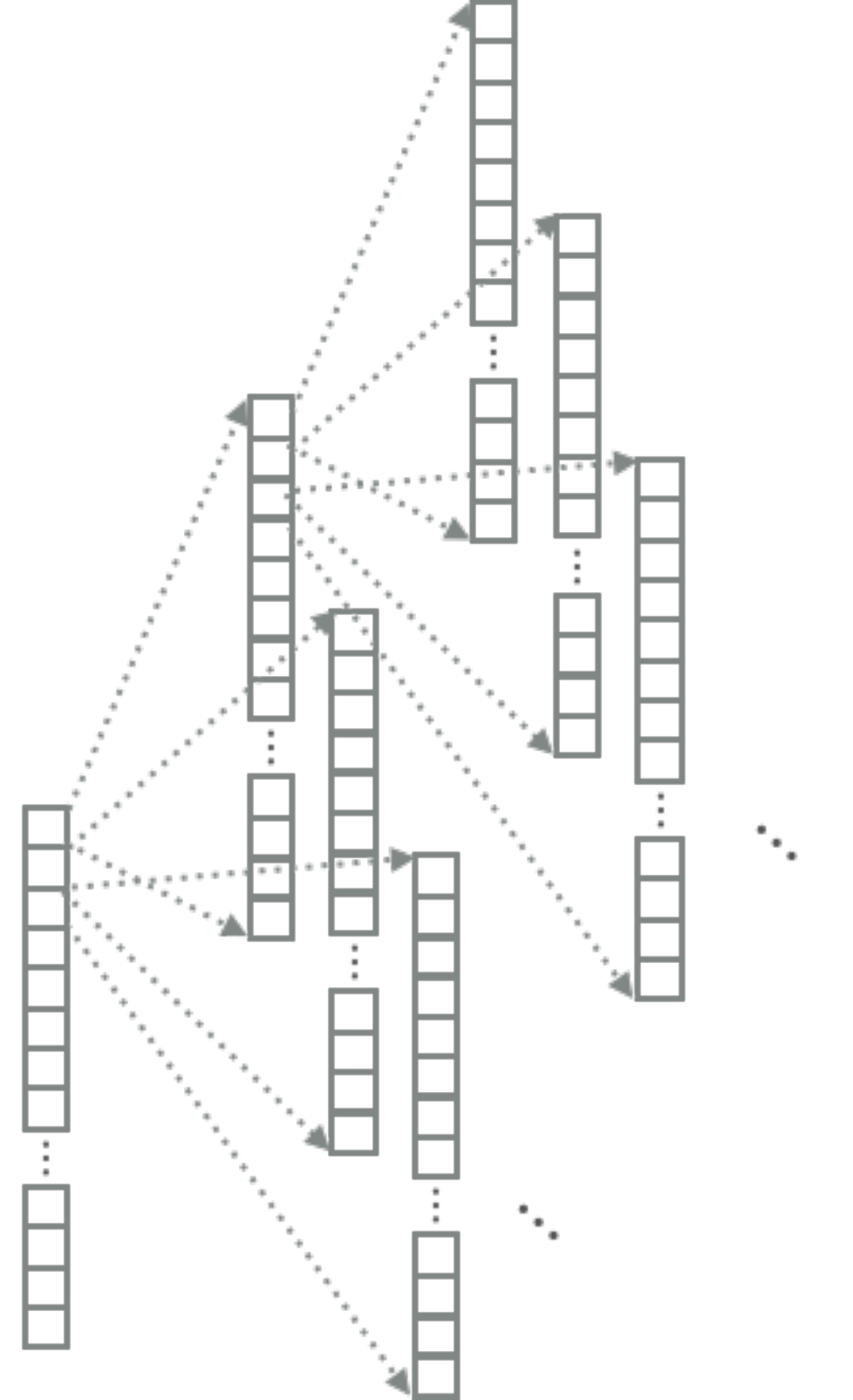}
    \end{minipage}
    \hfill
    \begin{minipage}{0.48\textwidth}
        \centering
        \includegraphics[width=0.98\columnwidth]{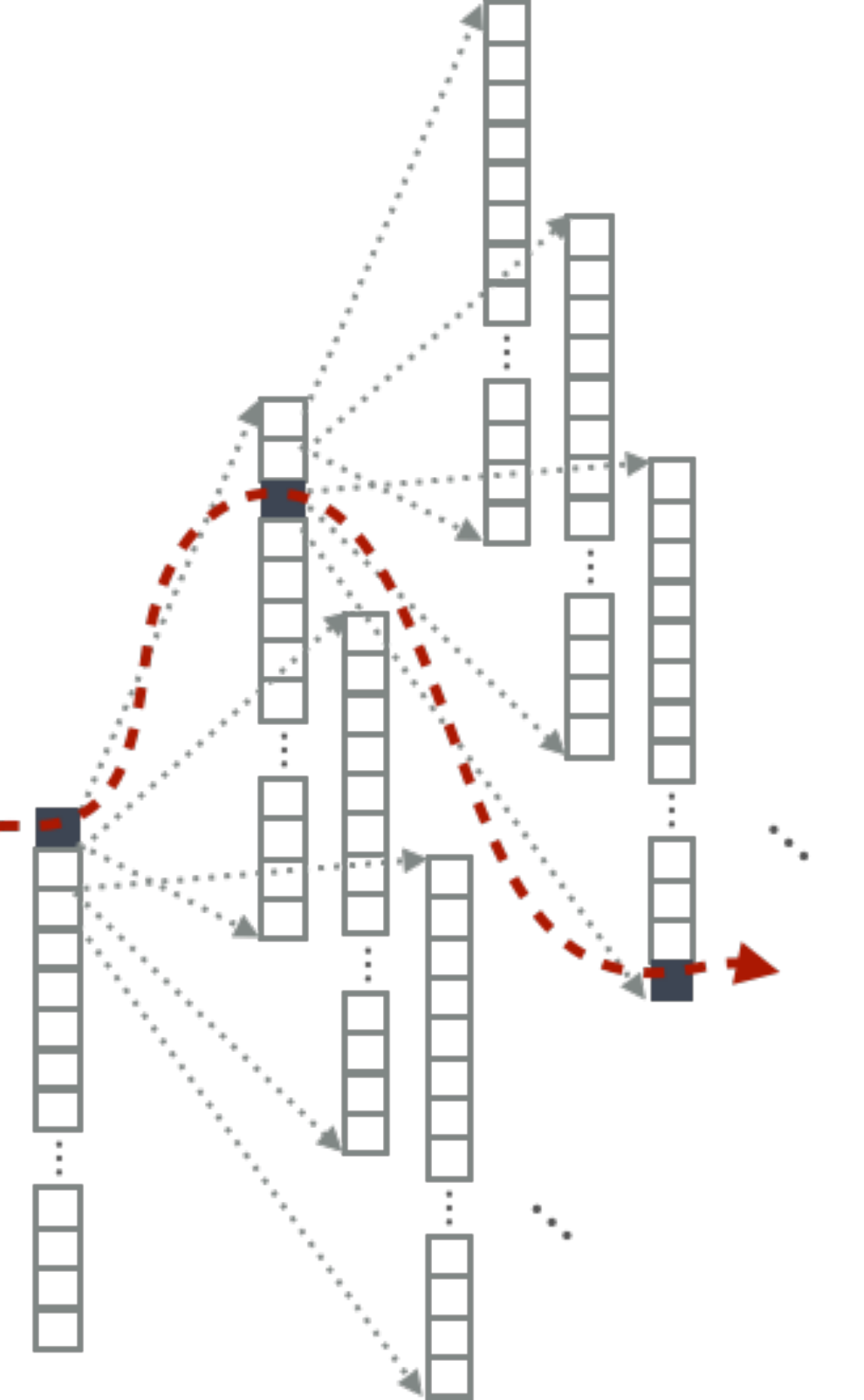}
    \end{minipage}

    \vspace{3mm}

    \begin{minipage}{0.48\textwidth}
        \centering
        (a)
    \end{minipage}
    \hfill
    \begin{minipage}{0.48\textwidth}
        \centering
        (b)
    \end{minipage}
    \caption{(a) Search space depicted as a tree. (b) Greedy search.}
    \label{fig:searchspace}
\end{figure}

One very natural way to enumerate all possible target sentences and
simultaneously computing the log-probability of each and every one of them is to
start from all possible first word, compute the probabilities of them, and from
each potential first word branch into all possible second words, and so on. This
procedure forms a tree, and any path from the root of this tree to any
intermediate node is a valid, but perhaps very unlikely, sentence. See
Fig.~\ref{fig:searchspace} for the illustration.  The conditional probabilities
of all these paths, or sentences, can be computed as we expand this tree down by
simply following Eq.~\eqref{eq:condlm}.

Of course, we cannot compute the conditional probabilities of all possible
sentences. Hence, we must resort to some kind of {\em approximate search}. Wait,
{\em search}? Yes, this whole procedure of finding the most likely translation
is equivalent to searching through a space, in this case a tree, of all possible
sentences for one sentence that has the highest conditional probability.

The most basic approach to approximately searching for the most likely
translation is to choose only a single branch at each time step $t$. In other
words, 
\begin{align*}
    \hat{y}_t = \argmax_{w' \in V} \log p(y_t = w' | \hat{y}_{<t}, X),
\end{align*}
where the conditional probability is defined in Eq.~\eqref{eq:nmt_decoder_out},
and $\hat{y}_{<t}=(\hat{y}_1, \hat{y}_2, \ldots, \hat{y}_{t-1})$ is a sequence
of greedily-selected target words up to the $(t-1)$-th step. This procedure is
repeated until the selected $\hat{y}_t$ is a symbol corresponding to the end of
the translation (often denoted as \eos{}.) See Fig.~\ref{fig:searchspace}~(b)
for illustration.

There is a big problem of this {\em greedy search}. That is, as soon as it makes
one mistake at one time step, there is no way for this search procedure to
recover from this mistake. This happens because the conditional distributions at
later steps depend on the choices made earlier. 

Consider the following two sequences: $\left(w_1, w_2\right)$ and $\left(w_1',
w_2\right)$. These sequences' probabilities are 
\begin{align*}
    &p(w_1, w_2) = p(w_1) p(w_2 | w_1), \\
    &p(w_1', w_2)= p(w_1') p(w_2 | w_1')
\end{align*}
Let's assume that 
\[
    \lambda p(w_1) = p(w_1'),
\]
where $0 < \lambda < 1$, meaning that $p(w_1) > p(w_1')$. In this case, the
greedy search will choose $w_1$ over $w_1'$ and ignore $w_1'$.

Now we can see that there's a problem with this. Let's assume that
\begin{align*}
    \lambda p(w_2 | w_1) < p(w_2 | w_1') \iff p(w_2 | w_1) < \frac{1}{\lambda}
    p(w_2 | w_1')
\end{align*}
where $\lambda$ was defined earlier. In this case,
\begin{align*}
    p(w_1, w_2) =& p(w_1) p(w_2 | w_1) 
    = \lambda p(w_1') p(w_2 | w_1) \\ <&
    \cancel{\lambda} p(w_1') \frac{1}{\cancel{\lambda}} p(w_2 | w_1') 
    = p(w_1') p(w_2 | w_1') = p(w_1', w_2).
\end{align*}
In short,
\begin{align*}
    p(w_1, w_2) < p(w_1', w_2).
\end{align*}
It means that the sequence $(w_1',w_2)$ is more likely than $(w_1, w_2)$, but
the greedy search algorithm is unable to notice this, because simply $p(w_1) >
p(w_1')$.  

Unfortunately, the only way to completely avoid this undesirable situation is to
consider all the possible paths starting from the very first time step. This is
exactly the reason why we introduced the greedy search in the first place, but
the greedy search is {\em too} greedy. The question is then whether there is
something in between the exact search and the greedy search.

\paragraph{Beam Search}

Let us start from the very first position $t=1$. First, we compute the
conditional probabilities of all the words in the vocabulary:
\begin{align*}
    p(y_1=w|X)\text{ for all }w\in V.
\end{align*}
Among these, we choose the $K$ most likely words and initialize the $K$
hypotheses:
\begin{align*}
    (w_1^1),(w_2^1), \ldots, (w_K^1)
\end{align*}
We use the subscript to denote the hypothesis and the subscript the time step.
As an example, $w_1^1$ is the first hypothesis at time step $1$.

For each hypothesis, we compute the next conditional probabilities of all the
words in the vocabulary:
\begin{align*}
    p(y_2=w|y_{<1}=(w_i^1), X)\text{ for all }w \in V,
\end{align*}
where $i=1, \ldots, K$. 
We then have $K \times |V|$ candidates with the corresponding probabilities:
\begin{align*}
    K \underbrace{\left\{
        \begin{array}{c c c}
                p(w_1^1, w_{c,1}^2),& \ldots,& p(w_1^1, w_{c,|V|}^2) \\
                p(w_2^1, w_{c,1}^2),& \ldots,& p(w_2^1, w_{c,|V|}^2) \\
                                   &\vdots& \\
            p(w_K^1, w_{c,1}^2),& \ldots,& p(w_K^1, w_{c,|V|}^2) \\
        \end{array}
    \right.}_{|V|}
\end{align*}
Among these $K \times |V|$ candidates, we choose the $K$ most likely candidates:
\begin{align*}
    (w_1^1, w_1^2), (w_2^1, w_2^2), \ldots, (w_K^1,w_K^2).
\end{align*}
Starting from these $K$ new hypotheses, we repeat the process of computing the
probabilities of all $K \times |V|$ possible candidates and choosing among them
the $K$ most likely new hypotheses. 

\begin{figure}[t]
    \centering
    \centering
    \includegraphics[width=0.7\columnwidth]{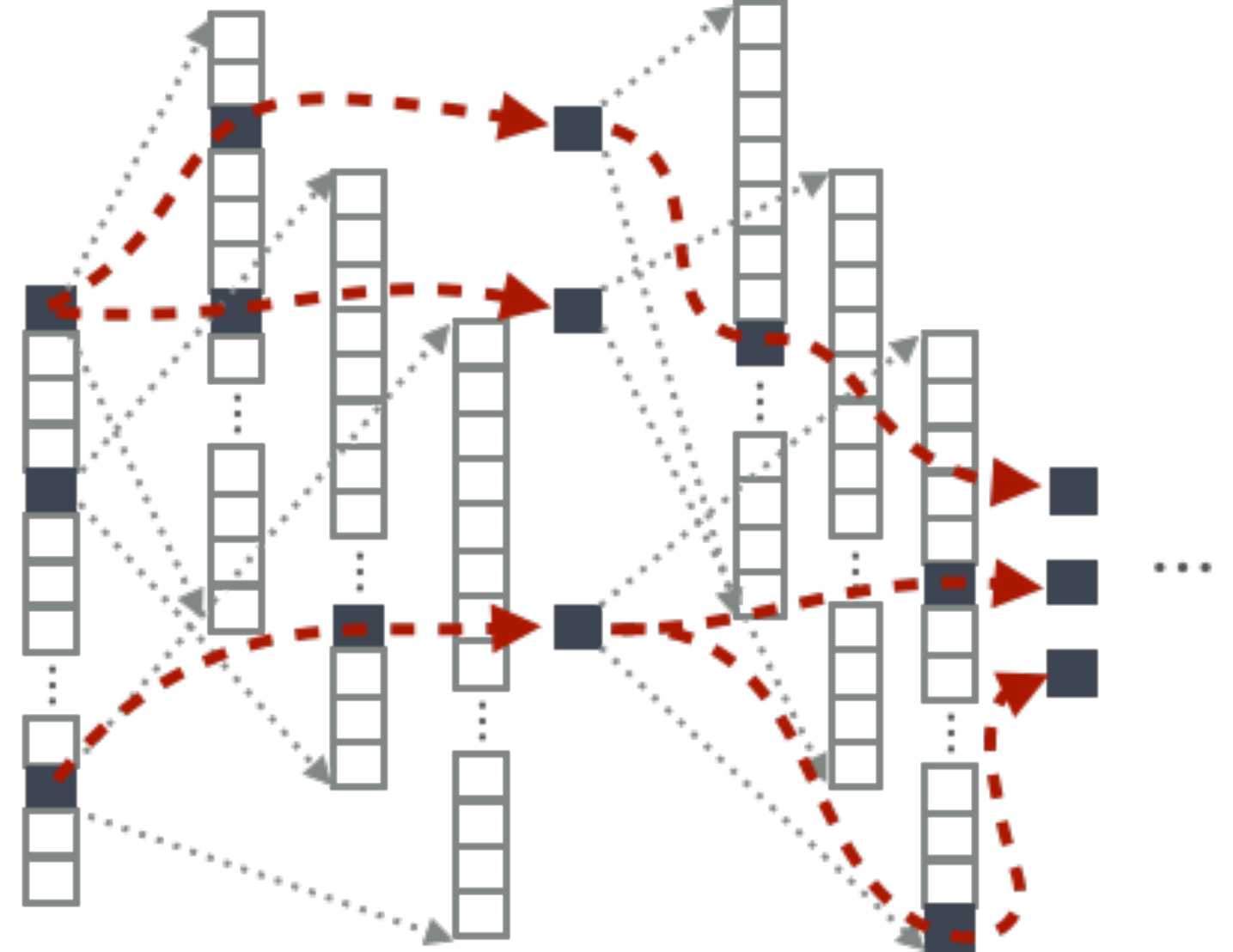}

    \caption{Beam search with the beam width set to $3$.}
    \label{fig:beamsearch}
\end{figure}

It should be clear that this procedure, called {\em beam search} and shown in
Fig.~\ref{fig:beamsearch}, becomes
equivalent to the exact search, as $K\to \infty$. Also, when $K=1$, this
procedure is equivalent to the greedy search. In other words, this beam search
interpolates between the exact search, which is computationally intractable but
exact, and the greedy search, which is computationally very cheap but probably
quite inexact, by changing the size $K$ of hypotheses maintained throughout the
search procedure. 

How do we choose $K$? One might mistakenly think that we can simply use as large
$K$ as possible given the constraints on computation and memory. Unfortunately,
this is not necessarily true, as this interpolation by $K$ is not monotonic.
That is, the quality of the translation found by the beam search with a larger
$K$ is not necessarily better than the translation found with a smaller $K$.

Let us consider the case of vocabulary having three symbols $\{ a, b, c\}$ and
any valid translation being of a length $3$. In the first step, we have
\begin{align*}
    p(a) = 0.5, p(b) = 0.15, p(c) = 0.45.
\end{align*}
In the case of $K=1$, i.e., {\em greedy search}, we choose $a$. If $K=2$, we
will keep $(a)$ and $(c)$. 

Given $a$ as the first symbol, we have
\begin{align*}
    p(a|a) = 0.4, p(b|a) = 0.3, p(c|a) = 0.3,
\end{align*}
in which case, we keep $(a,a)$ with $K=1$. With $K=2$, we should check also
\begin{align*}
    p(a|c) = 0.45, p(b|c) = 0.45, p(c|c) = 0.1,
\end{align*}
from which we maintain the hypotheses $(c,a)$ and $(c,b)$ ($0.45 \times 0.45$
and $0.45 \times 0.45$, respectively.) Note that with $K=2$,
we have discarded $(a,a)$.

Now, the greedy search ends by computing the last conditional probabilities:
\begin{align*}
    p(a|a,a) = 0.9, p(b|a,a) = 0.05, p(c|a,a) = 0.05.
\end{align*}
The final verdict from the greedy search is therefore $(a,a,a)$ with its
probability being $0.5 \times 0.4 \times 0.9 = 0.18$.

What happens with the beam search having $K=2$? We need to check the following
conditional probabilities:
\begin{align*}
    &p(a|c,a) = 0.7, p(b|c,a) = 0.2, p(c|c,a) =0.1 \\
    &p(a|c,b) = 0.4, p(b|c,b) = 0.0, p(c|c,b) =0.6 \\
\end{align*}
From here we consider $(c,a,a)$ and $(c,b,c)$ with the corresponding
probabilities $0.45 \times 0.45 \times 0.7=0.14175$ and $0.45 \times 0.45 \times
0.6 = 0.1215$. Among these two, $(c,a,a)$ is finally chosen, due to its higher
probability than that of $(c,b,c)$.

In summary, the greedy search found $(a,a,a)$ whose probability is
\begin{align*}
    p(a,a,a) = 0.18,
\end{align*}
and the beam search with $K=2$ found $(c,a,a)$ whose probability is
\begin{align*}
    p(c,a,a) = 0.14175.
\end{align*}
Even with a larger $K$, the beam search found a worse translation!

Now, clearly, what one can do is to set the maximum beam width $\bar{K}$ and try
with all possible $1 \leq K \leq \bar{K}$. Among the translations given by
$\bar{K}$ beam search procedures, the best translation can be selected based on
their corresponding probabilities. From the point of view of computational
complexity, this is perhaps the best approach to upper-bound the worst-case
memory consumption. Doing the beam search once with $\bar{K}$ or multiple beam
searches with $K=1,\ldots,\bar{K}$ are equivalent in terms of memory
consumption, i.e., both are $O(K|V|)$. Furthermore, the worst-case computation
is $O(K|V|)$ (assuming a constant time computation for computing each
conditional probability.) In practice however, the constant in front of $K|V|$
does matter, and we often choose $K$ based on the translation quality of the
validation set, after trying a number of values--$\left\{ 1, 2, 4, 8, 16
\right\}$.

If you're interested in how to improve beam search by backtracking so that the
beam search becomes {\em complete}, refer to, e.g.,
\cite{furcy2005limited,zhou2005beam}. If you're interested in general search
strategies, refer to \cite{russell1995artificial}.  Also, in the context of
statistical machine translation, it is useful to read \cite{koehn2004pharaoh}.

\section{Attention-based Neural Machine Translation}
\label{sec:att_mt}

One important property of the simple encoder-decoder model for neural machine
translation (from Sec.~\ref{sec:nmt_simple}) is that a whole source sentence is
compressed into a single real-valued vector $\vc$. This sounds okay, since the
space of all possible source sentences is {\em countable}, while the context
vector space $\left[ -1, 1 \right]^d$ is {\em uncountable}. There exists a
mapping from this sentence space to the context vector space, and all we need to
ensure is that training the simple encoder-decoder model finds this mapping.
This is conditioned on the assumption that the hypothesis space\footnote{
    See Sec.~\ref{sec:hypothesis_space}.
}
defined by the
model architecture--the number of hidden units and parameters-- includes this
mapping from any source sentence to a context vector.

Unfortunately, considering the complexity of any natural language sentence, it
is quite easy to guess that this mapping must be highly nonlinear and will require
a huge encoder, and consequently, a huge decoder to map back from a context
vector to a target sentence. In fact, this fact was empirically validated last
year (2014), when the almost identical models from two groups
\cite{sutskever2014sequence,cho2014properties} showed vastly different
performances on the same English--French translation task. The only difference
there was that the authors of \cite{sutskever2014sequence} used a {\em much}
larger model than the authors of \cite{cho2014properties} did.

At a more fundamental level there's a question of whether a natural language
sentence {\em should} be fully represented as a single vector. For instance,
there is now a famous quote by Prof. Raymond Mooney\footnote{
    \url{https://www.cs.utexas.edu/~mooney/}
} 
of the University of Texas at Austin: ``You can't cram the meaning of a whole
\%\&!\$\# sentence into a single \$\&!\#* vector!''\footnote{
    \url{http://nlpers.blogspot.com/2014/09/amr-not-semantics-but-close-maybe.html}
} Though, our goal is not in answering this fundamental question from
linguistics.

Our goal is rather to investigate the possibility of avoiding this situation of
having to learn a highly nonlinear, complex mapping from a source sentence to a
single vector. The question we are more interested in is whether {\em there
    exists a neural network that can handle a variable-length sentence by
building a variable-length representation of it}. Especially, we are interested
in whether we can build a neural machine translation system that can exploit a
variable-length context representation.

\paragraph{Variable-length Context Representation}

In the simple encoder-decoder model, a source sentence, regardless of its
length, was mapped to a single context vector by a recurrent neural network:
\begin{align*}
    \vh_t = \phi_{\enc}\left( \vh_{t-1}, \mE_x^\top \vx_t \right).
\end{align*}
See Eq.~\eqref{eq:nmt_encoder} and the surrounding text for more details.

Instead, here we will encode a source sentence $X=(x_1, x_2, \ldots, x_{T_x})$
with a {\em set $C$ of context vectors} $\vh_t$'s. This is achieved by having
two recurrent neural networks rather than a single recurrent neural networks, as
in the simple encoder-decoder model. The first recurrent neural network, to
which we will refer as a {\em forward recurrent neural network}, reads the
source sentence as usual and results in a set of {\em forward memory states}
$\overrightarrow{\vh}_t$, for $t=1,\ldots,T_x$. The second recurrent neural
network, a {\em reverse recurrent neural network}, reads the source sentence
in a reverse order, starting from $x_{T_x}$ to $x_1$. This reverse network will
output a sequence of {\em reverse memory states} $\overleftarrow{\vh}_t$, for
$t=1, \ldots, T_x$.

\begin{figure}[t]
    \centering
    \centering
    \includegraphics[width=0.7\columnwidth]{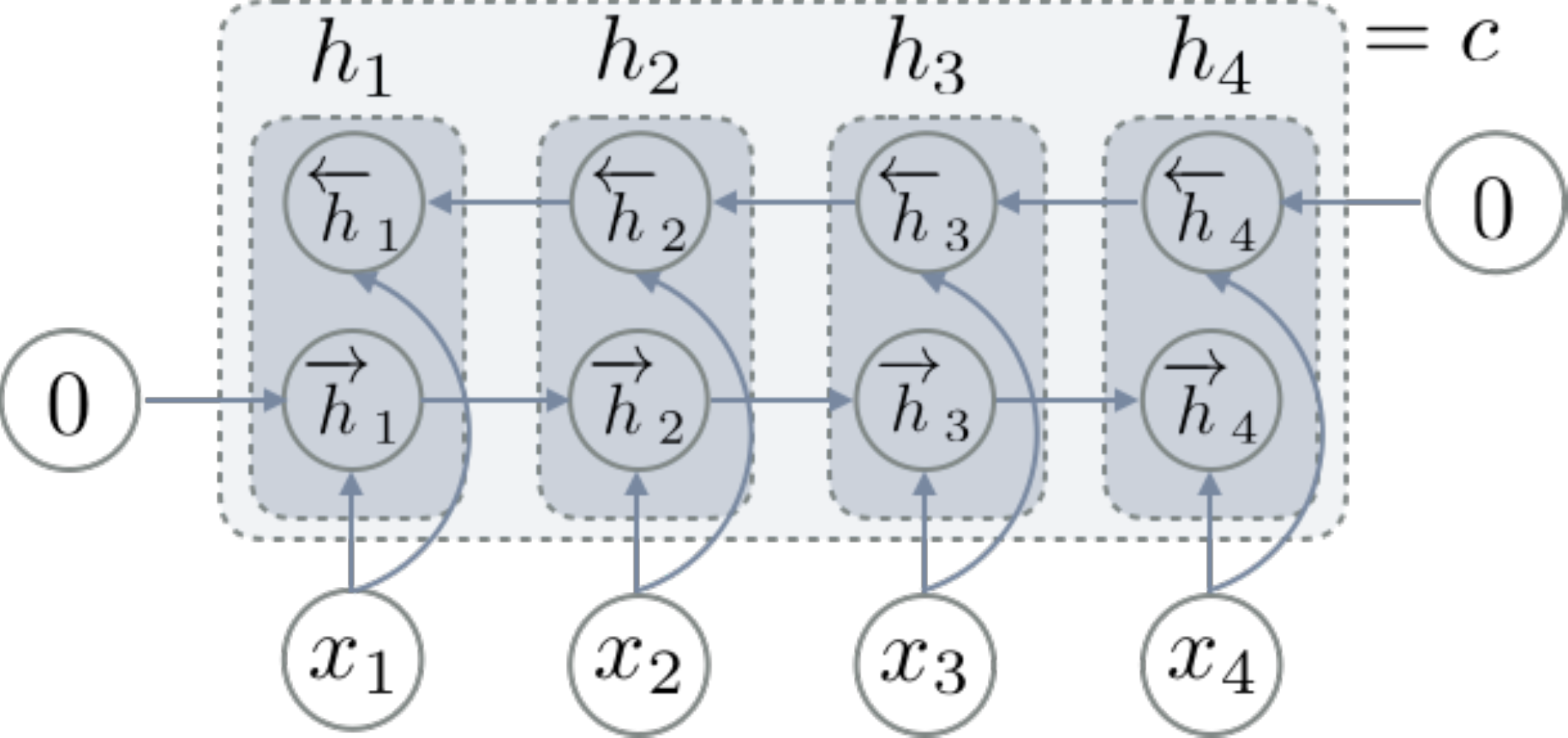}

    \caption{An encoder with a bidirectional recurrent neural network}
    \label{fig:bienc}
\end{figure}

For each $x_t$, we will concatenate $\overrightarrow{\vh}_t$ and
$\overleftarrow{\vh}_t$ to form a {\em context-dependent vector} $\vh_t$:
\begin{align}
    \label{eq:cd_rep}
    \vh_t = \left[ \begin{array}{c}
            \overrightarrow{\vh}_t \\
            \overleftarrow{\vh}_t
    \end{array} \right]
\end{align}
We will form a {\em context set} with these context-dependent vectors $c=\left\{
\vh_1, \vh_2, \ldots, \vh_{T_x}\right\}$. See Fig.~\ref{fig:bienc} for the
graphical illustration of this process.

Now, why is $\vh_t$ a {\em context-dependent vector}? We should look at what the
input was to a function that computed $\vh_t$. The first half of $\vh_t$,
$\overrightarrow{\vh}_t$, was computed by
\begin{align*}
    \overrightarrow{\vh}_t = \phi_{\fenc}\left( \phi_{\fenc}\left( \cdots,
    \mE_x^\top \vx_{t-1}\right), \mE_x^\top \vx_{t}\right),
\end{align*}
where $\phi_{\fenc}$ is a forward recurrent activation function. From this we
see that $\overrightarrow{\vh}_t$ was computed by all the source words up to
$t$,
i.e., $\vx_{\leq t}$. Similarly, 
\begin{align*}
    \overleftarrow{\vh}_t = \phi_{\renc}\left( \phi_{\renc}\left( \cdots,
    \mE_x^\top \vx_{t+1}\right), \mE_x^\top \vx_{t}\right),
\end{align*}
where $\phi_{\renc}$ is a reverse recurrent activation function, and
$\overleftarrow{\vh}_t$ depends on all the source words from $t$ to the end,
i.e., $\vx_{\geq t}$. 

In summary, $\vh_{t} =\left[ \overrightarrow{\vh}_t^\top;
\overleftarrow{\vh}_t^\top \right]^\top$ is a vector representation of the
$t$-th word, $x_t$, with respect to all the other words in the source sentence.
This is why $\vh_t$ is a context-dependent representation. But, then, what is
the difference among all those context-dependent representations $\left\{
    \vh_1, \ldots, \vh_{T_x}
\right\}$? We will discuss this shortly.

\paragraph{Decoder with Attention Mechanism}

Before anything let us think of what the memory state $\vz_t$ of the decoder
(from Eq.~\eqref{eq:nmt_decoder}) does:
\begin{align*}
    \vz_t = \phi_{\dec}\left( 
        \phi_{\dec}\left(
            \phi_{\dec}\left(
                \cdots, 
                \left[ \mE_y^\top \vy_{t-3}; \vc \right]
            \right),
            \left[ \mE_y^\top \vy_{t-2}; \vc \right]
        \right)
        \left[ \mE_y^\top \vy_{t-1}; \vc \right]
    \right)
\end{align*} 

It is computed based on all the generated target words so far $(\tilde{y}_1,
\tilde{y}_2, \ldots, \tilde{y}_{t-1})$ and the context vector\footnote{
    We will shortly switch to using a context {\em set} instead.
}
$\vc$ which is the summary of the source sentence. The very reason why I
designed the decoder in this way is so that the memory state $\vz_t$ is
informative of which target word should be generated at time $t$ {\em after}
generating the first $t-1$ target words {\em given} the source sentence. In
order to do so, $\vz_t$ must encode {\em what have been translated so far} among
the words that are supposed to be translated (which is encoded in the context
vector $\vc$.) Let's keep this in mind.

In order to compute the new memory state $\vz_t$ with a {\em context set}
$C=\left\{ \vh_1, \vh_2, \ldots, \vh_{T_x}\right\}$,
we must first get one vector out of $T_x$ context vectors. Why is this
necessary? Because we cannot have an infinitely large number of parameters to
cope with any number of context vectors. Then, how can we get a single vector
from an unspecified number of context vectors $\vh_t$'s?

\begin{figure}[t]
    \centering
    \centering
    \includegraphics[width=0.7\columnwidth]{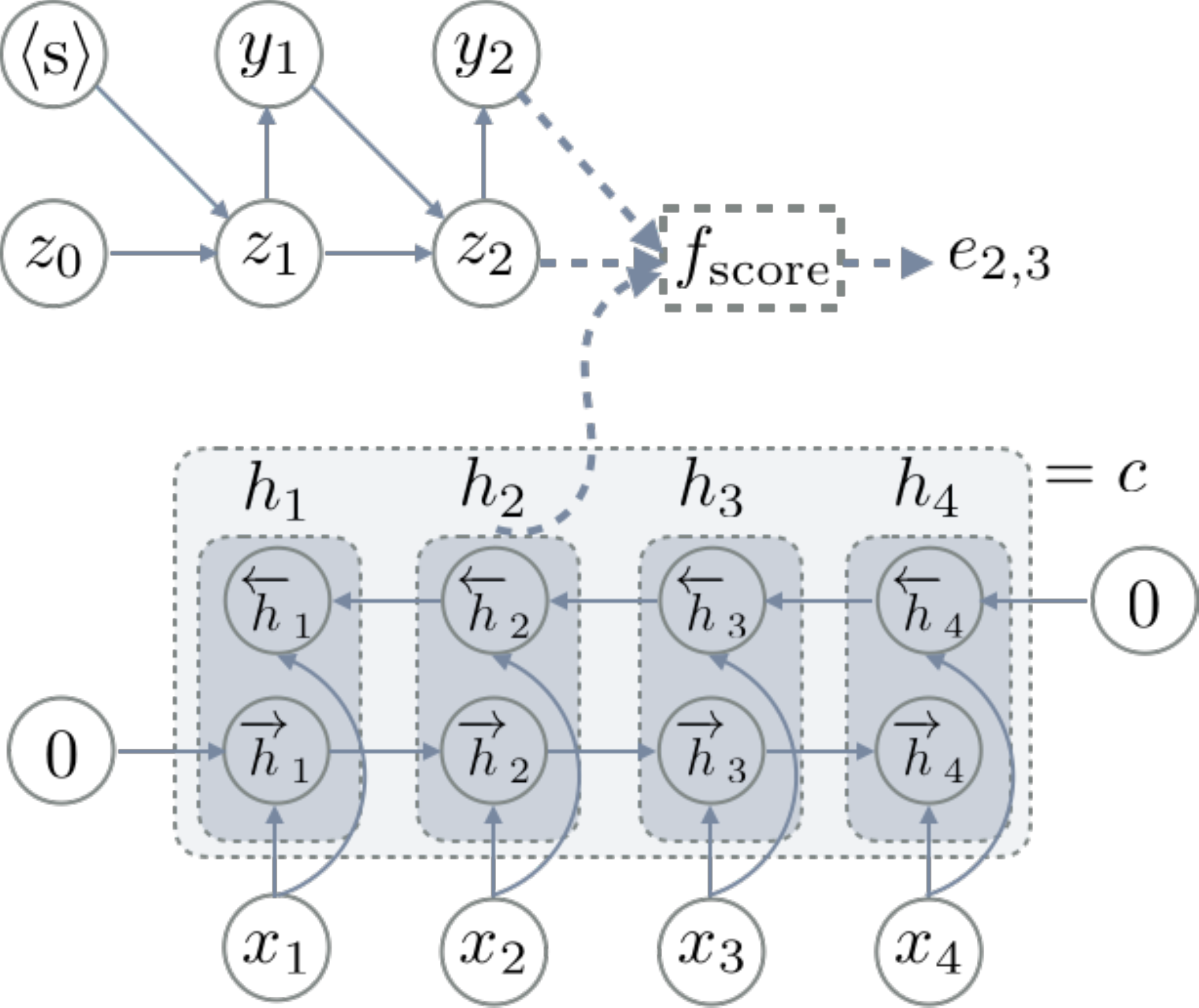}

    \caption{Illustration of how the relevance score $e_{2,3}$ of the second
    context vector $\vh_2$ at time step $3$ (dashed curves and box.)}
    \label{fig:att1}
\end{figure}

First, let us score each context vector $\vh_j$ ($j=1, \ldots, T_x$) based on
how {\em relevant} it is for translating a next target word. This scoring needs
to be based on (1) the previous memory state $\vz_{t-1}$ which summarizes what
has been translated up to the $(t-2)$-th word\footnote{
    Think of why this is only up to the $(t-2)$-th word not up to the $(t-1)$-th
    one.
}, (2) the previously generated target word $\tilde{y}_{t-1}$, and (3) the
$j$-th context vector $\vh_j$: 
\begin{align}
    \label{eq:nmt_score}
    e_{j, t} = f_{\text{score}}(\vz_{t-1}, \mE_y^\top \tilde{y}_{t-1}, \vh_j).
\end{align}
Conceptually, the score $e_{j,t}$ will be computed by {\em comparing}
$(\vz_{t-1}, \tilde{y}_{t-1})$ with the context vector $\vc_j$. See
Fig.~\ref{fig:att1} for graphical illustration.

Once the scores for all the context vectors $\vh_j$'s ($j=1, \ldots, T_x$) are
computed by $f_{\text{score}}$, we normalize them with a softmax function:
\begin{align}
    \label{eq:att_weight}
    \alpha_{j, t} = \frac{\exp(e_{j,t})}{\sum_{j'=1}^{T_x} \exp(e_{j',t})}.
\end{align}
We call these normalized scores the {\em attention weights}, as they correspond
to how much the {\em decoder attends} to each of the context vectors. This whole
process of computing the attention weights is often referred to as an {\em
attention mechanism} (see, e.g., \cite{cho2015describing}.)

\begin{figure}[t]
    \centering
    \centering
    \includegraphics[width=0.7\columnwidth]{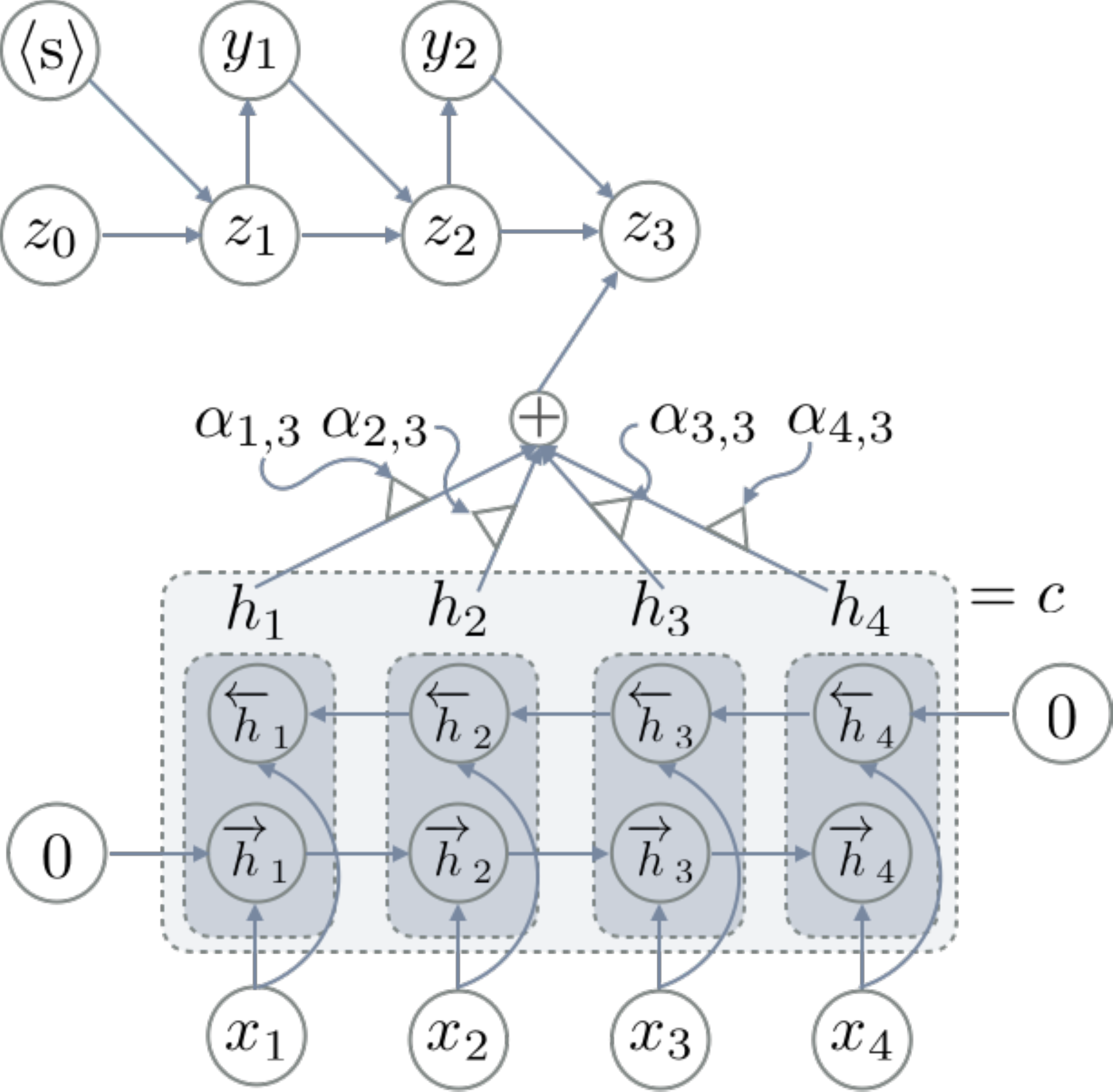}

    \caption{Computing the new memory state $\vz_t$ of the decoder based on the
        previous memory state $\vz_{t-1}$, the previous target word
    $\tilde{y}_{t-1}$ and the weighted average of context vectors according to
the attention weights.}
    \label{fig:att2}
\end{figure}

We take the weighted average of the context vectors with these attention
weights:
\begin{align}
    \label{eq:td_context}
    \vc_t = \sum_{j=1}^{T_x} \alpha_{j, t} \vh_j
\end{align}
This weighted average is used to compute the new memory state $\vz_t$ of the
decoder, which is identical to the decoder's update equation from the simple
encoder-decoder model (see Eq.~\eqref{eq:nmt_decoder}) except that $\vc_t$ is
used instead of $\vc$ ((a) in the equation below):
\begin{align*}
    \vz_t = \phi_{\dec}\left( \vz_{t-1}, \left[ \mE_y^\top \vy_{t-1};
            \underbrace{\vc_t}_{\text{(a)}}
    \right]\right)
\end{align*}
See Fig.~\ref{fig:att2} for the graphical illustration of how it works.

Given the new memory state $\vz_t$ of the decoder, the output probabilities of
all the target words in a vocabulary happen without any change from the simple
encoder-decoder model in Sec.~\ref{sec:nmt_simple}.

We will call this model, which has a bidirectional recurrent neural network as
an encoder and a decoder with the attention mechanism, an {\em attention-based
encoder-decoder model}. This approach was proposed last year (2014) in the
context of machine translation in \cite{bahdanau2014neural} and has been studied
extensively in \cite{luong2015effective}.

\subsection{What does the Attention Mechanism do?}

One important thing to notice is that this attention-based encoder-decoder model
can be reduced to the simple encoder-decoder model easily. This happens when the
attention mechanism $f_{\text{score}}$ in Eq.~\eqref{eq:nmt_score} returns a
constant regardless of its input. When this happens, the context vector $\vc_t$
at each time step $t$ (see Eq.~\eqref{eq:td_context}) is same for all the time
steps $t=1,\ldots,T_y$:
\begin{align*}
    \vc_t = \frac{1}{T_x} \sum_{j=1}^{T_x} \vh_j.
\end{align*}
The encoder effectively maps the whole input sentence into a single vector,
which was at the core of the simple encoder-decoder model from
Sec.~\ref{sec:nmt_simple}. 

This is not the only situation in which this type of behaviour happens. Another
possible scenario is for the encoder to make the last memory states,
$\overrightarrow{\vh}_{T_x}$ and $\overleftarrow{\vh}_1$, of the forward and
reverse recurrent neural networks to have a special mark telling that these are
the last states. The attention mechanism then can exploit this to assign a large
score to these two memory states (but still constant across time $t$.) This will
become even closer to the simple encoder-decoder model.

The question is how we can avoid these degenerate cases. Or, {\em is it
necessary for us to explicitly make these degenerate cases unlikely?} Of course,
there is no single answer to this question. Let me give you my answer, which may
differ from others' answer: {\em no}.

The goal of introducing a novel network architecture is to guide a model
according to our intuition or scientific observation so that it will do a better
job at a target task. In our case, the attention mechanism was introduced based
on our observation, and some intuition, that it is not desirable to ask the
encoder to compress a whole source sentence into a single vector. 

This incorporation of {\em prior knowledge} however should not put a {\em hard
constraint}. We give a model a possibility of exploiting this prior knowledge,
but should not force the model to use this prior knowledge exclusively. As this
prior knowledge, based on our observation of a small portion of data, is not
likely to be true in general, the model must be able to ignore this, if the data
does not exhibit the underlying structure corresponding to this prior knowledge.
In this case of attention-based encoder-decoder model, the existence of those
degenerate cases above is a direct evidence of what this attention-based model
can do, if there is no such underlying structure present in the data. 

Then, a natural next question is whether there are such structures that can be
well exploited by this attention mechanism in real data. If we train this
attention-based encoder-decoder model on the parallel corpora we discussed
earlier in Sec.~\ref{sec:parallel_corpora}, what kind of structure does this
attention-based model learn?

In order to answer this question, we must first realize that we can easily
visualize what is happening inside this attention-based model. First, note that
given a pair of source $X$ and target $Y$ sentences,\footnote{
    Note that if you're given only a source sentence, you can let the model
    translate and align simultaneously.
}
the attention-based model
computes an {\em alignment} matrix $A \in \left[ 0, 1\right]^{|X| \times |Y|}$:
\begin{align*}
    A = \left[ 
        \begin{array}{c c c c}
            \alpha_{1,1} & \alpha_{1,2} & \cdots & \alpha_{1, |Y|} \\
            \alpha_{2,1} & \alpha_{2,2} & \cdots & \alpha_{2, |Y|} \\
            \vdots & \vdots & \ddots & \vdots \\
            \alpha_{|X|,1} & \alpha_{|X|, 2} & \cdots & \alpha_{|X|, |Y|} 
        \end{array}
    \right],
\end{align*}
where $\alpha_{j,t}$ is defined in Eq.~\eqref{eq:att_weight}.

Each column $\va_t$ of this alignment matrix $A$ is how well each source word
(based on its {\em context-dependent vector representation} from
Eq.~\eqref{eq:cd_rep}) is aligned to the $t$-th target word. Each row $\vb_{j}$
similarly shows how well each target word is aligned to the content-dependent
vector of the $j$-th source word. In other words, we can simply draw the
alignment matrix $A$ as if it were a gray scale 2-D image. 

\begin{figure}[ht]
    \centering
    \begin{minipage}[b]{0.48\textwidth}
        \raggedleft
        \includegraphics[width=1.\columnwidth]{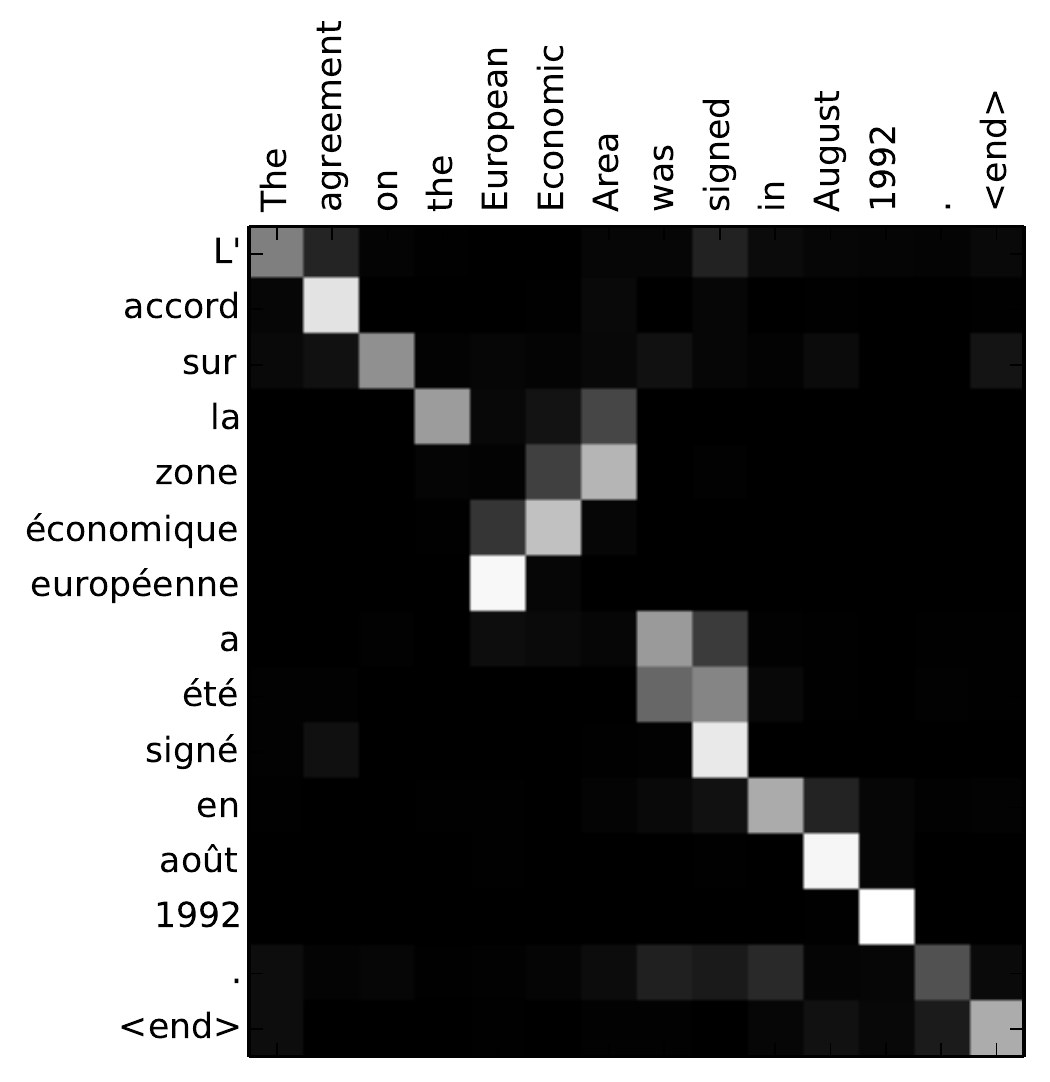}
    \end{minipage}
    \hfill
    \begin{minipage}[b]{0.48\textwidth}
        \raggedleft
        \includegraphics[width=1.\columnwidth]{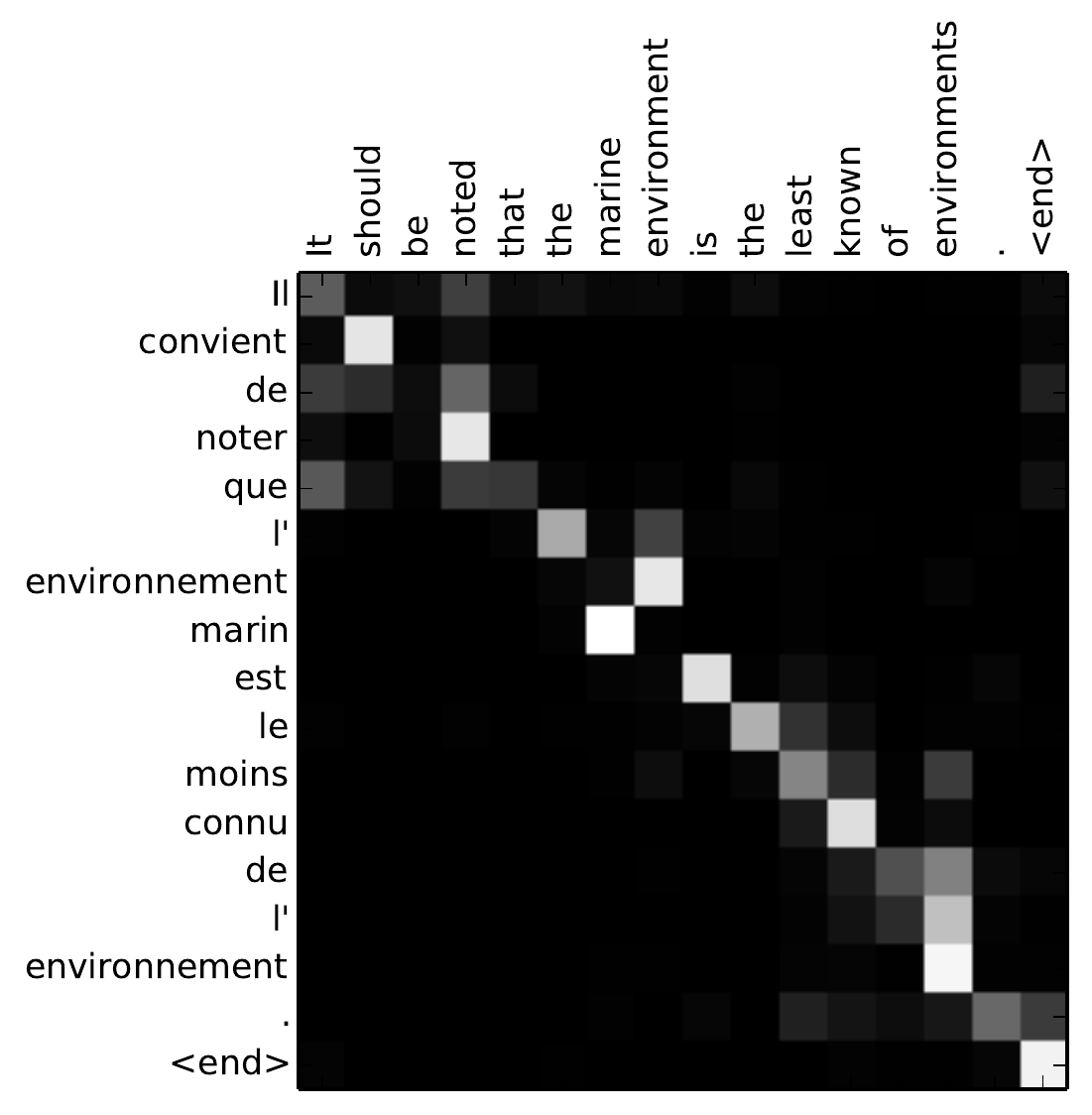}
    \end{minipage}

    \begin{minipage}{0.48\textwidth}
        \centering
        (a)
    \end{minipage}
    \hfill
    \begin{minipage}{0.48\textwidth}
        \centering
        (b)
    \end{minipage}

    \begin{minipage}[b]{0.48\textwidth}
        \raggedleft
        \includegraphics[width=1.\columnwidth]{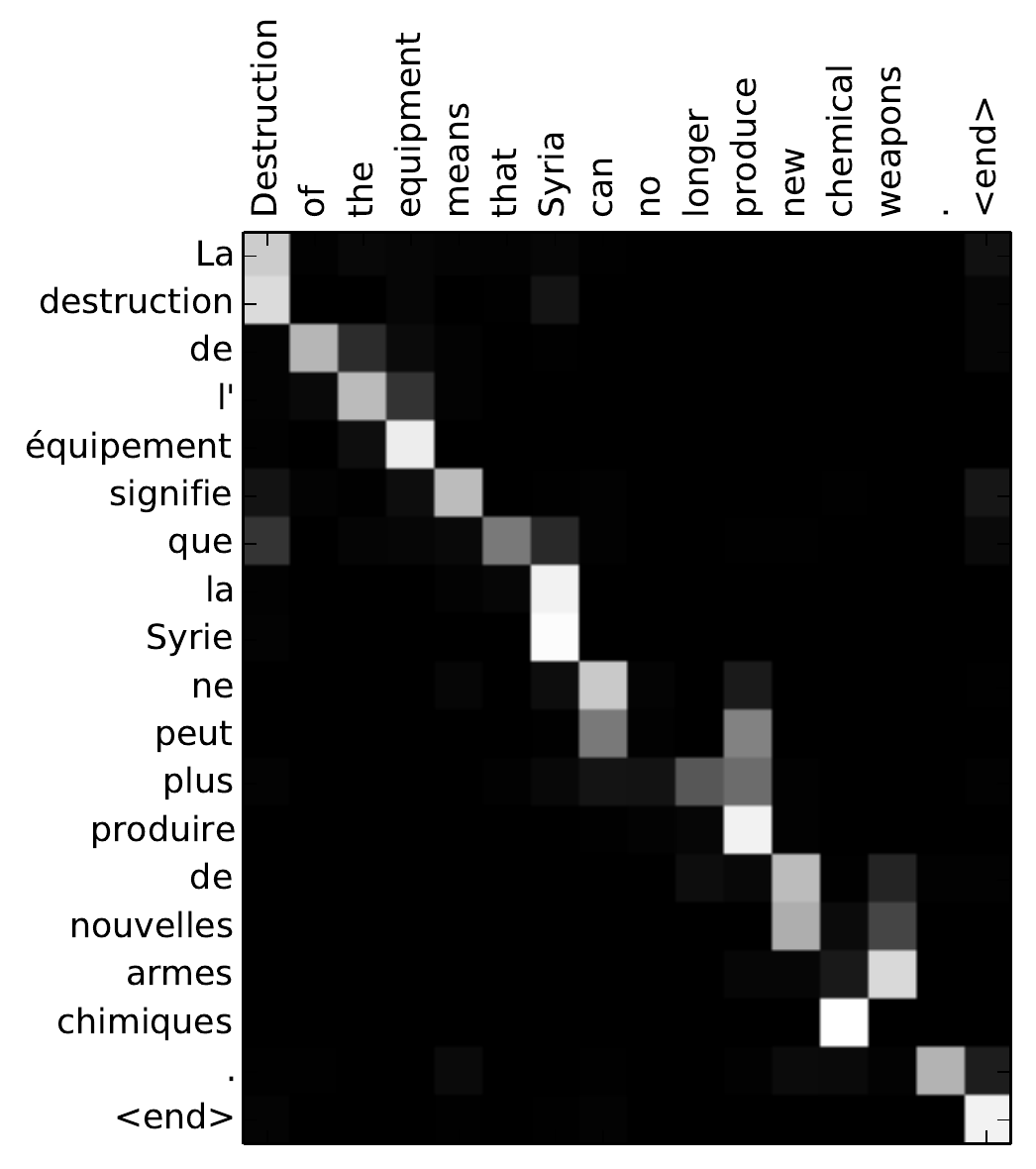}
    \end{minipage}
    \hfill
    \begin{minipage}[b]{0.48\textwidth}
        \raggedleft
        \includegraphics[width=1.\columnwidth]{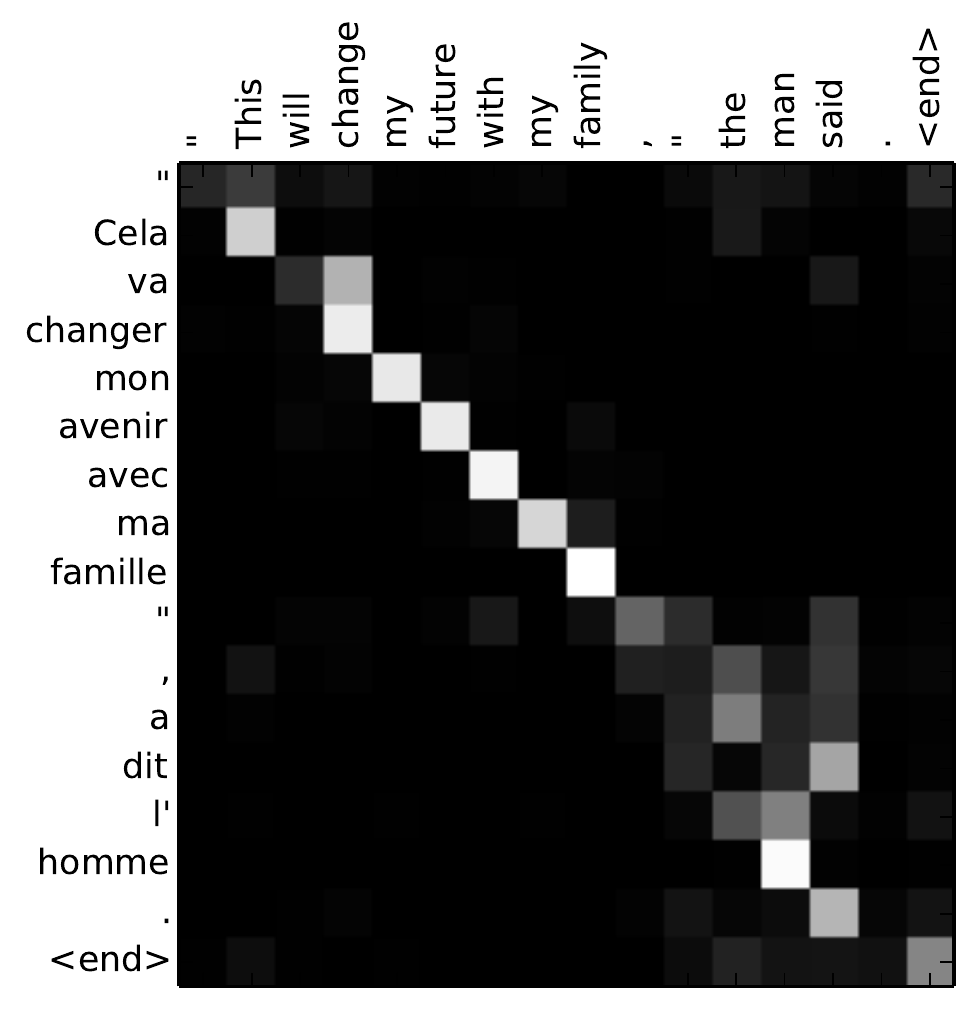}
    \end{minipage}

    \begin{minipage}{0.48\textwidth}
        \centering
        (c)
    \end{minipage}
    \hfill
    \begin{minipage}{0.48\textwidth}
        \centering
        (d)
    \end{minipage}

    \caption{
        Visualizations of the four sample alignment matrices. The alignment
        matrices were computed from an attention-based translation model trained
        to translate a sentence in English to French.
        Reprinted from
        \cite{bahdanau2014neural}.
    }
    \label{fig:alignment}
\end{figure}

In Fig.~\ref{fig:alignment}, the visualization of four alignment matrices is
presented. It is quite clear, especially to a French-English bilingual speaker,
that the model indeed captured the underlying structure of word/phrase mapping
between two languages. For instance, focus on ``European Economic Area'' in
Fig.~\ref{fig:alignment}~(a). The model correctly noticed that ``Area''
corresponds to ``zone'', ``Economic'' to ``\'economique'', and ``European'' to
``europ\'eenne'', without any supervision about this type of alignment.  

This is nice to see that the model was able to notice these regularities from
data without any explicit supervision. However, the goal of introducing the
attention mechanism was not to get these pretty figures. After all, our goal is
not to build an interpretable model, but a model that is predictive of the
correct output given an input (see Chapter~\ref{chap:intro} and
\cite{breiman2001statistical}.) In this regard, how much does the introduction
of the attention mechanism help?

In \cite{bahdanau2014neural}, the attention-based encoder-decoder model was
compared against the simple encoder-decoder model in the task of English-French
translation. They observed the relative improvement of up to 60\% (in terms of
BLEU, see Sec.~\ref{sec:bleu},) as shown in Table~\ref{tab:att_nmt}.
Furthermore, by using some of the latest techniques, such as handling large
vocabularies \cite{jean2014using}, building a vocabulary of subword units
\cite{sennrich2015neural} and variants of the attention mechanism
\cite{luong2015effective}, it has been found possible to achieve a better translation
quality with neural machine translation than the existing state-of-the-art
translation systems.

\begin{table}[ht]
    \centering
    \begin{tabular}{c|c|c}
    Model & BLEU & Rel. Improvement \\
    \hline
    \hline
    Simple Enc--Dec & 17.82 &  -- \\
    Attention-based Enc--Dec & 28.45 & +59.7\% \\
    Attention-based Enc--Dec (LV) & 34.11 & +90.7\% \\
    Attention-based Enc--Dec (LV)$^\star$ & {\bf 37.19} & {\bf +106.0\%} \\
    \hline
    State-of-the-art SMT$^\circ$ & 37.03 & -- 
    \end{tabular}
    \caption{The translation performances and the relative improvements over the
        simple encoder-decoder model on an English-to-French translation task
        (WMT'14),
    measured by BLEU \cite{bahdanau2014neural,jean2014using}. 
$\star$: an ensemble of multiple attention-based models. $\circ$: the
state-of-the-art phrase-based statistical machine translation
system~\cite{Durrani2014}.}
    \label{tab:att_nmt}
\end{table}

\section{Warren Weaver's Memorandum}

In 1949 Warren Weaver\footnote{
    Yes, this is the very same Weaver after which the building of the Courant
    Institute of Mathematical Sciences has been named.
} wrote a memorandum, titled $\left<\text{Translation}\right>$ on machine
translation \cite{weaver1955translation}. Although this text was written way
before computers have become ubiquitous,\footnote{
    Although Weaver talks about modern computers over and over in his
    memorandum, what he refers to is not exactly what we think of computers as
    these days.
} there are many interesting ideas that are closely related to what we have
discussed so far in this chapter. Let us go over some parts of the Weaver's
memorandum and see how the ideas there corresponds to modern-day machine
translation.

\paragraph{Necessity of Linguistic Knowledge}

Weaver talks about a distinguished mathematician P who was surprised by his
colleague. His colleague ``had an amateur interest in cryptography'', and one
day presented P his method to ``decipher'' an encrypted Turkish text
successfully. ``The most important point'', according to Weaver, from this
instance is that ``the decoding was done by someone who did not know Turkish.''
Now, this sounds familiar, doesn't it? 

As long as there was a parallel corpus, we are able to use neural machine
translation models, described throughout this chapter, without ever caring about
which languages we are training a model to translate between.  Especially if we
decide to consider each sentence as a sequence of {\em characters},\footnote{
    In fact, only very recently people have started investigating the
    possibility of building a machine translation system based on character
    sequences \cite{ling2015character}. This has been made possible due to the
    recent success of neural machine translation.
}
there is almost no need for any linguistic knowledge when building these neural
machine translation systems. 

This lack of necessity for linguistic knowledge is not new. In fact, the most
widely studied and used machine translation approach, which is (count-based)
statistical machine translation
\cite{brown1990statistical,koehn2003statistical}, does not require any prior
knowledge about source and target languages. All it needs is a large corpus.

\paragraph{Importance of Context}

Recall from Sec.~\ref{sec:att_mt} that the encoder of an attention-based neural
machine translation uses a {\em bidirectional} recurrent neural network in order
to obtain a context set. Each vector in the context set was considered a {\em
context-dependent vector}, as it represents what the center word means with
respect to all the surrounding words. This context dependency is a necessary
component in making the whole attention-based neural machine translation, as it
helps disambiguating the meaning of each word and also distinguishing multiple
occurrences of a single word by their context.

Weaver discusses this extensively in Sec.~3--4 in his memorandum. First, to
Weaver, it was ``amply clear that a translation procedure that does little more
than handle a one-to-one correspondence of words can not hope to be useful .. in
which the problems of .. multiple meanings .. are frequent.'' In other words, it
is simply not possible to look at each word separately from surrounding words
(or context) and translate it to a corresponding target word, because there is
uncertainty in the meaning of the source word which can only be resolved by
taking into account its context.

So, what does Weaver propose in order to address this issue? He proposes in
Sec.~5 that if ``one can see not only the central word in question, but also say
$N$ words on either side, then if [sic] $N$ is large enough one can
unambiguously decide the meaning of the central word.'' If we consider only a
single sentence and take the infinite limit of $N \to \infty$, we see that what
Weaver refers to is exactly the bidirectional recurrent neural network used by
the encoder of the attention-based translation system.  Furthermore, we see that
the continuous bag-of-words language model, or Markov random field based
language model, from Sec.~\ref{sec:cbow} exactly does what Weaver proposed by
setting $N$ to a finite number.

In Sec.~\ref{sec:data_sparsity}, we talked about the issue of data sparsity, and
how it is desirable to have a larger $N$ but it's often not a good idea
statistically to do so. Weaver was also worried about this by saying that ``it
would hardly be practical to do this by means of a generalized dictionary which
contains all possible phases [sic] $2N+1$ words long; for the number of such
phases [sic] is horrifying.'' We learned that this issue of data sparsity can be
largely avoided by adopting a fully parametric approach instead of a
table-based approach in Sec.~\ref{sec:nlm}.

\paragraph{Common base of human communications}

Weaver suggested in the last section of his memorandum that ``perhaps the way''
for translation ``is to descend, from each language, down to the common base of
human communication -- the real but as yet undiscovered universal language --
and then re-emerge by whatever particular route is convenient.'' He specifically
talked about a ``universal language'', and this makes me wonder if we can
consider the memory state of the recurrent neural networks (both of the encoder
and decoder) as this kind of intermediate language. This intermediate language
radically departs from our common notion of natural languages. Unlike
conventional languages, it does not use discrete symbols, but uses continuous
vectors. This use of continuous vectors allows us to use simple arithmetics to
manipulate the meaning, as well as its surface realization.\footnote{
    If you find this view too radical or fascinating, I suggest you to look at
    the presentation slides by Geoff Hinton at
    \url{https://drive.google.com/file/d/0B16RwCMQqrtdMWFaeThBTC1mZkk/view?usp=sharing}
}

This view may sound radical, considering that what we've discussed so far has
been confined to translating from one language to another. After all, this
universal language of ours is very specific to only a single source language
with respect to a single target language. This is however not a constraint on
the neural machine translation by design, but simply a consequence of our having
focused on this specific case. 

Indeed, in this year (2015), researchers have begun to report that it is
possible to build a neural machine translation model that considers multiple
languages, and even further multiple tasks \cite{dong2015multi,luong2015}. More
works in this line are expected, and it will be interesting to see if Weaver's
prediction again turns out to be true.

\chapter{Final Words}

Let me wrap up this lecture note by describing some aspects of natural language
understanding with distributed representations that I have not discussed in this
course. These are the topics that I would have spent time on, had the course
been scheduled to last twice the duration as it is now. Afterward, I will
finalize this whole lecture note with a short summary.

\section{Multimedia Description Generation as Translation}

Those who have followed this course closely so far must have noticed that the
neural machine translation model described in the previous chapter is quite
general in the sense that the input to the model does not have to be a sentence.
In the case of the simple encoder-decoder model from Sec.~\ref{sec:nmt_simple},
it is clear that any type of input $X$ can be used instead of a sentence, as
long as there is a feature extractor that returns the vector representation
$\vc$ of the input.

And, fortunately, we already learned how to build a feature extractor throughout
this course. Almost every single model (that is, a neural network in our case)
converts an input into a continuous vector. Let us take a multilayer perceptron
from Sec.~\ref{sec:mlp} as an example. Any classifier built as a multilayer
perceptron can be considered as a two-stage process (see
Sec.~\ref{sec:example2}.) First, the feature vector of the input is extracted
(see Eq.~\eqref{eq:phi}):
\begin{align*}
    \phi(x) = \sigma(u x + c).
\end{align*}
The extracted feature vector $\phi(x)$ is then affine-transformed, followed by
softmax function. This results in a conditional distribution over all possible
labels (see Eq.~\eqref{eq:multiclass}.)

This means that we can make the simple encoder-decoder model to work with
non-language input simply by replacing the recurrent neural network based
encoder with the feature extraction stage of the multilayer perceptron.
Furthermore, it is possible to {\em pretrain} this feature extractor by training
the whole multilayer perceptron on a separate classification dataset.\footnote{
    This way of using a feature extractor pretrained from another network has
    become a {\em de facto} standard in many of the computer vision tasks
    \cite{sermanet2013overfeat}. This
    is also closely related to semi-supervised learning with pretrained word
    embeddings which we discussed in Sec.~\ref{sec:semi_emb}. In that case, it
    was only the first input layer that was pretrained and used later (see
    Eqs.~\eqref{eq:nlm_first_layer}--\eqref{eq:word_emb}.)
}

This approach of using the encoder-decoder model for describing non-language
input has become popular in recent years (especially, 2014 and 2015,) and has
been applied to many applications, including image/video description generation
and speech recognition. For an extensive list of these applications, I refer the
readers to a recent review article by Cho et al.~\cite{cho2015describing}.

\paragraph{Example: Image Caption Generation}

Let me take as an example the task of image caption generation.  The possibility
of using the encoder-decoder model for image caption generation was noticed by
several research groups (almost simultaneously) last year (2014)
\cite{Kiros-et-al-ICML2014,vinyals2014show,Karpathy+Li-arxiv2014,Mao+al-arxiv2014,Donahue-et-al-arxiv2014,Fang-et-al-arxiv2014,Chen+Zitnick-arxiv2014}.\footnote{
    I must however make a note that Kiros et al.~\cite{Kiros-et-al-ICML2014}
    proposed a fully neural network based image caption generation earlier than
    all the others cited here did.
} The success of neural machine translation in \cite{sutskever2014sequence} and
earlier success of deep convolutional network on object recognition (see, e.g.,
\cite{krizhevsky2012imagenet,simonyan2014very,szegedy2014going}) inspired them
the idea to use the deep convolutional network's feature extractor together with
the recurrent neural network decoder for the task of {\em image caption
generation}. 

\begin{wrapfigure}{l}{0.5\textwidth}
    \centering
    \includegraphics[width=0.5\columnwidth]{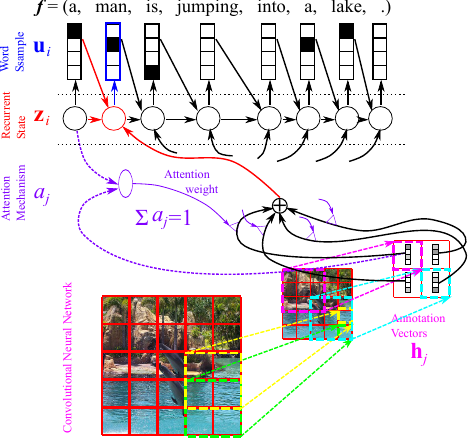}
    \caption{Image caption generation with the attention-based
    encoder-decoder model \cite{xu2015show}.}
    \label{fig:att_img_cap}
\end{wrapfigure}

Right after these, Xu et al.~\cite{xu2015show} realized that it is possible to
use the attention-based encoder-decoder model from Sec.~\ref{sec:att_mt} for
image caption generation. Unlike the simple model, the attention-based model
requires a context set instead of a context vector. The context set should
contain multiple context vectors, and each vector should represent a spatial
location with respect to the whole image, meaning each context vector is a {\em
spatially-localized, context-dependent image descriptor}. This was achieved by
using the last {\em convolutional} layer's activations of the pretrained deep
convolutional network instead of the last {\em fully-connected} layer's. See
Fig.~\ref{fig:att_img_cap} for graphical illustration of this approach.

These approaches based on neural networks, or in other words based on
distributed representations, have been successful at image caption generation.
Four out of five top rankers in the recent Microsoft CoCo Image Captioning
Challenge 2015\footnote{
    \url{http://mscoco.org/dataset/\#captions-leaderboard}
} 
were using variants of the neural encoder-decoder model, based on human
evaluation of the captions. 

\section{Language Understanding with World Knowledge}

In Sec.~\ref{sec:intro}, we talked about how we view natural languages as a
function. This function of natural language maps from a tuple of a speaker's
speech, a listener's mental state and {\em the surrounding world} to the
listener's reaction, often as a form of natural language response.
Unfortunately, in order to make it manageable, we decided to build a model that
approximates only a part of this true function. 

\paragraph{Immediate state of the surrounding world} 

In this course of action, one thing we have dropped out is the surrounding
world. The surrounding world may mean many different things. One of them is
the current state of the surrounding world. As an example, when I say ``look at
this cute llama,'' it is quite likely that the surrounding world at the current
state contains either an actual llama or at least a picture of a llama. A
listener then understands easily what a llama is even without having known what
a llama is in advance. By looking at the picture of llama, the listener makes a
mental note that the llama looks similar to a camel and therefore must be a
four-legged animal.

If the surrounding world is not taken into account, as we've been doing so far,
the listener can only generalize based on the context words. Just like how the
neural language model from Sec.~\ref{sec:nlm} generalized to unseen, or rarely
seen words, the listener can infer that ``llama'' must be a type of animal by
remembering that the phrase ``look at this cute'' has mainly been followed by an
animal such as ``cat'' or ``dog''. However, it is quite clear that ``look at
this cute'' is also followed by many other nouns, including ``baby'', ``book''
and so on.

The question is then how to exploit this. How can we incorporate, for instance,
vision information from the surrounding world into natural language
understanding? 

The simplest approach is to simply concatenate a word embedding vector (see
Eq.~\eqref{eq:word_emb}) and a corresponding image vector (obtained from an
existing feature extractor, see above) \cite{kiela2014learning}. This can be
applied to any existing language models such as neural language model (see
Sec.~\ref{sec:nlm}) and neural machine translation model (see
Chapter~\ref{chap:nmt}.) This approach gives a strong signal to the model the
similarities among different words based on the corresponding objects'
appearances. This approach of concatenating vectors of two different modalities,
e.g., language and vision, was earlier proposed in \cite{weston2010large}. 

A more sophisticated approach is to design and train a model to solve a task
that requires tight interaction between language and other modalities. As our
original goal is to build a {\em natural language function}, all we need to do
is to build a function approximator that takes as input both language and other
modalities. Recently, Antol et al.~\cite{VQA} built a large-scale dataset of
question-answer-image triplets, called visual question answering (VQA) for this
specific purpose. They have carefully built the dataset such that many, if not
most, questions can only be answered when the accompanying image is taken into
consideration. Any model that's able to solve the questions in this dataset well
will have to consider both language and vision.

\paragraph{Knowledge base: Lost in a library} 

So far, we have talked about incorporating an immediate state of the surrounding
world. However, our use of languages is more sophisticated. This is especially
apparent in written languages. Let us take an example of me writing this lecture
note. It is not the case where I simply sit and start writing the whole text
based purely on my mental state (with memory of my past research) and the
immediate surrounding world state (which has almost nothing to do with.) Rather,
a large part of this writing process is spent on going through various research
articles and books written by others in order to find relevant details of the
topic. 

In this case, the surrounding world is a database in which human knowledge is
stored. You can think of a library or the Internet. As the amount of knowledge
is simply too large to be memorized in the entirety, it is necessary for a
person to be able to search through the vast knowledge base. But, wait, what
does it have to do with natural language understanding?

Consider the case where the context phrase is ``Llama is a domesticated camelid
from''. Without access to the knowledge base, or in this specific instance,
access to Wikipedia, any language model can only say as much as that this
context phrase is likely followed by a name of some place. This is especially
true, if we assume that the training corpus did not mention ``llama'' at all.
However, if the language model is able to search Wikipedia and condition on its
search result, it suddenly becomes so obvious that this context phrase is
followed by ``South America'' or the name of any region on Andean mountain
rages. 

Although this may sound too complicated a task to incorporate into a neural
network, the concept of how to incorporate this is not necessarily complicated.
In fact, we can use the attention mechanism, discussed in Sec.~\ref{sec:att_mt},
almost as it is. Let us describe here a conceptual picture of how this can be
done.

Let $D=\left\{ \vd_1, \vd_2, \ldots, \vd_M \right\}$ be a set of knowledge
vectors. Each knowledge vector $\vd_i$ is a vector representation of a piece of
knowledge. For instance, $\vd_i$ can be a vector representation of one Wikipedia
article. It is certainly unclear what is the best way to obtain this vector
representation of an entire article, but let us assume that an oracle gave us a
means to do so.

Let us focus on recurrent language modelling from Sec.~\ref{sec:rlm}.\footnote{ 
    This approach of using attention mechanism for external knowledge pieces has
    been proposed recently in \cite{bordes2015large}, in the context of
    question-answering. Here, we stick to language modelling, as the course has
    not dealt with question-answering tasks.
}
At each time step, we have access to the following vectors:
\begin{enumerate}
    \itemsep 0em
    \item Context vector $\vh_{t-1}$: the summary all the preceding words
    \item Current word $w_t$: the current input word
\end{enumerate}
Similarly to what we have done in Sec.~\ref{sec:att_mt}, we will define a
scoring function $f_{\text{score}}$ which scores each knowledge vector $\vd_i$ with respect to
the context vector and the current word: 
\begin{align*}
    \alpha_{i,t} \propto \exp\left( f_{\text{score}}(\vd_i, \vh_{t-1}, \ve_{w_t}) \right),
\end{align*}
where $\ve_{w_t}$ is a vector representation of the current word $w_t$.  

This score reflects the relevance of the knowledge in predicting the next word,
and once it is computed for every knowledge vector, we compute the weighted sum
of all the knowledge:
\begin{align*}
    \tilde{\vd}_t = \sum_{i=1}^M \alpha_{i,t} \vd_i.
\end{align*}
This vector $\tilde{\vd}_t$ is a vector summary of the knowledge relevant to the
next word, taking into account the context phrase. In the case of an earlier
example, the scoring function gives a high score to the Wikipedia article on
``llama'' based on the history of preceding words ``Llama is a domesticated
camelid from''. 

This knowledge vector is used when updating the memory state of the recurrent
neural network:
\begin{align*}
    \vh_t = f_{\text{rec}}\left( \vh_{t-1}, \ve_{w_t}, \tilde{\vd}_t \right).
\end{align*}
From this updated memory state, which also contains the knowledge extracted from
the selected knowledge vector, the next word's distribution is computed
according to Eq.~\eqref{eq:rnn_y_h}.

One important issue with this approach is that the size of knowledge set $D$ is
often extremely large. For instance, English Wikipedia contains more than 5M
articles as of 23 Nov 2015.\footnote{
    \url{https://en.wikipedia.org/wiki/Wikipedia:Statistics}
} It easily becomes impossible to score each and every knowledge vector, not to
mention to extract knowledge vectors of all the articles.\footnote{
    This is true especially when those knowledge vectors are also updated during
    training.
} It is an open question how this unreasonable amount of computation needed for
search can be avoided.

\paragraph{Why is this any significant?}

One may naively think that if we train a large enough network with a large
enough data which contains all those world knowledge, a trained network will be
able to contain all those world knowledge (likely in a compressed form) in its
parameters together with its network architecture. This is true up to a certain
level, but there are many issues here. 

First, the world knowledge we're talking about here contains all the knowledge
accumulated so far. Even a human brain, arguably the best working neural network
to date, cannot store all the world knowledge and must resort to searching over
the external database of knowledge. It is no wonder we have libraries where
people can go and look for relevant knowledge. 

Second, the world knowledge is dynamic. Every day some parts of the world
knowledge become obsolete, and at the same time previously unknown facts are
added to the world knowledge. If anyone looked up ``Facebook'' before 2004, they
would've ended up with yearly facebooks from American universities. Nowadays, it
is almost certain that when a person looks up ``Facebook'', they will find
information on ``Facebook'' the social network site. Having all the {\em
current} world knowledge encoded in the model's parameters is not ideal in this
sense.

\section{Larger-Context Language Understanding: \\
    Beyond Sentences and Beyond Words}

If we view natural language as a function, it becomes clear that what we've
discussed so far throughout the course is heavily restrictive. There are two
reasons behind this restriction. 

First, what we have discussed so far has narrowly focused on handling a
sentence. In Sec.~\ref{sec:lm}, I have described language model as a way to
model a {\em sentence} probability $p(S)$. This is a bit weird in the sense that
we've been using a term ``language'' modelling not ``sentence'' modelling.
Keeping it in mind, we can start looking at a probability of a document or
discourse $D$ as a whole rather than as a product of sentence probabilities:
\begin{align*}
    p(D) = \prod_{k=1}^N p(S_k | S_{<k}),
\end{align*}
where the document $D$ consists of $N$ sentences. This approach is readily
integrated into the language modelling approaches we discussed earlier in
Chapter~\ref{chap:nlm} by
\begin{align*}
    p(D) = \prod_{k=1}^N \prod_{j=1}^{T_k} p(w_j | w_{<j}, S_{<k}).
\end{align*}
This is applicable to any language-related models we have discussed so far,
including neural language model from Sec.~\ref{sec:nlm}, recurrent language
model from Sec.~\ref{sec:rlm}, Markov random
field language model from Sec.~\ref{sec:cbow} and neural machine translation
from Chapter~\ref{chap:nmt}.

In the context of language modelling, two recent articles proposed to explore
this direction. I refer the readers to \cite{wang2015larger} and
\cite{ji2015document}.

Second, we have stuck to representing a sentence as a sequence of words so far,
despite a short discussion in Sec.~\ref{sec:ling_unit} where I strongly claim
that this does not have to be. This is indeed true, and in fact, even if we
replace most occurrence of ``word'' in this course with, for instance,
``character'', all the arguments stand. Of course, by using smaller units than
words, we run into many practical and theoretical issues. One most severe
practical issue is that each sentence suddenly becomes much longer. One most
sever theoretical issue is that it is a highly nonlinear mapping from a sequence
of characters to its meaning, as we discussed earlier in
Sec.~\ref{sec:ling_unit}. Nevertheless, the advance in computing and deep neural
networks, which are capable of learning such a highly nonlinear mapping, have
begun to let researchers directly work on this problem of using subword units
(see, e.g., \cite{kim2015character,ling2015character}.) Note that I am not
trying to say that characters are the only possible sub-word units, and recently
an effective statistical approach to deriving sub-word units off-line was
proposed and applied to neural machine translation in \cite{sennrich2015neural}.

\section{Warning and Summary}

Before I finish this lecture note with the summary of what we have discussed
throughout this course, let me warn you by quoting Claude Shannon
\cite{shannon1956bandwagon}:\footnote{
    I would like to thank Adam Lopez for pointing me to this quote.
}

\begin{quote}
    \it
    It will be all too easy for our somewhat artificial prosperity to collapse
    overnight when it is realized that the use of a few exciting words like
    information, entropy, redundancy, do not solve all our problems.
\end{quote}

Natural language understanding with distributed representation is a fascinating
topic that has recently gathered large interest from both machine learning and
natural language processing communities. This may give a wrong sign that this
approach with neural networks is an ultimate winner in natural language
understanding/processing, though without any ill intention. As Shannon pointed
out, this prosperity of distributed representation based natural language
understanding may collapse overnight, as can any other approaches out
there.\footnote{
    Though, it is interesting to note that information theory never really
    collapsed overnight. Rather its prosperity has been continuing for more than
    half a century since Shannon warned us about its potential overnight
    collapse in 1956.
} Therefore, I warn the readers, especially students, to keep this quote in
their mind and remember that it is not a few recent successes of this approach
to natural language understanding but the fundamental ideas underlying this
approach that matter and should be remembered after this course.

\paragraph{Summary}

Finally, here goes the summary of what we have learned throughout this semester.
We began our journey by a brief discussion on how we view human language as, and
we decided to stick to the idea that a language is a function not an entity
existing independent of the surrounding world, including speakers and listeners.
Is this a correct way to view a human language? Maybe, maybe not.. I will leave
it up to you to decide.

In order to build a machine that can approximate this language function, in
Chapter~\ref{chap:function_approx}, we studied basic ideas behind supervised
learning in machine learning. We defined what a cost function is, how we can
minimize it using an iterative optimization algorithm, specifically stochastic
gradient descent, and learned the importance of having a validation set for both
early-stopping and model selection. These are all basic topics that are dealt in
almost any basic machine learning course, and the only thing that I would like
to emphasize is the importance of {\em not looking at a held-out test set.} One
must always select anything related to learning, e.g., hyperparameters, networks
architectures and so on, based solely on a validation set. As soon as one tunes
any of those based on the test set performance, any result from this tuning
easily becomes invalid, or at least highly disputable.

In Chapter~\ref{chap:nn}, we finally talked about {\em deep} neural networks, or
more traditionally called multilayer perceptron.\footnote{
    I personally prefer ``multilayer perceptron'', but it seems like it
    has gone out of fashion.
} I tried to go over basic, but important details as slowly as possible,
including how to build a deep neural network based classifier, how to define a
cost function and how to compute the gradient w.r.t. the parameters of the
network. However, I must confess that there are better materials for this topic
than this lecture note.

We then moved on to recurrent neural networks in Chapter~\ref{chap:rnn}. This
was a necessary step in order to build a neural network based model that can
handle both variable-length input and output. Again, my goal here was to take as
much time as it is needed to motivate the need of recurrent networks and to give
you basic ideas underlying them. Also, I spent quite some time on why it has
been considered difficult to train recurrent neural networks by stochastic
gradient descent like algorithms, and as a remedy, introduced gated recurrent
units and long short-term memory units. 

Only after these long four to five weeks, have I started talking about how to
handle language data in Chapter~\ref{chap:nlm}. I motivated neural language
models by the lack of generalization and the curse of data sparsity. It is my
regret that I have not spent much time on discussing the existing techniques for
count-based $n$-gram language models, but again, there are much better materials
and better lecturers for these techniques already. After the introduction of
neural language model, I spent some time on describing how this neural language
model is capable of generalizing to unseen phrases. Continuing from this neural
language model, in Sec.~\ref{sec:rlm}, language modelling using recurrent neural
networks was introduced as a way to avoid Markov assumption of $n$-gram language
model.

This discussion on neural language model naturally continued on to neural
machine translation in Chapter~\ref{chap:nmt}. Rather than going directly into
describing neural machine translation models, I have spent a full week on two
issues that are often overlooked; data preparation in
Sec.~\ref{sec:parallel_corpora} and evaluation in Sec.~\ref{sec:bleu}. I wish
the discussion of these two topics has reminded students that machine learning
is not only about algorithms and models but is about a full pipeline starting
from data collection to evaluation (often with loops here and there.) This
chapter finished with where we are in 2015, compared to what Weaver predicted in
1949.

Of course, there are so many interesting topics in this area of natural language
understanding. I am not qualified nor knowledgeable to teach many, if not most,
of those topics unfortunately, and have focused on those few topics that I have
worked on myself. I hope this lecture note will serve at least as a useful
starting point into more advanced topics in natural language understanding with
distributed representations.

\bibliographystyle{abbrv}
\bibliography{lecture_note}

\end{document}